\documentclass{article}
\usepackage[T1]{fontenc}   %
\usepackage{microtype}
\usepackage{graphicx}
\usepackage{subcaption}
\usepackage{booktabs,multirow,array}
\usepackage{hyperref}

\usepackage[preprint]{icml2026}
\hypersetup{pdfsubject={}}
\usepackage{amsmath,amssymb,mathtools,amsthm}
\usepackage{xcolor}
\usepackage{colortbl}
\usepackage{tikz}
\usepackage{algorithm}
\usepackage{algorithmic}
\usepackage[capitalize,noabbrev]{cleveref}

\newtheoremstyle{plainnote}{\topsep}{\topsep}{\itshape}{}{\bfseries}{.}{.5em}{\thmname{#1}\thmnumber{ #2}\thmnote{. #3}}
\theoremstyle{plainnote}
\newtheorem{theorem}{Theorem}
\newtheorem{proposition}{Proposition}
\newtheorem{corollary}{Corollary}
\newtheorem{lemma}{Lemma}
\theoremstyle{definition}

\newtheoremstyle{proofhead}{\topsep}{\topsep}{\normalfont}{}{\bfseries}{.}{.5em}{\thmnote{#3}}
\theoremstyle{proofhead}
\newtheorem{proofA}{Proof}[section]
\crefname{proofA}{Proof}{Proofs}
\theoremstyle{plainnote}
\newcommand{\mtname}{}
\newtheorem{mtinner}{\mtname}
\newenvironment{mt}[2]{\renewcommand{\mtname}{#1}\renewcommand{\themtinner}{#2}\mtinner}{\endmtinner}

\newtheorem{slemma}{Lemma}[section]
\newtheorem{sprop}{Proposition}[section]
\newtheorem{scor}{Corollary}[section]
\newtheorem{sthm}{Theorem}[section]
\newtheoremstyle{defnote}{\topsep}{\topsep}{\normalfont}{}{\bfseries}{.}{.5em}{\thmname{#1}\thmnumber{ #2}\thmnote{. #3}}
\theoremstyle{defnote}
\newtheorem{sex}{Example}[section]
\newtheorem{sdef}{Definition}[section]
\theoremstyle{plain}
\providecommand{\M}{\mathbb{M}}
\providecommand{\Corr}{\operatorname{Corr}}
\providecommand{\rank}{\operatorname{rank}}
\providecommand{\spn}{\operatorname{span}}
\providecommand{\dist}{\operatorname{dist}}
\providecommand{\Ind}{\perp\!\!\!\perp}

\newcommand{\E}{\mathbb{E}}
\newcommand{\R}{\mathbb{R}}
\newcommand{\col}{\operatorname{col}}
\newcommand{\Cov}{\operatorname{Cov}}
\newcommand{\Var}{\operatorname{Var}}
\newcommand{\proj}{\operatorname{proj}}
\newcommand{\tr}{\operatorname{tr}}
\newcommand{\Zt}{\tilde Z}
\newcommand{\Wt}{\tilde W}
\newcommand{\Lb}{\bar L}
\newcommand{\Tz}{\Theta_z}
\newcommand{\Tw}{\Theta_w}
\newcommand{\pct}[1]{#1\%}
\newcommand{\ns}{$^{\dagger}$}
\newcommand{\Bias}{$|$Bias$|$}
\newcommand{\pending}[1]{\textcolor{red}{#1}}

\newcommand{\cmark}{\textcolor{green!45!black}{\checkmark}}
\newcommand{\xmark}{\textcolor{red!70!black}{$\boldsymbol{\times}$}}
\newcommand{\figorpending}[2]{\IfFileExists{#1}{\includegraphics[width=#2\textwidth]{#1}}{\fbox{\parbox{\dimexpr#2\textwidth-2\fboxsep-2\fboxrule}{\centering\vspace{2em}\pending{pending: \texttt{\detokenize{#1}}}\vspace{2em}}}}}

\icmltitlerunning{Completeness-Preserving Proxy Learning}

\begin{document}

\twocolumn[
  \icmltitle{Contrastive Proxy Forests: Completeness-Preserving Proxy Learning\texorpdfstring{\\}{ }from Multimodal Records}
  \icmlsetsymbol{corr}{*}
  \begin{icmlauthorlist}
    \icmlauthor{Kyoungeui Hong}{cs}
    \icmlauthor{Zijun Gao}{dso,corr}
    \icmlauthor{Ruishan Liu}{cs,corr}
  \end{icmlauthorlist}
  \icmlaffiliation{cs}{Department of Computer Science, University of Southern California, Los Angeles, CA, USA}
  \icmlaffiliation{dso}{Department of Data Sciences and Operations, USC Marshall School of Business, University of Southern California, Los Angeles, CA, USA}
  \icmlcorrespondingauthor{Zijun Gao}{zijungao@marshall.usc.edu}
  \icmlcorrespondingauthor{Ruishan Liu}{ruishanl@usc.edu}
  \icmlkeywords{proximal causal inference, negative controls, completeness, contrastive learning, multimodal, survival analysis}
  \vskip 0.3in
]
\printAffiliationsAndNotice{\textsuperscript{*}Co-corresponding authors. }

\begin{abstract}
Proximal causal inference removes unmeasured confounding using two proxies of the confounder, an outcome-side $W$ and a treatment-side $Z$. The proxies must satisfy exclusion restrictions and be \emph{complete}, meaning that they reflect every aspect of the confounder. In practice, they are a few variables hand-picked from a structured table, even though clinical notes and physiological waveforms carry rich information about the confounder. Such records are too high-dimensional for a proximal estimator, and generic compression such as PCA may keep modality-specific variation (style) instead of the confounder, destroying completeness. Our key observation is that, under the proximal condition, the confounder is the only link between a treatment-side and an outcome-side modality beyond the covariates. Learning what they \emph{share} therefore captures the confounder and filters out modality-specific style. Contrastive Proxy Forests (CPF) learns both proxies by contrasting the two modalities, given the covariates, on a separate unlabeled sample, and passes them to a proximal causal forest. This contrastive objective has an exact closed-form solution, ridge canonical correlation analysis. We prove that CPF preserves the exclusion restrictions exactly and that its proxies inherit the completeness of the raw records under conditions we characterize. Where style is strong, CPF lowers the error of conditional effects by 10--22\% relative to PCA proxies, and on MIMIC-IV it is the only method whose point estimate of warfarin's effect on stroke falls inside the range of randomized \mbox{trials}.
\end{abstract}

\newpage
\section{Introduction}\label{sec:intro}

Proximal causal inference \citep{miao2018,tchetgen2020} identifies a treatment effect despite an unmeasured confounder $U$, such as a patient's frailty, using two proxies of $U$. An outcome-side proxy $W$ may predict the outcome but is unaffected by the treatment, and a treatment-side proxy $Z$ may drive the treatment but has no direct effect on the outcome. Identification asks two things of them. The \emph{exclusion restrictions} are a design judgment that no data can verify, and every proximal analysis assumes them. \emph{Completeness} asks that the proxies vary with every aspect of $U$. When it fails, the estimate is biased, not just noisy. If the proxies reflect only part of $U$, the estimate adjusts for that part alone, and the rest of the confounding remains however large the sample (\cref{app:further}).

\begin{figure*}[t]
\centering
\begin{minipage}[b]{0.355\textwidth}
\centering
\resizebox{\linewidth}{!}{%
\renewcommand{\arraystretch}{1.3}%
\setlength{\tabcolsep}{3pt}%
\begin{tabular}{@{}l ccc@{}}
\toprule
& Uses the & Keeps their & Few, stable\\
Proxies from the records & records & completeness & dimensions\\
\midrule
A few hand-picked variables & \xmark & \xmark & \cmark\\
PCA / embeddings & \cmark & \xmark$^{*}$ & \cmark\\
Raw records & \cmark & \cmark & \xmark\\
\rowcolor{red!8}\textbf{CPF} & \cmark & \cmark & \cmark\\
\bottomrule
\multicolumn{4}{@{}l}{\footnotesize $^{*}$\,lost when style dominates (\cref{prop:pca})}
\end{tabular}}\\[6pt]
{\small (a) Four ways to obtain proxies}
\end{minipage}\hfill
\begin{minipage}[b]{0.625\textwidth}
\centering
\includegraphics[width=\textwidth]{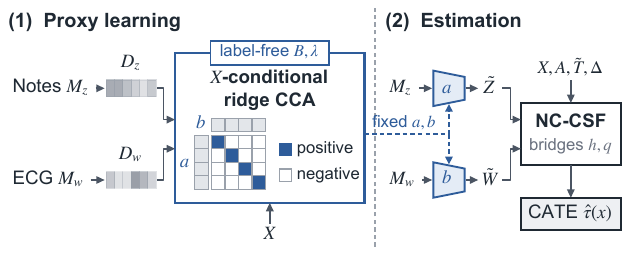}\\[-1pt]
{\small (b) The CPF pipeline}
\end{minipage}
\caption{(a) Four ways to obtain proxies from high-dimensional records. Only CPF, which keeps what the two modalities share, has all three properties. (b) Stage 1 builds raw-input dictionaries on an unlabeled pool and learns the proxy maps $a$ and $b$ in closed form by $X$-conditional ridge CCA, with the bandwidth $B$ and ridge $\lambda$ chosen without labels. Stage 2 applies the fixed maps to the analysis sample and estimates the CATE with NC-CSF.}
\label{fig:overview}
\end{figure*}

NC-CSF \citep{nccsf} integrates proximal inference with causal survival forests \citep[CSF;][]{cui2023csf} to estimate conditional average treatment effects (CATEs) on right-censored outcomes even when an unmeasured confounder exists. However, it is confined to a single modality, with $W$ and $Z$ hand-picked from one structured table. Thus, its conclusions depend on which variables are declared as proxies: in the ACTG175 analysis of NC-CSF, three clinically reasonable choices of $W$ and $Z$ moved the average CATE from 3.76 to 5.93 days.

Clinical notes and physiological waveforms carry rich information on frailty and severity, so they can play a central role as proxies of $U$. However, the common ways to use them each fail for a different reason (\cref{fig:overview}a). Hand-picking a few features discards most of what the records contain. Feeding the raw records directly forces a proximal estimator to learn its bridge functions over thousands of dimensions from the analysis sample alone, which makes the estimate unstable and the fit far slower (\cref{sec:raw}). PCA and pretrained encoders keep the directions of largest variance, which in records often capture \emph{style} rather than $U$. Style is modality-specific variation, such as the author and template of a note or the electrode placement of a waveform. For example, in the MIMIC-IV clinical notes, we found that the number, types and length of the reports explain 40\% of the variation in the top principal components beyond the covariates, whereas laboratory values explain less than 1\% (\cref{app:style}). Once style dominates, PCA proxies become uncorrelated with $U$ and completeness fails (\cref{prop:pca}).

Our starting point is one observation. Under the proximal condition, a treatment-side and an outcome-side modality are conditionally independent given the confounder and the covariates $X$, so \emph{whatever the two modalities share given $X$ is information about the confounder}, and style, which lives in one modality only, is not shared. This turns proxy selection into a representation-learning problem with an exact target: keep what the modalities share. Contrastive Proxy Forests (CPF; \cref{fig:overview}b, \cref{alg:cpf}) learns the two proxies by $X$-conditional contrastive learning between the modalities on an unlabeled pool disjoint from the analysis sample, using only the modalities and $X$, and passes them to a proximal causal forest.

What is new is not an estimator but a principled way to learn proxies, with guarantees for both requirements of proximal identification. Exclusion remains a design assumption, now made for whole modalities, and no learner can create it. We prove that learning on a disjoint pool preserves it exactly, whereas fitting CCA on the analysis sample breaks it. Completeness is where learning matters, and CPF is, to our knowledge, the first proxy learner with a completeness guarantee. Its proxies inherit the completeness of the raw records whenever the learned summary is sufficient for the confounder. In a linear factor model, this holds with as many proxies as the confounder's dimension. For a discrete confounder, it holds with one fewer proxy than the number of states, in the limit of rich dictionaries. We claim three contributions.
\begin{itemize}
\item \textbf{From proxy selection to proxy learning.} Instead of hand-picking proxies, CPF learns them from high-dimensional multimodal records without labels, keeping only what the two modalities share given $X$. This learning problem has an exact closed-form solution, $X$-conditional ridge CCA.
\item \textbf{Guarantees for both requirements.} Learning preserves the exclusion restrictions exactly for any pool size. The learned proxies are posterior summaries of the confounder that ignore style, and they inherit the completeness of the raw records under stated conditions. In a linear factor model, the best label-free compression is as efficient as the best compression learned with outcomes.
\item \textbf{General-purpose proxies from one label-free fit.} CPF learns the proxies from the two modalities and $X$ alone, without the treatment or the outcome, so the learned proxies are not tied to survival analysis. A single closed-form fit on an unlabeled pool can then serve any proximal estimator and outcome type, and we show this for survival, continuous and binary outcomes.
\end{itemize}

\section{Related work}\label{sec:related}

\textbf{Proximal inference.} \citet{miao2018} and \citet{tchetgen2020} established bridge-function identification from two kinds of proxies. \citet{cui2023semi} derived efficient estimators, \citet{sverdrup2023plearner} learn CATEs, \citet{ying2022survival} treat survival curves, and NC-CSF \citep{nccsf} uses causal survival forests. Kernel \citep{mastouri2021} and neural \citep{xu2021} bridge estimators accept high-dimensional proxies, and the neural one learns proxy features jointly with the outcome. \citet{sun2025} learn spectral representations of the proxies with contrastive objectives that also involve the outcome, fitted on the estimation sample. All of these take $W$ and $Z$ as given and give no completeness guarantee for the features they learn. CPF learns both proxies on a disjoint pool without the treatment or the outcome, which preserves exclusion (\cref{thm:exclusion}), and guarantees their completeness under stated conditions (\cref{prop:complete}).

\textbf{Choosing proxies.} Negative controls are usually chosen by subject-matter knowledge \citep{lipsitch2010negative}. DANCE \citep{kummerfeld2024dance} and Proxy-Rank \citep{xie2024proxy} use tetrad or rank tests to select, among given scalar variables, proxies that satisfy the exclusion restrictions. CPF instead assumes exclusion at the level of modalities and targets completeness, so the two approaches are complementary. \citet{chen2024} infer proxies from two text snippets with zero-shot models. Substitute confounders inferred from many observed variables \citep{kallus2018noisy,louizos2017cevae,wang2019blessings} are adjusted for as covariates, which cannot remove proxy noise (\cref{prop:bridge}). CPF instead uses the shared information inside proximal bridges.

\textbf{Contrastive learning and CCA.} Spectral contrastive losses are matrix factorizations \citep{haochen2021}, and self-supervised learning recovers spectral embeddings \citep{balestriero2022}. \Cref{thm:contrastive} extends this to $X$-conditional negatives. Multiview contrastive learning identifies the content shared by two views and discards view-specific style \citep{vonkugelgen2021content,daunhawer2023}. We read this causally. Under the proximal condition, the shared content is information about the confounder, and we formalize this with nonlinear canonical analysis \citep{lancaster1958,buja1990}.

\section{Problem setup}\label{sec:setup}

For each unit we observe covariates $X$, a binary treatment $A$, an outcome $Y$, a treatment-side modality $M_z$ and an outcome-side modality $M_w$. Each side may bundle several modalities. The target is the CATE $\tau(x)=\E[Y(1)-Y(0)\mid X=x]$. For right-censored survival we observe $\tilde T=\min(T,C)$ and $\Delta$, and the targets are the restricted mean survival time (RMST) effect, with $Y=\min(T,t_0)$, and the survival-probability effect, with $Y=1\{T>t_0\}$. Let $L=(U,V)$ collect the unmeasured confounder $U$ and a shared factor $V$ that enters both modalities but affects neither treatment nor outcome. Unlike style, which is specific to one modality, $V$ is shared by both, so the information CPF learns concerns $L$ rather than $U$ alone.

\textbf{The two requirements.} The proximal condition (P) states the exclusion restrictions for the modalities, $M_z\perp(Y(0),Y(1),C,M_w)\mid A,L,X$ and $M_w\perp(A,M_z)\mid L,X$, together with latent ignorability, positivity given $L$ and, for survival outcomes, NC-CSF's censoring assumptions. A proxy $S$ is \emph{complete} for $L$ if $\E[g(L)\mid S,A,X]=0$ implies $g(L)=0$. In words, no variation of $L$ is invisible to $S$. Stage 1 sees only an unlabeled pool $\{(M_z,M_w,X)\}$ independent of the analysis sample (S). Its task is to choose maps $a$ and $b$ such that the proxies $\Wt=b(M_w)$ and $\Zt=a(M_z)$ are complete and satisfy the exclusion restrictions of a proximal estimator, (E1) $\Zt\perp(Y(0),Y(1),C,\Wt)\mid A,L,X$ and (E2) $\Wt\perp(A,\Zt)\mid L,X$.

\textbf{Condition (P) in practice.} (P) states NC-CSF's exclusion restrictions (E1)--(E2) for whole modalities rather than for a few chosen variables. Like conventional proximal analyses, CPF assumes (P) and does not test it. Because every function of a modality inherits (P), assuming it for a whole modality is stronger than assuming it for a few variables. The roles of the modalities are therefore assigned by design. The treatment side holds records written before the treatment decision, with mentions of the treatment removed. The outcome side holds measurements taken before treatment, which the treatment cannot change. A record that may drive the decision can sit on the treatment side, as (P) allows.

\section{Method: Contrastive Proxy Forests}\label{sec:method}

\textbf{Raw-input dictionary.} Pretrained embeddings capture high-variance style first and can lose a weak shared signal (\cref{prop:pca}). CPF therefore represents each modality by a dictionary that concatenates its embedding $E$ with generic raw features: $D_z=[E_z,\ \text{position}\times\text{token one-hot}]$ for tokens and $D_w^{(B)}=[E_w,\ \log(1+|\mathrm{DFT}_{1..B}|)]$ for waveforms, whose magnitude spectrum does not depend on where the recording starts (\cref{prop:phase}). Columns are standardized with pool statistics.

\textbf{$X$-conditional contrastive learning, in closed form.} Let $R_z,R_w$ be the residuals of the dictionaries on $(1,X)$, $\Sigma_{zz},\Sigma_{ww}$ their covariances and $C$ their cross-covariance. With linear heads $f=\Tz^\top R_z$, $g=\Tw^\top R_w$, critic $s=f^\top g$, the two modalities of a unit as a positive pair, an outcome-side modality drawn independently \emph{at the same $X$} as a negative, and feature noise $\lambda I$ added to each view, the $X$-conditional spectral contrastive loss \citep{haochen2021} is $\mathcal L_\lambda=-2\E_{\rm pos}[s]+\E_{\rm neg}[s^2]$. Conditioning the negatives on $X$ removes what the modalities share only through the covariates. Its minimizer is $X$-conditional ridge CCA (\cref{thm:contrastive}): with $\Gamma_d,\Omega_d$ the top-$d$ singular vectors of $K_\lambda=(\Sigma_{zz}+\lambda I)^{-1/2}C(\Sigma_{ww}+\lambda I)^{-1/2}$, $\hat\Theta_z=(\Sigma_{zz}+\lambda I)^{-1/2}\Gamma_d$ and $\hat\Theta_w=(\Sigma_{ww}+\lambda I)^{-1/2}\Omega_d$.

\textbf{Label-free selection and Stage 2.} The bandwidth $B$ and the ridge $\lambda$ are chosen to maximize the held-out canonical correlation within the pool, without the treatment or the outcome. Stage 2 applies the fixed maps to the analysis sample and passes the proxies to a proximal estimator. We use NC-CSF for survival outcomes and its non-censored counterpart NC-CF \citep{nccsf} for continuous and binary outcomes. \cref{alg:cpf} summarizes the procedure.

\begin{algorithm}[t]
\caption{Contrastive Proxy Forests (CPF)}
\label{alg:cpf}
\begin{algorithmic}[1]
\STATE {\bfseries Input:} unlabeled pool $\mathcal P=\{(M_z,M_w,X)\}$; analysis sample $\mathcal A=\{(X,A,Y,M_z,M_w)\}$; proxy dimension $d$; grid $\mathcal G$ of bandwidths $B$ and ridges $\lambda$
\STATE {\bfseries Stage 1 (pool only; $A$, $Y$ never used)}
\FOR{$(B,\lambda)\in\mathcal G$}
\STATE Build dictionaries $D_z,D_w^{(B)}$; standardize and residualize them on $(1,X)$ with pool statistics
\STATE Fit ridge CCA (top-$d$ singular vectors of $K_\lambda$) on 80\% of $\mathcal P$
\STATE $J(B,\lambda)\gets$ sum of the $d$ canonical correlations on the other 20\%
\ENDFOR
\STATE $(\hat B,\hat\lambda)\gets\arg\max J$; refit on all of $\mathcal P$: $\hat\Theta_z,\hat\Theta_w$
\STATE Proxy maps $a(m_z)=\hat\Theta_z^\top(D_z(m_z)-\bar D_z)$, $b(m_w)=\hat\Theta_w^\top(D_w(m_w)-\bar D_w)$
\STATE {\bfseries Stage 2 (analysis sample)}
\STATE $\Wt_i\gets b(M_{w,i})$, $\Zt_i\gets a(M_{z,i})$ for every $i\in\mathcal A$
\STATE Fit the proximal estimator (NC-CSF for survival outcomes, NC-CF for continuous and binary outcomes) on $(X_i,A_i,Y_i,\Wt_i,\Zt_i)$
\STATE {\bfseries Output:} $\hat\tau(x)$
\end{algorithmic}
\end{algorithm}

\section{Theory}\label{sec:theory}

The results follow the two requirements. \Cref{thm:contrastive} solves the learning objective. \Cref{thm:exclusion} shows that exclusion is preserved, the most any procedure can offer since none can create (P). \Cref{thm:efficiency,thm:nonlinear} and \cref{prop:complete,prop:pca} concern completeness, where learning can help. \Cref{thm:contrastive,thm:efficiency} use three assumptions. (A1) states that $\E[D\mid X]$ is affine in $X$, and (A2) that the conditional covariance of $D$ given $X$ does not depend on $X$. (F) states that the dictionaries follow a linear factor model, $R_z=\Lambda_z\Lb+\varepsilon_z$ and $R_w=\Lambda_w\Lb+\varepsilon_w$ with $\Lb=L-\E[L\mid X]$ and each $\Lambda$ of column rank $k=\dim L$. The noises carry the style and are independent of each other and of $\Lb$, with arbitrary positive-definite covariances. Proof sketches are in \cref{app:theory}, further results in \cref{app:further}, and complete proofs in \cref{app:proofs}.

\begin{theorem}[Exact solution]\label{thm:contrastive}
Under (A1)--(A2), $\mathcal L_\lambda(\Tz,\Tw)=\lVert K_\lambda-\tilde\Theta_z\tilde\Theta_w^\top\rVert_F^2-\lVert K_\lambda\rVert_F^2$ with $\tilde\Theta=(\Sigma+\lambda I)^{1/2}\Theta$. Its minimizers are $X$-conditional ridge CCA up to an invertible map, for population and sample moments alike, so CPF has no optimization error.
\end{theorem}

\subsection{Learning preserves exclusion}
\begin{theorem}[Exact inheritance of exclusion]\label{thm:exclusion}
Under (P) and (S), the proxies of \cref{alg:cpf}, including the dictionary, standardization, selection of $(B,\lambda)$ and CCA, satisfy (E1)--(E2) exactly for every analysis unit, for any pool size and any outcome of the selection. The same CCA fitted on the analysis sample does not.
\end{theorem}
The maps depend on the pool alone, and conditional independence survives measurable maps. Fitting the same CCA on the analysis sample breaks exclusion. In the example of \cref{s:sec:thm2}, the partial correlation of the two proxies given $L$ and $X$ is $+0.64$ with in-sample fitting and $-0.02$ with a disjoint pool.

\begin{figure*}[t]
\centering
\figorpending{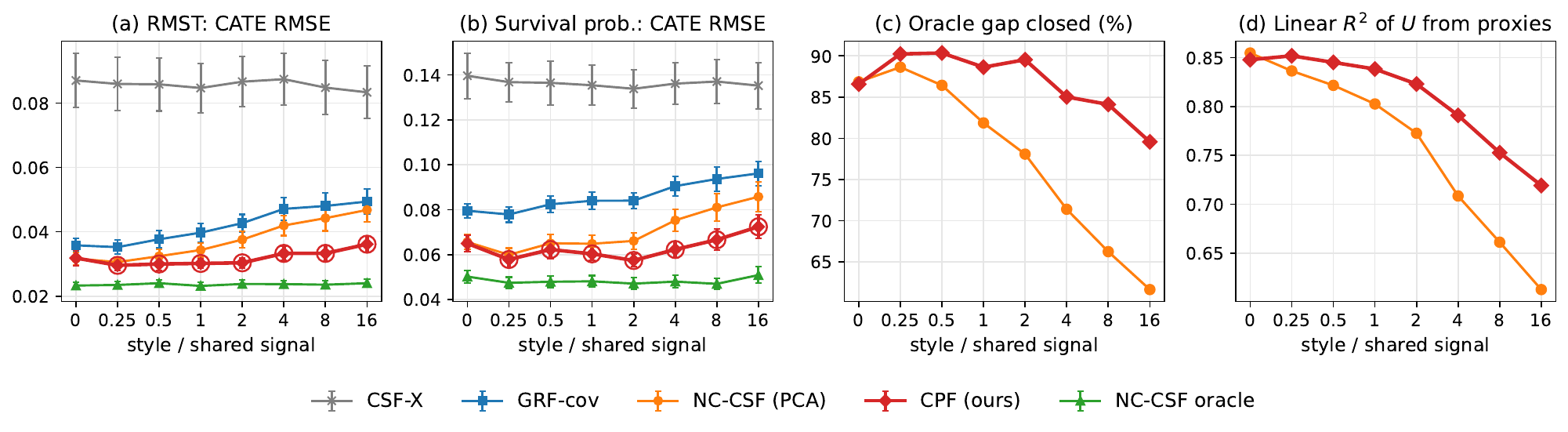}{0.98}
\caption{The predicted transition as style grows from 0 to 16 times the shared signal, with 60 datasets per point. (a, b) CATE RMSE with 95\% intervals. CSF-X adjusts for $X$ alone, GRF-cov uses the PCA proxies as covariates, and the oracle uses the true $U$. Circles mark significant paired differences between CPF and NC-CSF with PCA proxies. (c) Share of the RMST error gap between CSF-X and the oracle that each set of proxies closes. (d) Gain in the linear $R^2$ of $U$ when the proxies are added to $X$.}
\label{fig:phase}
\end{figure*}
\subsection{Learning preserves completeness}
\begin{theorem}[Information in the linear factor model]\label{thm:efficiency}
Assume (F). (a) The top $k$ canonical variates are an invertible transformation of the linear posterior mean $\E_{\rm lin}[\Lb\mid R_z]$, whatever the style covariance. (b) For proximal 2SLS, CCA minimizes the asymptotic variance for every outcome coefficient, so the best label-free compression is as good as the best compression that uses labels. (c) The cost of the ridge and of a finite pool of size $m$ is second order, $O_p((\lambda+m^{-1/2})^2)$.
\end{theorem}

Informally, without any linear model, whatever the two modalities share given $X$ is a posterior summary of $L$ that style cannot change.

\begin{theorem}[Without linearity]\label{thm:nonlinear}
Assume (P). Let $f_j$, $g_j$ and $\sigma_j$ be the nonlinear canonical functions and correlations of $(M_z,M_w)$ given $X$, defined as the singular system of the compact operator $g\mapsto\E[g(M_w)\mid M_z,X]$. (a) Each is a posterior mean of the confounder, $f_j(M_z)=\E[\phi_j(L)\mid M_z,X]$ and $g_j(M_w)=\E[\psi_j(L)\mid M_w,X]$. Style, any variable independent of the content and the other modality jointly, changes neither $f_j$ nor $\sigma_j$, whatever its size or form. (b) If the operator is Hilbert--Schmidt, the contrastive loss with nonlinear heads is minimized by the top-$d$ canonical functions. Under (A1)--(A2), CPF's CCA without ridge converges to them as the dictionaries grow.
\end{theorem}
The key step is that (P) makes the modalities independent given $L$ and $X$, so $\E[g(M_w)\mid M_z,X]=\E[\E[g(M_w)\mid L,X]\mid M_z,X]$ and whatever they share passes through $L$. The proxies are posterior summaries of $L$ rather than $L$ itself, so the bridges remain necessary, as \cref{prop:bridge} shows. When is such a summary complete?

\begin{proposition}[When the learned proxies are exactly complete]\label{prop:complete}
Let $S=s(M)$ be any proxy computed from a modality $M$ by a map fixed by the pool. (a) \emph{Learning cannot create completeness.} If $S$ is complete for $L$, so is $M$. (b) \emph{Sufficient summaries lose nothing.} If $L\perp M\mid S,A,X$, then $S$ is complete if and only if $M$ is. (c) \emph{CPF's proxies are complete in two cases.} (i) If (F) holds with $d\ge k$ and noise whose characteristic function vanishes nowhere, $\Wt$ is complete for every ridge and style covariance, and so is $\Zt$ when $M_z\perp A\mid L,X$. Neither is complete if $d<k$. (ii) If instead $L$ takes $K$ values and both modalities are complete, the top $d\ge K-1$ canonical functions are complete, and CPF's CCA converges to them by \cref{thm:nonlinear}(b). (d) \emph{Finite rank means a discrete confounder.} If both modalities are complete and the cross-modal operator has finite rank $r$, then $L$ takes at most $r+1$ values. For a continuous confounder, finitely many canonical functions are complete when they are sufficient, as in case (i), and in general the information they carry approaches that of the raw records as $d$ grows.
\end{proposition}

\begin{proposition}[Generic compression can destroy completeness]\label{prop:pca}
Assume (F). If style spans at least $d$ dimensions orthogonal to the signal, with variance above the largest eigenvalue of the shared signal, then PCA-$d$ proxies are uncorrelated with $\Lb$ and completeness fails. Linear autoencoders span the same subspace \citep{baldi1989}, whereas the CCA directions do not depend on the magnitude of style.
\end{proposition}

\textbf{Raw records versus learned proxies.} By \cref{prop:complete}(a), the raw records are complete whenever any summary of them is, and by parts (b) and (c), CPF loses none of their completeness when its summary is sufficient. The cost of the raw records lies in estimation. The proximal estimator must learn its bridges over thousands of columns from the analysis sample alone, while CPF learns the few directions that matter from an unlabeled pool. The proxy dimension $d$ therefore acts as a regularization level. Too few directions lose completeness, and many approach the raw records. \Cref{sec:raw} compares the two ends directly.

\section{Experiments}\label{sec:exp}
\begin{table}[t]
\centering\small
\caption{RMST CATE error of NC-CSF with proxies from different learners, all fitted on the same pool from the same dictionaries, and of covariate adjustment. $\Delta$: RMSE versus CPF (\%). Survival probability in \cref{app:results}. The values come from a separate run, so last digits can differ from \cref{tab:ablate,tab:outcomes}.}
\label{tab:main}
\setlength{\tabcolsep}{2.6pt}
\resizebox{\columnwidth}{!}{%
\begin{tabular}{@{}l rrr rrr@{}}
\toprule
& \multicolumn{3}{c}{Group A (strong style)} & \multicolumn{3}{c}{Group B (no style)}\\
\cmidrule(lr){2-4}\cmidrule(l){5-7}
Proxies & RMSE & \Bias & $\Delta$ & RMSE & \Bias & $\Delta$\\
\midrule
True $U$ (oracle) & .0232 & .0076 & & .0226 & .0069 & \\
\midrule
\textbf{CPF} & \textbf{.0374} & \textbf{.0221} & -- & \textbf{.0346} & \textbf{.0178} & --\\
PCA (NC-CSF) & .0452 & .0319 & $+20.8$ & .0356 & .0191 & $+2.9$\\
PLS & .0438 & .0299 & $+17.1$ & .0345 & .0177 & $-0.2$\ns\\
Deep CCA & .0430 & .0286 & $+14.9$ & .0356 & .0189 & $+2.8$\\
Embedding, uncompressed & .0562 & .0438 & $+50.1$ & .0377 & .0199 & $+9.0$\\
\midrule
GRF, all features & .0487 & .0349 & $+30.1$ & .0424 & .0254 & $+22.4$\\
$X$ only (CSF-X) & .0799 & .0694 & $+114$ & .0777 & .0655 & $+125$\\
\bottomrule
\end{tabular}}
\vspace{2pt}
{\footnotesize $^\dagger$\,$|z|<2$; every other difference is significant.}
\end{table}

\textbf{Synthetic data.} Starting from NC-CSF's data-generating model, we replace its tabular proxies with two records. A token sequence of length 64 over a 32-symbol vocabulary stands in for clinical notes, and a waveform of length 500 stands in for an electrocardiogram. $U$ enters the tokens through nonlinear features and the waveform through the amplitude, frequency and landmarks of a random-phase oscillation. To make the task harder, we also put a shared non-confounder $V$ and a linear index of $X$ into both records, so $U$ is not the only signal they share. CPF conditions on $X$ to remove the $X$ index, and the proxies may capture $V$ without breaking identification, because $V$ affects neither treatment nor outcome. Stage 1 uses an unlabeled pool of 10{,}000 units, and the analysis and test samples have 3{,}000 and 20{,}000 units. We score the CATE for the RMST and the survival probability at $t_0=0.75$ against the true unit-level effect, by RMSE and the absolute mean bias. \textbf{Group A} has strong style, so the theory predicts a gain. Its settings S8, HT3S, HBIS and HALL give 260 datasets, and in S8 style is eight times the shared signal. \textbf{Group B} has no style, so little gain is predicted. It holds the base setting D and its variations HV, HLOW, RARE, EMOD, MULTI and MISS, 440 datasets in all. \Cref{app:dgp} details every setting. All proxies are learned from the same pool and dictionaries, so they differ only in what they carry about $L$. Differences are tested with paired $z$ statistics.

\subsection{The predicted transition occurs}\label{sec:phase}
\begin{figure*}[t]
\centering
\IfFileExists{r6/fig_tradeoff.pdf}{\includegraphics[width=0.98\textwidth]{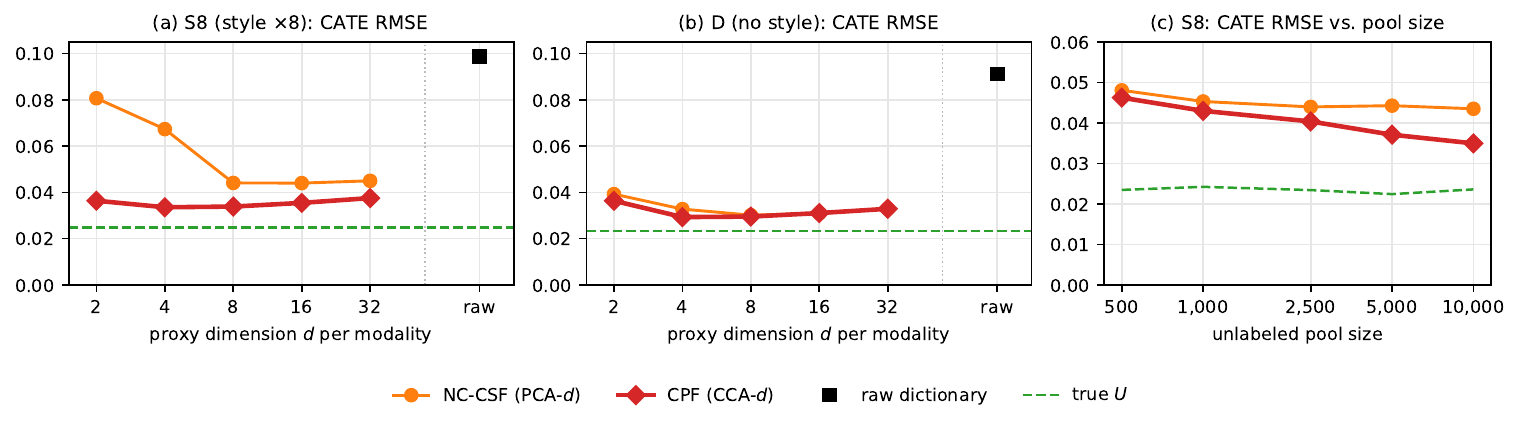}}{\pending{pending: r6/fig\_tradeoff.pdf}}
\caption{How much to compress, and how much unlabeled data. RMST CATE RMSE with 60 datasets per point. (a, b) Proxy dimension $d$ per modality, with and without style. The uncompressed raw dictionary of about 2{,}200 columns sits at the end of the axis. (c) Size of the unlabeled pool on the S8 design, with the analysis sample fixed at 3{,}000. CSF-X, which adjusts for $X$ alone, is off the scale at about .085. Dashed lines show NC-CSF with the true $U$.}
\label{fig:tradeoff}
\end{figure*}
\Cref{prop:pca} predicts that, as style grows with the signal fixed, PCA proxies lose the confounder while the canonical directions keep it. \Cref{fig:phase} shows this transition. Without style the two coincide, with RMST RMSEs of .0318 for CPF and .0317 for PCA proxies. As style grows, PCA proxies break down while CPF's hold up. The difference in completeness is large. PCA proxies lose much of the confounder, and the linear $R^2$ of $U$ falls from .85 to .61. CPF's proxies keep most of it, and their $R^2$ falls only to .72. The estimates follow the same pattern. NC-CSF with PCA proxies loses \pct{48} in accuracy, and its share of the oracle gap falls from \pct{87} to \pct{62}. CPF loses only \pct{13} and keeps \pct{80}. The paired difference is significant for both targets from style $\times$0.25 on.

\subsection{Main benchmark}\label{sec:main}
CPF was never significantly less accurate than NC-CSF with PCA proxies in any of 102 setting-level comparisons, 66 across the 11 settings and 36 more in the scale sweeps (\cref{app:results}). \textbf{Where style is strong (group A)}, CPF outperformed PCA proxies in all 24 combinations of four settings, two targets and three metrics, with $|z|\ge5.1$. It lowered RMSE by \pct{17.3} and \pct{12.5} and mean bias by \pct{30.9} and \pct{23.5} for RMST and the survival probability, and it closed \pct{71}--\pct{77} of the oracle gap against \pct{59}--\pct{63} for PCA proxies (\cref{tab:main}). Bias fell about twice as much as RMSE. This is what incompleteness predicts, because PCA proxies leave part of the confounder unadjusted. Every other learner, including PLS \citep{wegelin2000pls} and Deep CCA \citep{andrew2013dcca}, was significantly less accurate in group A. GRF-all \citep{athey2019grf}, which adjusts for more than 2{,}000 raw features, was \pct{30} worse. \Cref{prop:bridge} predicts this, since covariate adjustment cannot remove proxy noise. \textbf{Without style (group B)}, CPF performed on par with PCA, PLS and Deep CCA, as predicted.

\subsection{Raw records versus learned proxies}\label{sec:raw}
\begin{table}[t]
\centering\small
\caption{Raw records versus learned proxies in the same estimator. CATE RMSE on the d-sweep datasets, 60 per setting, with CPF's proxies, PCA proxies of the same dictionaries, the full raw dictionary without compression, and the true $U$. CPF and PCA use 8 proxies per modality, and the raw dictionary has about 2{,}200 columns on S8.}
\label{tab:raw}
\setlength{\tabcolsep}{3pt}
\begin{tabular}{@{}l rr rr@{}}
\toprule
& \multicolumn{2}{c}{Survival, RMST} & \multicolumn{2}{c}{Continuous}\\
& \multicolumn{2}{c}{(NC-CSF)} & \multicolumn{2}{c}{(NC-CF)}\\
\cmidrule(lr){2-3}\cmidrule(l){4-5}
Proxies & S8 & D & S8 & D\\
\midrule
\textbf{CPF} & \textbf{.0341} & \textbf{.0295} & \textbf{.1250} & \textbf{.0922}\\
PCA & .0440 & .0301 & .1846 & .0958\\
Raw, uncompressed & .0987 & .0911 & .1679 & .1115\\
True $U$ & .0248 & .0234 & .0770 & .0816\\
\bottomrule
\end{tabular}
\end{table}
\Cref{tab:raw} compares three ways of obtaining proxies from the records within the same estimator. PCA proxies lose to CPF's when style is present, by \pct{29} for RMST and \pct{48} for the continuous outcome on S8. On the style-free D the two tie, as \cref{prop:pca} predicts.
The raw dictionary was significantly less accurate than CPF's proxies in all eight comparisons, four outcomes in two settings (\cref{app:raw}). Its error was about three times as large for RMST, four times as large for the survival probability, and \pct{9}--\pct{34} larger for continuous and binary outcomes. The loss lies in estimation, not identification, because by \cref{prop:complete}(a) the raw records are never less complete than CPF's proxies. It persists without style, so it comes from dimension. The estimator must fit its nuisance models on about 2{,}200 columns using the analysis sample alone. In NC-CSF, each treatment arm of a training fold has only about 1{,}000 units, so the arm-specific and censoring Cox models \citep{cox1972} overfit and the weighted scores built on them become unstable. NC-CF has no Cox model and loses less. The raw records are also slow. A single NC-CSF fit took a median of almost five hours on them, and a day or more on the slowest datasets, against about 20 seconds on CPF's proxies. CPF's proxies are therefore both more accurate and more than 800 times faster to fit.

\subsection{How much to compress, and how much unlabeled data}\label{sec:tradeoff}
\Cref{fig:tradeoff}a,b places the raw records at the end of the dimension sweep, on the same datasets. Under style, CPF's error barely depends on $d$. Its RMST RMSE stays between .0336 and .0376 for $d=2$ to $32$ and is lowest at $d=4$ to $8$. PCA's error stops falling only at $d=8$ and never reaches CPF's. Its best value, .0440 at $d=16$, is \pct{31} above CPF at $d=4$. The leading PCA directions are style, so PCA needs more of them to reach the confounder. Without style the two compressions coincide from $d=8$ on. In both settings the uncompressed dictionary, about 2{,}200 columns, has about three times CPF's error. The proxy dimension thus governs a trade-off between completeness and estimation. Too few directions cannot carry $U$ (\cref{prop:complete}), and too many leave the estimator noise to fit. Because Stage 1 needs no labels, its input can also grow at no cost to the analysis (\cref{fig:tradeoff}c). With the analysis sample fixed, enlarging the unlabeled pool from 500 to 10{,}000 records lowered CPF's RMST error from .0463 to .0350, while NC-CSF with PCA proxies improved only from .0481 to .0435. CPF's advantage grew from \pct{3.8} to \pct{19.6} for RMST and from \pct{2.5} to \pct{17.3} for the survival probability, consistent with \cref{thm:efficiency}(c).

\subsection{What each component contributes}\label{sec:ablation}
\begin{table}[t]
\centering\small
\caption{RMST CATE error when one component of CPF is changed. In order, the rows remove $X$-conditioning, adjust for CPF's proxies as covariates without bridges, use InfoNCE heads \citep{oord2018cpc} instead of the closed form, use PCA instead of CCA on the same dictionary, use pretrained embeddings instead of the dictionary, and use PCA of pretrained embeddings, the common practice. $\Delta$ is the RMSE change relative to CPF in \%, paired over datasets. The two embedding rows cover 249 datasets in group A and 427--428 in group B. \Cref{app:results} adds MAE.}
\label{tab:ablate}
\setlength{\tabcolsep}{2.4pt}
\resizebox{\columnwidth}{!}{%
\begin{tabular}{@{}l rrr rrr@{}}
\toprule
& \multicolumn{3}{c}{Group A (strong style)} & \multicolumn{3}{c}{Group B (no style)}\\
\cmidrule(lr){2-4}\cmidrule(l){5-7}
Variant & RMSE & \Bias & $\Delta$ & RMSE & \Bias & $\Delta$\\
\midrule
\textbf{CPF} & \textbf{.0374} & .0221 & -- & \textbf{.0347} & .0179 & --\\
No $X$-conditioning & .0376 & .0223 & $+0.4$ & .0349 & .0180 & $+0.5$\\
No bridges & .0377 & .0205 & $+0.8$ & .0371 & .0174 & $+6.9$\\
InfoNCE heads & .0483 & .0350 & $+29.1$ & .0547 & .0403 & $+57.8$\\
PCA, not CCA & .0452 & .0320 & $+20.9$ & .0356 & .0191 & $+2.7$\\
Embeddings, CCA & .0473 & .0330 & $+23.3$ & .0402 & .0240 & $+14.5$\\
Embeddings, PCA & .0624 & .0508 & $+62.8$ & .0388 & .0227 & $+10.7$\\
\bottomrule
\end{tabular}}
\end{table}
\Cref{tab:ablate} changes one component of CPF at a time. Where style is strong, the canonical directions and the raw-input dictionary each matter. PCA in place of CCA on the same dictionary raised RMSE by \pct{21}, and pretrained embeddings in place of the dictionary raised it by \pct{23}. Together they separate CPF from common practice, NC-CSF on the PCA of pretrained embeddings, by \pct{39} in RMSE and \pct{55} in mean bias. Solving the objective in closed form matters in both groups. Learned InfoNCE heads raised RMSE by \pct{29} and \pct{58}, whereas \cref{thm:contrastive} gives the exact minimizer of the spectral contrastive loss. The bridges matter where adjustment alone falls short. Using CPF's proxies as covariates raised RMSE by \pct{7} in group B and by \pct{17}--\pct{20} for the survival probability (\cref{app:results}). \Cref{prop:bridge} predicts this, since covariate adjustment cannot remove proxy noise. Removing $X$-conditioning changed RMSE by less than \pct{1}, as \cref{cor:xcond} predicts when $X$ enters each modality through one linear index. Its role is to guard against covariates that dominate a modality.

\subsection{One set of proxies for survival, continuous and binary outcomes}\label{sec:outcomes}
\begin{table}[t]
\centering\small
\caption{The same CPF proxies for three outcome types. CATE RMSE with CPF and PCA proxies of the same dictionaries, and the change $\Delta$ in \%. Continuous and binary outcomes are redrawn on the benchmark datasets from each case's linear predictor, in the strong-confounding cases (\cref{app:general}). All 8 differences are significant.}
\label{tab:outcomes}
\setlength{\tabcolsep}{2.2pt}
\begin{tabular}{@{}l rrr rrr@{}}
\toprule
& \multicolumn{3}{c}{Group A} & \multicolumn{3}{c}{Group B}\\
\cmidrule(lr){2-4}\cmidrule(l){5-7}
Outcome (estimator) & CPF & PCA & $\Delta$ & CPF & PCA & $\Delta$\\
\midrule
RMST (NC-CSF) & \textbf{.0374} & .0452 & $-17.3$ & \textbf{.0347} & .0356 & $-2.6$\\
Surv.\ prob.\ (NC-CSF) & \textbf{.0703} & .0804 & $-12.5$ & \textbf{.0665} & .0679 & $-2.1$\\
Continuous (NC-CF) & \textbf{.1716} & .2252 & $-23.8$ & \textbf{.1887} & .1952 & $-3.3$\\
Binary (NC-CF) & \textbf{.0503} & .0571 & $-11.8$ & \textbf{.0521} & .0526 & $-1.0$\\
\bottomrule
\end{tabular}
\end{table}
\begin{table*}[t]
\centering\scriptsize
\caption{Real data with answers from randomized controlled trials (RCTs). Average RMST effects with 95\% bootstrap intervals from 250 replicates. MIMIC-IV follows death and ischemic stroke up to 360 and 243 days after discharge, in days, and NHANES follows death up to 10 years, in months. CSF ($X$ only) is the same forest without proxies, and NC-CSF (hand-picked) is NC-CSF's construction from tabular variables. RCT answers are carried to each cohort by applying the trial hazard ratio to its Kaplan--Meier curve (\cref{app:real}).}
\label{tab:real}
\setlength{\tabcolsep}{4pt}
\resizebox{\textwidth}{!}{\IfFileExists{r5/tab_real_body_en.tex}{\begin{tabular}{@{}ll rrrrr c@{}}
\toprule
& & Unadjusted & CSF ($X$ only) & NC-CSF (hand-picked) & NC-CSF (PCA) & \textbf{CPF} & RCT\\
\midrule
\multicolumn{8}{@{}l}{\textit{MIMIC-IV, atrial fibrillation: clinical notes (treatment side), 12-lead ECG (outcome side)}}\\
Warfarin $\to$ death & RMST (days) & $+18.92$ & $+18.46$ & $+17.98$ & $+16.53$ & $\mathbf{+15.95}$ & $+2.5$ to $+21.9$\\
 & & & {\tiny$(+14.99, +23.32)$} & {\tiny$(+13.54, +21.55)$} & {\tiny$(+12.12, +20.52)$} & {\tiny$(+11.86, +19.74)$} & \\[1pt]
\cmidrule(l){2-8}
Warfarin $\to$ stroke & RMST (days) & $+0.86$ & $+0.62$ & $+2.82$ & $+2.02$ & $\mathbf{+1.84}$ & $+1.15$ to $+1.86$\\
 & & & {\tiny$(-0.53, +1.94)$} & {\tiny$(+0.90, +4.43)$} & {\tiny$(+0.10, +3.31)$} & {\tiny$(-0.00, +2.85)$} & \\[1pt]
\midrule
\multicolumn{8}{@{}l}{\textit{NHANES 2003--2006: dietary recall (treatment side), accelerometry (outcome side)}}\\
Supplements & RMST (months) & $-1.14$ & $+0.50$ & $+0.49$ & $-0.23$ & $\mathbf{+0.19}$ & $\approx 0$\\
 & & & {\tiny$(-1.14, +1.96)$} & {\tiny$(-0.75, +1.78)$} & {\tiny$(-1.39, +0.96)$} & {\tiny$(-1.02, +1.43)$} & \\[1pt]
\cmidrule(l){2-8}
Statins & RMST (months) & $-5.34$ & $+3.38$ & $+3.41$ & $+3.03$ & $\mathbf{+2.94}$ & $+0.8$ to $+1.8$\\
 & & & {\tiny$(+1.88, +4.98)$} & {\tiny$(+2.02, +4.77)$} & {\tiny$(+1.58, +4.44)$} & {\tiny$(+1.64, +4.57)$} & \\[1pt]
\bottomrule
\end{tabular}
}{\pending{pending}}}
\end{table*}
Stage 1 uses only the modalities and $X$, never the treatment or the outcome, so the proxies are not tied to survival. We redrew the outcome of the benchmark datasets as a continuous or a binary variable and reused the same proxies. Across the three outcome types, the CPF proxies were significantly more accurate than the PCA proxies in all 8 comparisons (\cref{tab:outcomes}). Where style is strong, CPF lowered RMSE by \pct{23.8} for the continuous outcome and \pct{11.8} for the binary one, a gain comparable to that for survival. It closed \pct{81} and \pct{87} of the oracle gap for these two outcomes, against \pct{70} and \pct{78} with PCA proxies.

\subsection{Real data with answers known from trials}\label{sec:real}
Real records have no unit-level truth, so we ask questions whose answers are known from RCTs (\cref{tab:real}). In \textbf{MIMIC-IV} \citep{johnson2023mimiciv} we compare warfarin with no anticoagulation after a first admission with atrial fibrillation, in 7{,}336 patients. Frailty confounds this comparison, because untreated patients often have contraindications or a short life expectancy. The treatment side is the clinical notes \citep{johnson2023note} written before the decision, with every sentence on anticoagulation or the electrocardiogram removed. The outcome side is the latest 12-lead electrocardiogram before treatment \citep{gow2023ecg}. The trial answer comes from a Cochrane meta-analysis \citep{aguilar2005}. For stroke, CPF is the only method whose estimate falls inside the trial range. It estimates $+1.84$ days of RMST, against a trial range of $+1.15$ to $+1.86$ days. Adjustment for $X$ misses most of the benefit, with $+0.62$ days and an interval covering zero. The hand-picked and PCA proxies overstate it, with $+2.82$ and $+2.02$ days. For death, every estimate lies inside the wide trial range, and CPF's is the closest to the trial value. The notes show why PCA proxies fail here. Beyond $X$, the composition of the documentation explains 40\% of the variance of the notes' PCA proxies but only 7\% of CPF's, so the notes are in the style regime of \cref{prop:pca} (\cref{app:style}).

In \textbf{NHANES} \citep{nhanes}, with dietary recalls and accelerometry as the modalities, the methods differ little. Randomized trials found no effect of supplements \citep{gaziano2012,sesso2012mvm}, and CPF estimates $+0.19$ months of 10-year RMST, against $-1.14$ unadjusted. For statins, confounding by indication makes users appear to lose 5.34 months in the unadjusted comparison. Every adjusted estimate reverses this sign but overshoots the gain implied by the CTT trials \citep{ctt2010}, $+0.8$ to $+1.8$ months. CPF overshoots it least, at $+2.94$ months. CPF is thus the closest to the trial answers in NHANES as well, but only by small margins. In both cohorts the intervals overlap, so these comparisons check agreement with the trials rather than separate the methods statistically.

\section{Discussion}\label{sec:discussion}

CPF changes how proximal inference obtains its proxies, not what it assumes about them. The exclusion restrictions stay a design assumption, now made for whole modalities, and learning on a disjoint pool keeps them intact. Learning can protect completeness, whose failure biases the estimate. CPF's proxies keep the completeness of the raw records under conditions we characterize, whereas generic compression can lose it. Theory and experiments agree on when this matters. The gain grows with modality-specific variation, and without it CPF loses nothing. The raw records are no way around compression. They are never less complete, yet they left the estimator less accurate and far slower on an analysis sample of ordinary size. Naive CCA is not enough either. Fitted on the analysis sample it breaks exclusion (\cref{thm:exclusion}), without $X$-conditioning it can spend its directions on the covariates (\cref{cor:xcond}), and without a ridge it finds perfect correlations in noise (\cref{thm:selection}).

\textbf{Limitations.} CPF neither tests nor relaxes the exclusion restrictions, and (P) for whole modalities is stronger than for a few variables. Choosing \emph{valid} proxies from records remains open and is complementary to this work. Completeness is preserved, not created. A confounder the records do not reflect cannot be recovered (\cref{prop:complete}(a)). The exact conditions cover additive noise and discrete confounders. With heteroscedastic indicator noise or finitely many proxies for a continuous confounder, completeness holds only approximately. The efficiency results cover linear proximal estimators. Real-data trial answers are transported from trial populations to routine-care cohorts.

\bibliography{refs}
\bibliographystyle{icml2026}

\newpage
\appendix
\onecolumn

\section{Proofs}\label{app:theory}

This appendix gives the main steps of the proofs of \cref{thm:contrastive,thm:exclusion,thm:efficiency,thm:nonlinear} and the complete proof of \cref{prop:complete}. \Cref{app:proofs} gives complete proofs of all results.

\textbf{Notation.} Under (F), let $\Sigma_L=\Var\Lb$, and let $\Psi_z=\Var\varepsilon_z$ and $\Psi_w=\Var\varepsilon_w$ be the style covariances. Then
\[
\Sigma_{zz}=\Lambda_z\Sigma_L\Lambda_z^\top+\Psi_z,\qquad
\Sigma_{ww}=\Lambda_w\Sigma_L\Lambda_w^\top+\Psi_w,\qquad
C=\Lambda_z\Sigma_L\Lambda_w^\top .
\]
$\Theta_z^{(k)}(\lambda)$ and $\Theta_w^{(k)}(\lambda)$ are the top $k$ ridge CCA directions at ridge $\lambda$, $\proj(V\mid S)$ is the $L^2$ projection of $V$ onto the linear span of $S$, and $\perp$ denotes (conditional) independence.

\begin{proofA}[Proof of \cref{thm:contrastive}]\label{prf:contrastive}
Write $N=\Tz\Tw^\top$. The feature noises $\xi_z,\xi_w$ have mean zero and covariance $\lambda I$ and are independent of the data, so the positive term is
\[
\E_{\rm pos}[s]=\E\big[(R_z+\xi_z)^\top N(R_w+\xi_w)\big]=\E\big[R_z^\top NR_w\big]=\tr\big(\Tz^\top C\,\Tw\big).
\]
For a negative pair, let $a=R_z+\xi_z$ and $b=R'_w+\xi'_w$, where $R'_w$ is drawn given the same $X$, independently of $R_z$. Then
\begin{align*}
\E[s^2\mid X]&=\E\big[\tr\big(N^\top aa^\top N\,bb^\top\big)\mid X\big]\\
&=\tr\big(N^\top\,\E[aa^\top\mid X]\,N\,\E[bb^\top\mid X]\big)\\
&=\tr\big(\Tz^\top(\Sigma_{zz}+\lambda I)\Tz\,\Tw^\top(\Sigma_{ww}+\lambda I)\Tw\big),
\end{align*}
where the second equality uses $a\perp b\mid X$, and the third uses (A1)--(A2), which give $\E[aa^\top\mid X]=\Cov(D_z\mid X)+\lambda I=\Sigma_{zz}+\lambda I$, and likewise for $b$. With $\tilde\Theta_z=(\Sigma_{zz}+\lambda I)^{1/2}\Tz$ and $\tilde\Theta_w=(\Sigma_{ww}+\lambda I)^{1/2}\Tw$, the two terms combine into
\begin{align*}
\mathcal L_\lambda(\Tz,\Tw)&=-2\tr\big(\tilde\Theta_z^\top K_\lambda\tilde\Theta_w\big)+\tr\big(\tilde\Theta_z^\top\tilde\Theta_z\,\tilde\Theta_w^\top\tilde\Theta_w\big)\\
&=\big\lVert K_\lambda-\tilde\Theta_z\tilde\Theta_w^\top\big\rVert_F^2-\lVert K_\lambda\rVert_F^2 .
\end{align*}
As $(\Tz,\Tw)$ ranges over all pairs, $\tilde\Theta_z\tilde\Theta_w^\top$ ranges over all matrices of rank at most $d$. By the Eckart--Young theorem \citep{eckart1936}, the minimum is $-\sum_{j\le d}\rho_j(\lambda)^2$, where $\rho_j(\lambda)$ are the singular values of $K_\lambda$, attained at $\tilde\Theta_z\tilde\Theta_w^\top=\Gamma_dS_d\Omega_d^\top$ with $S_d=\mathrm{diag}(\rho_1(\lambda),\dots,\rho_d(\lambda))$. Undoing the whitening, if $\rho_d(\lambda)>\rho_{d+1}(\lambda)$, the minimizers are exactly
\[
\big(\Tz,\Tw\big)=\big(\hat\Theta_zS_d^{1/2}G,\ \hat\Theta_wS_d^{1/2}G^{-\top}\big),\qquad G\in\R^{d\times d}\ \text{invertible}.
\]
Only $\Sigma+\lambda I\succ0$ was used, so the same algebra holds verbatim for the pool moments $(\hat\Sigma_{zz},\hat\Sigma_{ww},\hat C)$, with positive and negative pairs drawn from the pool rows.
\end{proofA}

\begin{proofA}[Proof of \cref{thm:exclusion}]\label{prf:exclusion}
Let $\omega$ collect the pool and the seeds. Every step of Stage~1 (dropping rare one-hot columns, the spectra, standardization, residualization on $[1,X]$, the selection of $(B,\lambda)$ and the final ridge CCA) is a measurable function of $\omega$ alone. Treatment, time and event never enter. Hence $\Zt=a_\omega(M_z,X)$ and $\Wt=b_\omega(M_w,X)$ for maps fixed by $\omega$.

\emph{Fixed maps.} If $Q_1\perp Q_2\mid Q_3$, then $h_1(Q_1,Q_3)\perp h_2(Q_2,Q_3)\mid Q_3$ for all measurable $h_1,h_2$ \citep[Lemma~4.2]{dawid1979}. Applied to the two parts of (P),
\begin{align*}
M_z\perp\big(Y(0),Y(1),C,M_w\big)\mid A,L,X\quad&\Longrightarrow\quad a(M_z,X)\perp\big(Y(0),Y(1),C,b(M_w,X)\big)\mid A,L,X,\\
M_w\perp\big(A,M_z\big)\mid L,X\quad&\Longrightarrow\quad b(M_w,X)\perp\big(A,a(M_z,X)\big)\mid L,X,
\end{align*}
which are (E1) and (E2).

\emph{Maps fitted on the pool.} By (S), each analysis unit $O$ is independent of $\omega$. Conditionally on $\omega=\omega_0$, the law of $(O,a_\omega,b_\omega)$ is therefore that of $(O,a_{\omega_0},b_{\omega_0})$ with fixed maps, to which the first part applies.

\emph{Maps fitted on the analysis sample.} Let $M_z$ and $M_w$ be scalar noise independent of everything else, so that (P) holds. One-dimensional CCA fitted on $n$ analysis units gives standardized variates with
\[
\E\big[\Zt_i\Wt_i\big]=\E\,|\hat r|>0,\qquad \E\Zt_i=\E\Wt_i=0,
\]
where $\hat r$ is the sample correlation. The two proxies are therefore dependent, also given $(L_i,X_i)$, and (E2) fails for every unit.
\end{proofA}

\begin{proofA}[Proof of \cref{thm:efficiency}]\label{prf:efficiency}
(a) Since $\Sigma_L\Lambda_w^\top$ has full row rank $k$, $\col C=\col\Lambda_z$ and $K_\lambda$ has exactly $k$ nonzero singular values. Its top $k$ left singular vectors span $(\Sigma_{zz}+\lambda I)^{-1/2}\col\Lambda_z$, so
\[
\col\Theta_z^{(k)}(\lambda)=\col\big((\Sigma_{zz}+\lambda I)^{-1}\Lambda_z\big),\qquad \lambda\ge0 .
\]
At $\lambda=0$, the linear posterior mean of the confounder is
\[
\E_{\rm lin}[\Lb\mid R_z]=\Sigma_L\Lambda_z^\top\Sigma_{zz}^{-1}R_z=\Upsilon^\top R_z,\qquad \Upsilon=\Sigma_{zz}^{-1}\Lambda_z\Sigma_L ,
\]
and $\col\Upsilon=\col(\Sigma_{zz}^{-1}\Lambda_z)$. Both matrices have $k$ independent columns, so $\Theta_z^{(k)}(0)=\Upsilon N$ for an invertible $N$, and the top $k$ canonical variates equal $N^\top\E_{\rm lin}[\Lb\mid R_z]$, whatever $\Psi_z$ is.

(b) Consider proximal 2SLS \citep{tchetgen2020} in the linear proximal model $Y=c_0+\tau A+\beta^\top\Lb+\gamma^\top X+e$ (\cref{s:sec:thm3}, (L1)--(L4)), with instruments $(1,A,X,P_z^\top D_z)$ and regressors $(1,A,X,P_w^\top D_w)$, where $\Lambda_w^\top P_w$ is invertible. Its asymptotic variance factorizes into an outcome-side and a treatment-side factor,
\[
\mathcal V(P_z,P_w)=\underbrace{\big(\sigma_e^2+\vartheta_{P_w}^\top\Psi_w\vartheta_{P_w}\big)}_{\text{outcome side}}\ \underbrace{\Big(\frac1{s_A}+\alpha^\top H(P_z)^{-1}\alpha\Big)}_{\text{treatment side}},\qquad \vartheta_{P_w}=P_w(\Lambda_w^\top P_w)^{-1}\beta,
\]
where $s_A$ is the residual variance of $A$ given $(1,X)$, $\alpha$ is the coefficient of $A$ in $\proj(\Lb\mid1,X,A)$, and $H(P_z)$ is the second-moment matrix of $\proj(\Lb\mid\Zt^\circ)$, with $\Zt^\circ$ the residual of $P_z^\top R_z$ on $(1,X,A)$. The treatment-side factor is minimized by monotonicity of projected variance
\[
H(P_z)\preceq H(I)\quad\text{for every }P_z,\qquad\text{with equality if }\col P_z\supseteq\col\big(\Sigma_{zz}^{-1}\Lambda_z\big).
\]
The outcome-side factor is minimized by choosing, among the weights that reproduce the outcome coefficients, the one with the smallest noise variance
\[
\min_{\vartheta:\ \Lambda_w^\top\vartheta=\beta}\ \vartheta^\top\Psi_w\vartheta=\beta^\top\big(\Lambda_w^\top\Psi_w^{-1}\Lambda_w\big)^{-1}\beta,
\]
attained at $\vartheta^\ast\in\col(\Psi_w^{-1}\Lambda_w)$, and $\col(\Psi_w^{-1}\Lambda_w)=\col(\Sigma_{ww}^{-1}\Lambda_w)$ by the Woodbury identity. By (a), both minimizers are the CCA directions, and neither depends on $(\tau,\beta)$. Hence CCA attains, for every outcome coefficient,
\[
\mathcal V^\ast=\Big(\sigma_e^2+\beta^\top\big(\Lambda_w^\top\Psi_w^{-1}\Lambda_w\big)^{-1}\beta\Big)\Big(\frac1{s_A}+\alpha^\top H(I)^{-1}\alpha\Big).
\]

(c) For feasible $\vartheta$, the cross term vanishes at the minimizer, so the excess outcome-side variance is a quadratic form,
\[
\vartheta^\top\Psi_w\vartheta-\vartheta^{\ast\top}\Psi_w\vartheta^\ast=(\vartheta-\vartheta^\ast)^\top\Psi_w(\vartheta-\vartheta^\ast),
\]
and $H(I)-H(P_z)$ is the second moment of a projection residual. Both are therefore $O(\theta^2)$ in the distance $\theta$ between the subspaces used and those of CCA. A ridge moves the directions by $O(\lambda)$ and a pool of size $m$ by $O_p(m^{-1/2})$ by Wedin's theorem \citep{wedin1972}, so $\mathcal V-\mathcal V^\ast=O_p\big((\lambda+m^{-1/2})^2\big)$.
\end{proofA}

\begin{proofA}[Proof of \cref{thm:nonlinear}]\label{prf:nonlinear}
Fix $x$ and suppress it. Define the conditional-expectation operators
\[
\mathcal S_z\phi=\E[\phi(L)\mid M_z],\qquad \mathcal S_w\phi=\E[\phi(L)\mid M_w],\qquad \mathcal Tg=\E[g(M_w)\mid M_z],
\]
whose adjoints are $\mathcal S_z^\ast f=\E[f(M_z)\mid L]$, $\mathcal S_w^\ast g=\E[g(M_w)\mid L]$ and $\mathcal T^\ast f=\E[f(M_z)\mid M_w]$. We write $\mathcal T$ also for its restriction to mean-zero functions.

\emph{Factorization.} (P) implies $M_w\perp M_z\mid L$, so by the tower property
\[
\mathcal Tg=\E\big[\E[g(M_w)\mid M_z,L]\mid M_z\big]=\E\big[\E[g(M_w)\mid L]\mid M_z\big]=\mathcal S_z\mathcal S_w^\ast g .
\]
The singular system satisfies $\mathcal Tg_j=\sigma_jf_j$ and $\mathcal T^\ast f_j=\sigma_jg_j$, so every canonical pair consists of posterior means
\[
f_j(M_z)=\E[\phi_j(L)\mid M_z],\quad \phi_j=\sigma_j^{-1}\mathcal S_w^\ast g_j,\qquad\qquad
g_j(M_w)=\E[\psi_j(L)\mid M_w],\quad \psi_j=\sigma_j^{-1}\mathcal S_z^\ast f_j .
\]

\emph{Style.} Let $M_z=h(M_z^{\rm c},S)$ for an injective map $h$, where the style $S$ is independent of $(M_z^{\rm c},M_w)$ jointly. Then $\E[g(M_w)\mid M_z^{\rm c},S]=\E[g(M_w)\mid M_z^{\rm c}]$, so $\mathcal T^\ast\mathcal T$ does not change with $S$, and neither do the $\sigma_j$, the $g_j$, or the $f_j$ as functions of the content.

\emph{Contrastive loss.} Let $\mathcal T$ be Hilbert--Schmidt. For heads $u=(u_j)_{j\le d}$ and $v=(v_j)_{j\le d}$ of mean-zero functions and critic $s=u(M_z)^\top v(M_w)$, the negative is drawn independently of $M_z$ given $x$, so
\begin{align*}
\mathcal L(u,v)&=-2\sum_{j\le d}\langle u_j,\mathcal Tv_j\rangle+\sum_{i,j\le d}\langle u_i,u_j\rangle\langle v_i,v_j\rangle\\
&=\Big\lVert\mathcal T-\sum_{j\le d}u_j\otimes v_j\Big\rVert_{\rm HS}^2-\lVert\mathcal T\rVert_{\rm HS}^2 .
\end{align*}
Schmidt's theorem \citep{schmidt1907} gives the minimizers $u_j=\sigma_j^{1/2}f_j$ and $v_j=\sigma_j^{1/2}g_j$, up to an invertible linear map.

\emph{Growing dictionaries.} Let $\Pi_K^z$ and $\Pi_K^w$ project onto the spans of nested dictionaries whose unions are dense. Under (A1)--(A2), linear CCA without ridge of the $K$-th dictionaries computes the singular values of $\Pi_K^z\mathcal T\Pi_K^w$, and since $\mathcal T$ is compact,
\[
\big\lVert\Pi_K^z\mathcal T\Pi_K^w-\mathcal T\big\rVert\le\big\lVert(\Pi_K^z-I)\mathcal T\big\rVert+\big\lVert(\Pi_K^w-I)\mathcal T^\ast\big\rVert\ \longrightarrow\ 0 .
\]
Weyl's inequality and the Davis--Kahan theorem \citep{stewart1990,davis1970} then give convergence of the canonical correlations and subspaces.
\end{proofA}

\begin{proofA}[Proof of \cref{prop:complete}]\label{prf:complete}
Fix $(a,x)$ and write all expectations given $(A,X)=(a,x)$.

(a) Suppose $\E[g(L)\mid M]=0$. Since $S$ is a function of $M$, the tower property gives
\[
\E[g(L)\mid S]=\E\big[\E[g(L)\mid M]\mid S\big]=0,
\]
and completeness of $S$ gives $g=0$.

(b) If $L\perp M\mid S$, then $\E[g(L)\mid M]=\E[g(L)\mid S]$, so the conditions $\E[g(L)\mid M]=0$ and $\E[g(L)\mid S]=0$ coincide.

(c)(i) Under (F), CPF's outcome-side proxy is linear in the factor
\[
\Wt=\Lambda_{\Wt}\Lb+c(X)+\xi,\qquad \Lambda_{\Wt}=\mathrm{diag}(\sigma)^{-1}\hat\Theta_w^\top\Lambda_w,\qquad \xi=\mathrm{diag}(\sigma)^{-1}\hat\Theta_w^\top\varepsilon_w,
\]
where $\sigma$ holds the pool standard deviations of the variates. If the $d\ge k$ directions contain the top $k$ ridge directions, then by the proof of \cref{thm:efficiency}(a) the corresponding $k$ rows of $\Lambda_{\Wt}$ form, up to row scaling, the matrix $N^\top\Lambda_w^\top(\Sigma_{ww}+\lambda I)^{-1}\Lambda_w$ with $N$ invertible, which is invertible for every $\lambda\ge0$. Hence $\operatorname{rank}\Lambda_{\Wt}=k$. By (F) and (P), $\xi\perp(L,A)\mid X$, and \cref{lem:complete} gives completeness.

If $d<k$, then $\Wt$ depends on $\Lb$ only through $\Lambda_{\Wt}\Lb$, which does not determine $\Lb$. Take any $h$ that is not a function of $(\Lambda_{\Wt}\Lb,A,X)$ and set
\[
g(L)=h(L)-\E\big[h(L)\mid\Lambda_{\Wt}\Lb,A,X\big]\neq0 .
\]
Then
\begin{align*}
\E\big[g(L)\mid\Wt,A,X\big]&=\E\Big[\E\big[g(L)\mid\Lambda_{\Wt}\Lb,\xi,A,X\big]\,\Big|\,\Wt,A,X\Big]\\
&=\E\Big[\E\big[g(L)\mid\Lambda_{\Wt}\Lb,A,X\big]\,\Big|\,\Wt,A,X\Big]=0,
\end{align*}
where the first equality is the tower property, since $\Wt$ is a function of $(\Lambda_{\Wt}\Lb,\xi,X)$, and the second uses that $\xi$ is independent of $L$ given $(\Lambda_{\Wt}\Lb,A,X)$. Completeness therefore fails. On the treatment side the same argument needs $\varepsilon_z\perp A\mid L,X$, which holds when $M_z\perp A\mid L,X$.

(c)(ii) If $L$ takes $K$ values $l_1,\dots,l_K$ and both modalities are complete, the cross-modal operator has rank $K-1$, and its top $K-1$ canonical functions are an invertible affine map of the posterior probabilities (\cref{s:sec:thm4nl}, Theorem~4(g))
\[
\big(g_1,\dots,g_{K-1}\big)(M_w)=\mathcal A\Big(\big(P(L=l_j\mid M_w,X)\big)_{j<K}\Big),\qquad \mathcal A\ \text{invertible affine}.
\]
The posterior of $L$ given $M_w$ is therefore a function of $S=(g_1,\dots,g_d)(M_w)$ for any $d\ge K-1$. Because $M_w\perp A\mid L,X$ by (P), Bayes' rule carries this over to $(A,X)$
\[
P(L=l\mid M_w,A,X)=\frac{P(A\mid L=l,X)\,P(L=l\mid M_w,X)}{\sum_{j\le K}P(A\mid L=l_j,X)\,P(L=l_j\mid M_w,X)},
\]
which is a function of $(S,A,X)$. Hence $L\perp M_w\mid S,A,X$, and (b) applies. The treatment side is the same once $M_z\perp A\mid L,X$.

(d) Write $\mathcal T=\mathcal S_z\mathcal S_w^\ast$ as in the proof of \cref{thm:nonlinear}. Completeness of $M_z$ makes $\mathcal S_z$ injective. Completeness of $M_w$ makes $\mathcal S_w$ injective, so $\operatorname{range}\mathcal S_w^\ast$ is dense in $L^2(L)$. If $\mathcal T$ has finite rank $r$, then, counting constants,
\[
\dim\operatorname{range}\mathcal S_w^\ast=\dim\mathcal S_z\big(\operatorname{range}\mathcal S_w^\ast\big)=\dim\operatorname{range}\mathcal T\le r+1,
\]
where the first equality uses injectivity of $\mathcal S_z$. A finite-dimensional dense subspace is the whole space, so $\dim L^2(L)\le r+1$. $L$ has at most $r+1$ atoms. For a continuous $L$, the full sequence of canonical functions generates the posterior (\cref{s:sec:thm4nl}, Theorem~4(e)), and a finite set does so when it is sufficient, as in (i). In general, the $\sigma$-fields $\mathcal F_d=\sigma\big(g_1(M_w),\dots,g_d(M_w)\big)$ increase with $d$, and by martingale convergence
\[
\E\big[\phi(L)\mid\mathcal F_d\big]\ \longrightarrow\ \E\big[\phi(L)\mid\mathcal F_\infty\big]\quad\text{in }L^2\ \text{as }d\to\infty,
\]
where $\mathcal F_\infty$ carries the same information about $L$ as the raw modality (\cref{s:sec:thm4nl}, Theorem~4(e)). \hfill$\square$
\end{proofA}

\section{Further theoretical results}\label{app:further}

The results below support the main text. Their complete proofs are in \cref{app:proofs}, and \cref{lem:complete}, which \cref{prop:complete} rests on, is also proved here.

\begin{theorem}[Dictionary monotonicity]\label{thm:monotone}
Let $D$ be a compression of $D'$, that is, $D_z=T_zD'_z$ and $D_w=T_wD'_w$ for fixed matrices $T_z,T_w$ (a column subset is a row selection of the identity). Let $\mathcal I_z(D)=\E\big[\proj(\Lb\mid1,X,D_z)\,\proj(\Lb\mid1,X,D_z)^\top\big]$ be the linear confounding information.
\begin{enumerate}
\item[(a)] $\rho_j(D)\le\rho_j(D')$ for every $j$. For column subsets, also $\rho_j(\lambda;D)\le\rho_j(\lambda;D')$ for every $\lambda>0$.
\item[(b)] The optimal contrastive loss is no larger on $D'$, and $\mathcal I_z(D)\preceq\mathcal I_z(D')$. If (F) holds for $D'$, the $k$ CCA proxies of $D'$ attain $\mathcal I_z(D')$.
\item[(c)] Under the linear proximal model for $D'$, $\mathcal V^\ast(D')\le\mathcal V^\ast(D)$, and CPF on $D'$ attains $\mathcal V^\ast(D')$.
\item[(d)] PCA has no such property. Adding a single column can drop the relevance of PCA-1 from $R^2=0.5$ to $0$, while the canonical correlation stays at $0.5$.
\end{enumerate}
\end{theorem}

\begin{proposition}[Random-phase waveforms]\label{prop:phase}
Let a waveform of length $N$ be
\[
x_t=\mu(L,X)\sin\!\big(2\pi f t/N+\varphi\big)+s_t(L,X)+\xi_t,\qquad t=0,\dots,N-1,
\]
where the frequency is $f=f(L,X)$, $\E[\xi_t\mid L,X]=0$, and the phase $\varphi\sim{\rm Unif}[0,2\pi)$ is independent of everything else.
\begin{enumerate}
\item[(a)] $\E[x_t\mid L,X]=s_t(L,X)$ for every $t$. Hence, for every fixed $c\in\R^N$ and every $g$,
\[
\Cov\big(c^\top x,\,g(L,X)\big)=\Cov\big(c^\top s,\,g(L,X)\big),
\]
so no linear function of the raw samples sees the amplitude $\mu$ or the frequency $f$.
\item[(b)] Let $D_N(\nu)=\sum_{t<N}e^{2\pi i\nu t/N}$ and let $F_k=\big|\sum_{t<N}(x_t-\bar x)\,e^{-2\pi i(k/q)t/N}\big|$ be the magnitude spectrum with zero-padding factor $q$. For the oscillatory component and every phase,
\[
\Big|F_k-\tfrac{|\mu|}{2}\big|D_N(f-k/q)\big|\Big|\le\tfrac{|\mu|}{2}\big|D_N(f+k/q)\big|+\tfrac{|\mu|}{N}\big|D_N(f)\big|\,\big|D_N(k/q)\big| ,
\]
whose main term depends on $(L,X)$ only.
\end{enumerate}
\end{proposition}
Numerically, the out-of-sample $R^2$ for $U$ was $.97$ with 48 spectral bins and at most $.00$ with 64 raw samples.

\begin{theorem}[Finite pool and label-free selection]\label{thm:selection}
Let the pool have $m$ rows, let $r_z,r_w$ be the residuals of its dictionaries on $[1,X]$ with second moments $S_{zz},S_{ww},S_{zw}$, and let $m'=m-\operatorname{rank}[1,X]$. For each candidate $c=(B,\lambda)$ in the grid $\mathcal G$ of \cref{alg:cpf}, let $r_{cj}$ and $\hat r_{cj}$ be the population and held-out correlations of the $j$-th pair of variates fitted on the training split, $J(c)=\sum_{j\le d}r_{cj}$, and $\hat c=\arg\max_c\sum_j\hat r_{cj}$, with $m_{\rm va}$ held-out rows.
\begin{enumerate}
\item[(a)] If $\operatorname{rank}r_z+\operatorname{rank}r_w>m'$, at least $\operatorname{rank}r_z+\operatorname{rank}r_w-m'$ unregularized sample canonical correlations equal $1$, under every law, independence included.
\item[(b)] Every $\hat\rho_j(\lambda)$ is non-increasing in $\lambda$, and
\[
\hat\rho_1(\lambda)\le\frac{\lVert S_{zw}\rVert}{\sqrt{\big(\lambda_{\min}(S_{zz})+\lambda\big)\big(\lambda_{\min}(S_{ww})+\lambda\big)}}\le\frac{\lVert S_{zw}\rVert}{\lambda}.
\]
\item[(c)] $\hat c$ and the final maps are functions of the pool and the seeds, so \cref{thm:exclusion} holds whatever the selection chose.
\item[(d)] Conditionally on the training split, with $\eta=\max_{c,j}|\hat r_{cj}-r_{cj}|$,
\[
J(\hat c)\ge\max_{c\in\mathcal G}J(c)-2d\eta,\qquad
\eta\le c_0\kappa^2\Big(\sqrt{\ell/m_{\rm va}}+\ell/m_{\rm va}\Big)\ \ \text{with probability at least }1-\delta,
\]
where $\ell=\log(10|\mathcal G|d/\delta)$, $\kappa$ bounds the sub-Gaussian norms of the standardized variates, and $c_0$ is an absolute constant.
\end{enumerate}
\end{theorem}

\begin{lemma}[Completeness without Gaussianity]\label{lem:complete}
Suppose
\[
\Wt=\Lambda_{\Wt}\Lb+c(X)+\xi,\qquad \operatorname{rank}\Lambda_{\Wt}=k,\qquad \xi\perp(L,A)\mid X,
\]
and that, for a left inverse $G$ of $\Lambda_{\Wt}$ ($G\Lambda_{\Wt}=I_k$), the characteristic function of $G\xi$ given $X=x$ is nonzero on a dense subset of $\R^k$ for almost every $x$. Then $\Wt$ is complete for $L$. $\E[g(L)\mid\Wt,A,X]=0$ implies $g(L)=0$, whatever the distributions of $L$ and $\xi$.
\end{lemma}

\begin{proofA}[Proof of \cref{lem:complete}]\label{prf:lemcomplete}
Fix $(a,x)$. Since $G\Wt$ is a function of $\Wt$, the hypothesis gives $\E\big[g(L)\,1\{A=a\}\,\phi(G\Wt)\mid X=x\big]=0$ for every bounded $\phi$. Now $G\Wt=\Lb+Gc(x)+\eta$ with $\eta=G\xi\perp(L,A)\mid X$, so taking $\phi(y)=e^{it^\top y}$ gives
\[
0=e^{it^\top(Gc(x)-\E[L\mid x])}\,\varphi_\eta(t)\,\hat\mu_g(t),\qquad
\hat\mu_g(t)=\int e^{it^\top l}\,g(l)\,\pi_a(l,x)\,P_{L\mid X=x}(dl),
\]
where $\pi_a(l,x)=P(A=a\mid L=l,X=x)$ and $\varphi_\eta$ is the characteristic function of $\eta$. Hence $\hat\mu_g=0$ on the dense set where $\varphi_\eta\neq0$, and everywhere by continuity. A finite signed measure whose Fourier transform vanishes is zero, so $g\,\pi_a=0$ almost everywhere, and $g(L)=0$ because $\pi_a>0$ by positivity.
\end{proofA}
The condition cannot be dropped. With Fej\'er noise, whose characteristic function is $(1-|t|)_+$, and $W=L+\xi$, one has $\E[\cos3L\mid W]=0$ although $\cos3L\neq0$ (\cref{s:sec:lem1}).

\begin{proposition}[Why bridges are needed]\label{prop:bridge}
Let the confounder $L$ be scalar and
\[
A=\pi L+\kappa_A^\top X+\nu,\qquad \Zt=aL+\delta_z,\qquad \Wt=bL+\delta_w,\qquad Y=\tau A+\beta L+\gamma^\top X+e,
\]
with $a,b\neq0$ and $\nu,\delta_z,\delta_w,e$ mutually independent, independent of $(L,X)$ and of mean zero. Gaussianity is not needed.
\begin{enumerate}
\item[(a)] Least squares of $Y$ on $(1,A,X,\Zt,\Wt)$, which adjusts for the proxies as covariates, converges to
\[
\tau+\beta\,\frac{\Cov(\check A,\check L)}{\Var\check A}=\tau+\frac{\beta\pi\Var\check L}{\pi^2\Var\check L+\sigma_\nu^2},
\]
where checks denote residuals on $(1,X,\Zt,\Wt)$. The bias is nonzero whenever $\beta\pi\neq0$ and the proxies do not determine $L$ ($\Var\check L>0$).
\item[(b)] 2SLS of $Y$ on $(1,A,X,\Wt)$ with instruments $(1,A,X,\Zt)$ is consistent for $\tau$ for every noise level $\sigma_z^2,\sigma_w^2$, provided $ab\Var(L^\circ)\neq0$, where $L^\circ=L-\proj(L\mid1,A,X)$.
\end{enumerate}
\end{proposition}

\begin{corollary}[$X$-conditioning]\label{cor:xcond}
Let $r$ be the rank of the part of the modalities that depends on $X$.
\begin{enumerate}
\item[(a)] $X$-conditional CCA with $d\ge k$ attains the bound of \cref{thm:efficiency}(a),
\[
\proj(\Lb\mid1,X,D_z)\in\operatorname{span}(1,X,\Zt),
\]
however $X$ enters the modalities.
\item[(b)] Unconditional CCA attains it as well if that part is visible in both modalities and $d\ge k+r$.
\item[(c)] Otherwise unconditional CCA can fall strictly below the bound. If that part is invisible, it does so for every $d$.
\end{enumerate}
\end{corollary}

\paragraph{Incomplete proxies bias the estimate, not only its variance.} Let the confounder have independent components $U_1,U_2$ in the linear proximal model
\[
Y=\tau A+b_1^\top U_1+b_2^\top U_2+\varepsilon .
\]
Completeness is then the rank condition that the proxies' loadings on $U=(U_1,U_2)$ have full column rank. If both proxies load on $U_1$ only, the condition fails, and proximal 2SLS converges to
\[
\hat\tau_{\rm 2SLS}\ \xrightarrow{\ p\ }\ \tau+\frac{b_2^\top\Cov(U_2,A)}{\Var(A\mid U_1)},
\]
which is the limit of the regression of $Y$ on $(A,U_1)$. The estimator adjusts for the part of the confounder the proxies carry and leaves the rest as omitted-variable bias. With $n=2\times10^6$ and $\tau=1$.
\begin{center}
\small
\begin{tabular}{lc}
\toprule
& Estimate of $\tau=1$ \\
\midrule
Proximal 2SLS, proxies load on $U_1$ only & 1.747 \\
Formula above & 1.746 \\
Regression of $Y$ on $(A,U_1)$ & 1.747 \\
Proximal 2SLS, proxies load on $U_1$ and $U_2$ & 1.001 \\
\bottomrule
\end{tabular}
\end{center}
PCA proxies under \cref{prop:pca} are the extreme case, in which no part of $\Lb$ is carried.

\paragraph{Outcome factors present only on the outcome side.} If a factor $Q$ that affects the outcome enters only the outcome-side modality, (P) holds with respect to $L$ and CPF remains valid, but the outcome coefficients of $Q$ do not appear in the unlabeled distribution, so no label-free compression rule can be efficient for every law sharing that distribution. CPF's rule is efficient exactly when the covariance between the outcome and the outcome-side noise lies in $\col\Lambda_w$.

\clearpage
\section{Synthetic data-generating model}\label{app:dgp}
\cref{tab:dgp} summarizes the data-generating model, the cases, the settings and the sample sizes. The base structure is that of NC-CSF \citep{nccsf}, with the tabular proxies replaced by token sequences and waveforms. The three sweeps (style multiplier 0--16 on D; unlabeled pool 500--10{,}000 on S8; proxy dimension 2--32 on S8 and D) change one quantity at a time and were declared before they were run.

\begin{table}[H]
\centering\footnotesize
\caption{Synthetic data-generating model, cases, and evaluation settings. $\sigma(\cdot)$ is the logistic function and $B_\cdot$ are smooth basis functions over the length of a modality.}
\label{tab:dgp}
\renewcommand{\arraystretch}{1.04}
\begin{tabular}{@{}>{\raggedright\arraybackslash}p{0.17\textwidth}p{0.79\textwidth}@{}}
\toprule
\multicolumn{2}{@{}l}{\textbf{Latent variables and covariates}}\\
Covariates & $X\sim N(0,I_5)$\\
Confounder & $U\sim N(0,1)$\\
Shared factor & $V\sim N(0,1)$; enters both modalities only, unrelated to treatment, outcome, and censoring\\
\midrule
\multicolumn{2}{@{}l}{\textbf{Treatment, survival, censoring}}\\
Treatment & Linear: $\mathrm{logit}\,P(A=1)=b_0+X^\top\alpha+\gamma_uU$; nonlinear: $b_0+\sum_k\alpha'_ks_k(X)+\gamma_u\sigma(U)$, $s(X)=(\sigma(X_1),\sigma(X_2),\sigma(X_3),X_1X_2,X_2X_3)$; $\alpha\sim N(0,0.5^2I)$; $b_0$ calibrated to $P(A=1)=0.5$\\
Event time & Weibull proportional hazards (shape 1.5, scale 0.4); $\eta_0=X^\top\beta+\beta_uU$ (nonlinear outcome adds $X_1^2$, $X_2^2$, $X_1U$, $0.5\sigma(U)$), $\eta_1=\eta_0+\tau_{\log HR}$, $\beta\sim N(0,0.4^2I)$; the two potential outcomes share the same uniform draw\\
Censoring & Weibull (shape 1.2); $\eta_c=X^\top\beta_c+0.3U$, $\beta_c\sim N(0,0.3^2I)$; scale calibrated to \pct{35} censoring\\
Truth & Unit-level effects $\tau_{\rm RMST}$, $\tau_{\rm SP}$ given $(X,U)$, computed in closed form at $t_0=0.75$\\
\midrule
\multicolumn{2}{@{}l}{\textbf{Modalities}}\\
Treatment side (notes) & Continuous signal of length 64, $f(U)^\top B_u+0.7\,(X^\top c_z)B_x+V B_v$ (+ style) + noise, $f(U)=(\tanh U,\,(U^2-1)/\sqrt2,\,\sin(1.5U)/0.7)$; the noisy signal is tokenized into 32 global quantile bins\\
Outcome side (ECG) & Length 500, $t\in[0,1]$: $(1+0.6\tanh U)\sin(2\pi(3+2\sigma(U))t+\varphi)+0.8\exp\{-((t-0.3-0.15\tanh U)/0.05)^2\}$ $+\,0.7(X^\top c_w)(t-0.5)+V\exp\{-((t-0.7)/0.08)^2\}$ (+ style) + noise, $\varphi\sim{\rm Unif}[0,2\pi)$\\
Element-wise SNR & Variance of the $U$ component / noise variance $=(1.5/\sigma)^2$: informative cases 1.78 (treatment side) and 0.96 (outcome side); noisy cases 0.042 and 0.099\\
Base embedding & Per-modality 64-dimensional autoencoder (reconstruction loss, 20 epochs), trained on the pool only and then frozen\\
\midrule
\multicolumn{2}{@{}l}{\textbf{Cases} \;(treatment/outcome model; proxies; $\beta_u$ = $U\to T$, $\gamma_u$ = $U\to A$, $\tau_{\log HR}$)}\\
1, 3 & Linear/linear; informative (0.8, 0.8, $-0.6$), noisy (0.9, 0.9, $-0.5$)\\
5, 7 & Linear/nonlinear; informative (0.8, 0.9, $-0.7$), noisy (1.1, 0.9, $-0.5$)\\
9, 11 & Nonlinear/nonlinear; informative (1.0, 1.0, $-0.7$), noisy (1.1, 0.9, $-0.5$)\\
2, 10 (S8, D only) & Weak confounding, informative: linear/linear (0.3, 0.2, $-0.25$), nonlinear/nonlinear (0.2, 0.2, $-0.3$)\\
\midrule
\multicolumn{2}{@{}l}{\textbf{Group A: modality-specific variation} \;(3 style factors per modality, $N(0,1)$, smooth bases,}\\
\multicolumn{2}{@{}l}{\quad independent across modalities; element-wise variance = multiplier $\times$ variance of the $U$ component)}\\
S8 & Style $\times8$\\
HT3S & $t_3$ noise (matched to the Gaussian interquartile range) + style $\times8$\\
HBIS & Binarization (2-symbol tokens, 1-bit waveform at the median) + style $\times2$\\
HALL & $t_3$ noise + SNR $\times0.25$ + style $\times8$ + $V\times2$\\
\midrule
\multicolumn{2}{@{}l}{\textbf{Group B: base setting, survival realism, and generality}}\\
D & Base setting: no style, SNR and loadings as above\\
HV & $V$ loading $\times3$ (the strongest shared signal is a non-confounder)\\
HLOW & SNR $\times0.05$\\
RARE & Event-time scale $\times10$ (\pct{10}--\pct{25} events before the horizon)\\
EMOD & Treated-arm log hazard ratio $+0.4\,U$ (effect modification by the confounder)\\
MULTI & Treatment side = notes + medication tokens (length 32, vocabulary 16); outcome side = waveform + lab panel (12 values)\\
MISS & \pct{40} of outcome-side modality missing completely at random (zero-filled + missingness indicator)\\
\midrule
\multicolumn{2}{@{}l}{\textbf{Samples and repetitions}}\\
Sizes & Unlabeled pool 10{,}000, training 3{,}000, test 20{,}000 (independent samples from the same population)\\
Datasets & 6 strong-confounding cases $\times$ 10 repetitions = 60 per setting (80 for S8 and D, which add the 2 weak-confounding cases); 260 in group A, 440 in group B\\
\bottomrule
\end{tabular}
\end{table}

\section{Additional synthetic results}\label{app:results}
\paragraph{Group results with the oracle gap.} \cref{tab:summary} reports both targets and three metrics for CPF, NC-CSF with PCA proxies and the oracle; \cref{tab:delta} gives the change by setting, and Tables~\ref{tab:fullA}--\ref{tab:fullB} the values.

\begin{table}[h]
\centering\small
\caption{CATE error (group means) of CPF, NC-CSF (the same estimator with PCA proxies) and the oracle (NC-CSF with the true $U$). In parentheses: share of the gap between $X$-only adjustment (CSF-X) and the oracle closed (\%). All CPF--NC-CSF differences are significant ($|z|\ge2.0$).}
\label{tab:summary}
\setlength{\tabcolsep}{3pt}
\begin{tabular}{@{}ll ccc ccc@{}}
\toprule
& & \multicolumn{3}{c}{Group A (strong style)} & \multicolumn{3}{c}{Group B (base and survival realism)}\\
\cmidrule(lr){3-5}\cmidrule(l){6-8}
Target & Method & RMSE & MAE & \Bias & RMSE & MAE & \Bias\\
\midrule
\multirow{3}{*}{RMST} & Oracle & .0232 & .0186 & .0076 & .0226 & .0177 & .0069\\
 & \textbf{CPF} & \textbf{.0374} (75.0) & \textbf{.0322} (75.5) & \textbf{.0221} (76.5) & \textbf{.0347} (78.1) & \textbf{.0287} (79.2) & \textbf{.0179} (81.3)\\
 & NC-CSF & .0452 (61.2) & .0398 (61.6) & .0320 (60.6) & .0356 (76.4) & .0295 (77.6) & .0191 (79.2)\\
\midrule
\multirow{3}{*}{Surv.\ prob.} & Oracle & .0472 & .0352 & .0161 & .0473 & .0347 & .0142\\
 & \textbf{CPF} & \textbf{.0703} (71.6) & \textbf{.0566} (70.6) & \textbf{.0331} (76.7) & \textbf{.0665} (77.0) & \textbf{.0520} (76.6) & \textbf{.0279} (81.8)\\
 & NC-CSF & .0804 (59.2) & .0652 (58.9) & .0433 (62.7) & .0679 (75.3) & .0532 (75.0) & .0286 (80.8)\\
\bottomrule
\end{tabular}
\end{table}

\begin{table}[h]
\centering\small
\caption{Change in CPF's error relative to NC-CSF with PCA proxies, by setting (\%). $^\dagger$: $|z|<2$.}
\label{tab:delta}
\setlength{\tabcolsep}{3.2pt}
\begin{tabular}{@{}l rrr rrr@{}}
\toprule
& \multicolumn{3}{c}{RMST} & \multicolumn{3}{c}{Survival probability}\\
\cmidrule(lr){2-4}\cmidrule(l){5-7}
Setting & RMSE & MAE & \Bias & RMSE & MAE & \Bias\\
\midrule
S8    & $-20.9$ & $-23.5$ & $-42.3$ & $-14.8$ & $-15.8$ & $-29.8$\\
HT3S  & $-22.0$ & $-24.2$ & $-37.1$ & $-14.8$ & $-15.9$ & $-25.6$\\
HBIS  & $-14.5$ & $-16.2$ & $-28.9$ & $-10.7$ & $-11.2$ & $-23.5$\\
HALL  & $-12.5$ & $-13.7$ & $-19.7$ & $-9.9$  & $-9.9$  & $-17.0$\\
\midrule
D     & $-2.0$\ns & $-3.0$    & $-5.3$\ns & $+0.1$\ns & $-0.4$\ns & $+3.9$\ns\\
HV    & $-1.8$\ns & $-1.8$\ns & $-9.5$\ns & $-1.8$\ns & $-1.9$\ns & $-2.5$\ns\\
HLOW  & $-1.2$\ns & $-1.1$\ns & $-2.0$\ns & $-1.2$\ns & $-0.9$\ns & $-0.8$\ns\\
RARE  & $-5.5$    & $-3.7$    & $+0.4$\ns & $-4.6$    & $-3.7$    & $-3.2$\ns\\
EMOD  & $-1.0$\ns & $-1.0$\ns & $-5.1$\ns & $-1.9$\ns & $-2.6$    & $-0.1$\ns\\
MULTI & $-4.7$    & $-7.3$    & $-20.0$   & $-4.1$    & $-5.6$    & $-13.2$\\
MISS  & $-4.5$    & $-4.2$    & $-8.1$    & $-2.9$    & $-2.7$    & $-4.6$\ns\\
\bottomrule
\end{tabular}
\end{table}

\begin{table}[H]
\centering\small
\caption{CATE error in group A (260 datasets; 80 for S8 and 60 for each other setting). $|$Bias$|$ is the absolute value of the per-dataset mean bias $\mathrm{mean}(\hat\tau-\tau)$, averaged over datasets, and $\Delta$ is the relative change of CPF versus NC-CSF. The paired difference is significant in every cell ($|z|\ge5.1$).}
\label{tab:fullA}
\setlength{\tabcolsep}{4.5pt}
\begin{tabular}{@{}l rrr rrr rrr@{}}
\toprule
& \multicolumn{3}{c}{RMSE} & \multicolumn{3}{c}{MAE} & \multicolumn{3}{c}{$|$Bias$|$}\\
\cmidrule(lr){2-4}\cmidrule(lr){5-7}\cmidrule(l){8-10}
Setting & \textbf{CPF} & NC-CSF & $\Delta$ & \textbf{CPF} & NC-CSF & $\Delta$ & \textbf{CPF} & NC-CSF & $\Delta$\\
\midrule
\multicolumn{10}{@{}l}{\emph{RMST}}\\
S8   & \textbf{.0294} & .0372 & $-20.9\%$ & \textbf{.0245} & .0321 & $-23.5\%$ & \textbf{.0142} & .0245 & $-42.3\%$\\
HT3S & \textbf{.0371} & .0475 & $-22.0\%$ & \textbf{.0318} & .0420 & $-24.2\%$ & \textbf{.0215} & .0342 & $-37.1\%$\\
HBIS & \textbf{.0337} & .0394 & $-14.5\%$ & \textbf{.0281} & .0335 & $-16.2\%$ & \textbf{.0172} & .0242 & $-28.9\%$\\
HALL & \textbf{.0520} & .0594 & $-12.5\%$ & \textbf{.0467} & .0541 & $-13.7\%$ & \textbf{.0381} & .0475 & $-19.7\%$\\
Overall & \textbf{.0374} & .0452 & $-17.3\%$ & \textbf{.0322} & .0398 & $-19.2\%$ & \textbf{.0221} & .0320 & $-30.9\%$\\
\midrule
\multicolumn{10}{@{}l}{\emph{Survival probability}}\\
S8   & \textbf{.0573} & .0673 & $-14.8\%$ & \textbf{.0453} & .0538 & $-15.8\%$ & \textbf{.0242} & .0345 & $-29.8\%$\\
HT3S & \textbf{.0737} & .0865 & $-14.8\%$ & \textbf{.0589} & .0700 & $-15.9\%$ & \textbf{.0360} & .0484 & $-25.6\%$\\
HBIS & \textbf{.0660} & .0739 & $-10.7\%$ & \textbf{.0527} & .0593 & $-11.2\%$ & \textbf{.0254} & .0332 & $-23.5\%$\\
HALL & \textbf{.0884} & .0981 & $-9.9\%$  & \textbf{.0734} & .0815 & $-9.9\%$  & \textbf{.0499} & .0601 & $-17.0\%$\\
Overall & \textbf{.0703} & .0804 & $-12.5\%$ & \textbf{.0566} & .0652 & $-13.2\%$ & \textbf{.0331} & .0433 & $-23.5\%$\\
\bottomrule
\end{tabular}
\end{table}

\begin{table}[H]
\centering\small
\caption{CATE error in group B (80 datasets for D, which adds the weak-confounding cases, and 60 for each other setting; 440 in total). Metrics and $\Delta$ are defined as in Table~\ref{tab:fullA}.}
\label{tab:fullB}
\setlength{\tabcolsep}{4.5pt}
\begin{tabular}{@{}l rrr rrr rrr@{}}
\toprule
& \multicolumn{3}{c}{RMSE} & \multicolumn{3}{c}{MAE} & \multicolumn{3}{c}{$|$Bias$|$}\\
\cmidrule(lr){2-4}\cmidrule(lr){5-7}\cmidrule(l){8-10}
Setting & \textbf{CPF} & NC-CSF & $\Delta$ & \textbf{CPF} & NC-CSF & $\Delta$ & \textbf{CPF} & NC-CSF & $\Delta$\\
\midrule
\multicolumn{10}{@{}l}{\emph{RMST}}\\
D     & \textbf{.0264} & .0270 & $-2.0\%$ & \textbf{.0217} & .0224 & $-3.0\%$ & \textbf{.0116} & .0122 & $-5.3\%$\\
HV    & \textbf{.0311} & .0317 & $-1.8\%$ & \textbf{.0255} & .0260 & $-1.8\%$ & \textbf{.0134} & .0148 & $-9.5\%$\\
HLOW  & \textbf{.0616} & .0623 & $-1.2\%$ & \textbf{.0567} & .0573 & $-1.1\%$ & \textbf{.0486} & .0496 & $-2.0\%$\\
RARE  & \textbf{.0172} & .0182 & $-5.5\%$ & \textbf{.0112} & .0117 & $-3.7\%$ & .0053 & .0053 & $+0.4\%$\\
EMOD  & \textbf{.0385} & .0389 & $-1.0\%$ & \textbf{.0292} & .0294 & $-1.0\%$ & \textbf{.0120} & .0126 & $-5.1\%$\\
MULTI & \textbf{.0300} & .0315 & $-4.7\%$ & \textbf{.0244} & .0263 & $-7.3\%$ & \textbf{.0115} & .0143 & $-20.0\%$\\
MISS  & \textbf{.0408} & .0427 & $-4.5\%$ & \textbf{.0344} & .0359 & $-4.2\%$ & \textbf{.0250} & .0272 & $-8.1\%$\\
Overall  & \textbf{.0347} & .0356 & $-2.6\%$ & \textbf{.0287} & .0295 & $-2.8\%$ & \textbf{.0179} & .0191 & $-6.4\%$\\
\midrule
\multicolumn{10}{@{}l}{\emph{Survival probability}}\\
D     & .0528 & .0527 & $+0.1\%$ & \textbf{.0417} & .0418 & $-0.4\%$ & .0194 & .0187 & $+3.9\%$\\
HV    & \textbf{.0649} & .0661 & $-1.8\%$ & \textbf{.0508} & .0518 & $-1.9\%$ & \textbf{.0250} & .0256 & $-2.5\%$\\
HLOW  & \textbf{.1033} & .1045 & $-1.2\%$ & \textbf{.0881} & .0889 & $-0.9\%$ & \textbf{.0659} & .0665 & $-0.8\%$\\
RARE  & \textbf{.0393} & .0412 & $-4.6\%$ & \textbf{.0280} & .0291 & $-3.7\%$ & \textbf{.0113} & .0117 & $-3.2\%$\\
EMOD  & \textbf{.0748} & .0763 & $-1.9\%$ & \textbf{.0530} & .0544 & $-2.6\%$ & \textbf{.0170} & .0170 & $-0.1\%$\\
MULTI & \textbf{.0578} & .0603 & $-4.1\%$ & \textbf{.0446} & .0472 & $-5.6\%$ & \textbf{.0207} & .0238 & $-13.2\%$\\
MISS  & \textbf{.0769} & .0793 & $-2.9\%$ & \textbf{.0611} & .0628 & $-2.7\%$ & \textbf{.0386} & .0405 & $-4.6\%$\\
Overall  & \textbf{.0665} & .0679 & $-2.1\%$ & \textbf{.0520} & .0532 & $-2.3\%$ & \textbf{.0279} & .0286 & $-2.7\%$\\
\bottomrule
\end{tabular}
\end{table}

\begin{table}[H]
\centering\small
\caption{Survival-probability version of Table~\ref{tab:baselines}: CATE error of every method (group means).}
\label{tab:baselinesSP}
\begin{tabular}{@{}l rrr rrr@{}}
\toprule
& \multicolumn{3}{c}{Group A} & \multicolumn{3}{c}{Group B}\\
\cmidrule(lr){2-4}\cmidrule(l){5-7}
Method & RMSE & MAE & $|$Bias$|$ & RMSE & MAE & $|$Bias$|$\\
\midrule
NC-CSF oracle & .0472 & .0352 & .0161 & .0473 & .0347 & .0142\\
\midrule
\textbf{CPF} & \textbf{.0703} & \textbf{.0566} & \textbf{.0331} & \textbf{.0665} & \textbf{.0520} & \textbf{.0279}\\
NC-CSF & .0804 & .0652 & .0433 & .0679 & .0532 & .0286\\
GRF-cov & .0923 & .0779 & .0488 & .0828 & .0696 & .0331\\
RSF-T & .1090 & .0863 & .0558 & .0988 & .0750 & .0418\\
RSF-S & .1139 & .0902 & .0866 & .1103 & .0847 & .0818\\
CSF-X & .1285 & .1082 & .0891 & .1306 & .1084 & .0891\\
GRF-X & .1246 & .1057 & .0888 & .1269 & .1059 & .0883\\
\bottomrule
\end{tabular}
\end{table}

\begin{table}[H]
\centering\small
\caption{Survival-probability CATE error of the ablation ladder (group means).}
\label{tab:ablationSP}
\begin{tabular}{@{}l rrr rrr@{}}
\toprule
& \multicolumn{3}{c}{Group A} & \multicolumn{3}{c}{Group B}\\
\cmidrule(lr){2-4}\cmidrule(l){5-7}
Variant & RMSE & MAE & $|$Bias$|$ & RMSE & MAE & $|$Bias$|$\\
\midrule
\textbf{CPF} & \textbf{.0703} & \textbf{.0566} & .0331 & \textbf{.0665} & \textbf{.0520} & \textbf{.0279}\\
w/o $X$-conditioning & .0706 & .0569 & .0334 & .0668 & .0523 & .0280\\
w/o bridges & .0819 & .0694 & .0325 & .0799 & .0664 & .0289\\
InfoNCE heads & .0843 & .0688 & .0478 & .0970 & .0774 & .0559\\
CCA $\to$ PCA & .0804 & .0652 & .0433 & .0679 & .0532 & .0286\\
w/o dictionary (CPF) & .0848 & .0695 & .0465 & .0752 & .0596 & .0360\\
w/o dictionary (NC-CSF) & .1057 & .0885 & .0677 & .0728 & .0572 & .0333\\
\bottomrule
\end{tabular}
\end{table}

\begin{table}[H]
\centering\footnotesize
\caption{Scale experiments around the base setting D, varying the training sample size $n$ and the covariate dimension $p_x$ one at a time (6 strong-confounding cases $\times$ 10 repetitions = 60 datasets per row; pool fixed at 10{,}000). The D row has $n=3{,}000$ and $p_x=5$; it excludes the weak-confounding cases and therefore differs slightly from Table~\ref{tab:fullB}. RMSE columns give the values for CPF and NC-CSF, $\Delta$ is the relative change (\%) of CPF versus NC-CSF, and $^\dagger$ marks $|z|<2$. Removing $X$-conditioning increased RMSE by at most 1.7\% in any cell.}
\label{tab:scale}
\setlength{\tabcolsep}{2.4pt}
\begin{tabular}{@{}lrr rrrrr rrrrr@{}}
\toprule
& & & \multicolumn{5}{c}{RMST} & \multicolumn{5}{c}{Survival probability}\\
\cmidrule(lr){4-8}\cmidrule(l){9-13}
Setting & $n$ & $p_x$ & CPF & NC-CSF & $\Delta$RMSE & $\Delta$MAE & $\Delta|$Bias$|$ & CPF & NC-CSF & $\Delta$RMSE & $\Delta$MAE & $\Delta|$Bias$|$\\
\midrule
N1000 & 1{,}000 & 5 & .0395 & .0387 & $+1.9^\dagger$ & $+2.3^\dagger$ & $+2.6^\dagger$ & .0787 & .0803 & $-2.0^\dagger$ & $-1.8^\dagger$ & $-4.1^\dagger$\\
D     & 3{,}000 & 5 & .0297 & .0302 & $-1.8^\dagger$ & $-2.9$ & $-5.9^\dagger$ & .0592 & .0587 & $+0.9^\dagger$ & $+0.3^\dagger$ & $+8.5^\dagger$\\
N8000 & 8{,}000 & 5 & .0263 & .0270 & $-2.5^\dagger$ & $-3.3$ & $-8.1^\dagger$ & .0499 & .0514 & $-2.8$ & $-4.1$ & $+2.1^\dagger$\\
\midrule
X2    & 3{,}000 & 2 & .0279 & .0274 & $+1.8^\dagger$ & $+1.0^\dagger$ & $+5.4^\dagger$ & .0559 & .0563 & $-0.7^\dagger$ & $-1.3^\dagger$ & $-0.8^\dagger$\\
X10   & 3{,}000 & 10 & .0349 & .0346 & $+1.0^\dagger$ & $+0.3^\dagger$ & $-4.0^\dagger$ & .0650 & .0648 & $+0.3^\dagger$ & $-0.1^\dagger$ & $+2.4^\dagger$\\
X20   & 3{,}000 & 20 & .0403 & .0407 & $-1.0^\dagger$ & $-1.7^\dagger$ & $-6.3^\dagger$ & .0651 & .0647 & $+0.7^\dagger$ & $+0.3^\dagger$ & $+0.5^\dagger$\\
X40   & 3{,}000 & 40 & .0435 & .0441 & $-1.4^\dagger$ & $-1.2^\dagger$ & $-3.3^\dagger$ & .0690 & .0701 & $-1.6^\dagger$ & $-0.4^\dagger$ & $-6.3^\dagger$\\
\bottomrule
\end{tabular}
\end{table}

\paragraph{External baselines.} CPF was significantly more accurate than every external baseline (\cref{tab:baselines}) in both groups, both targets and all three metrics. Its RMST error was \pct{24}--\pct{40} (group A) and \pct{14}--\pct{22} (group B) lower than that of covariate adjustment with the same PCA proxies (GRF-cov), and adjustment for $X$ alone had about twice its RMSE. GRF-cov and GRF-X are causal survival forests from the grf package \citep{athey2019grf,cui2023csf} on $X$ with and without the PCA proxies; RSF-T and RSF-S are T- and S-learners with random survival forests \citep{ishwaran2008rsf} on $X$ and the PCA proxies; CSF-X is the NC-CSF engine without proxies.

\begin{table}[h]
\centering\small
\caption{RMST CATE error of CPF and the external baselines, group means.}
\label{tab:baselines}
\begin{tabular}{@{}l rrr rrr@{}}
\toprule
& \multicolumn{3}{c}{Group A} & \multicolumn{3}{c}{Group B}\\
\cmidrule(lr){2-4}\cmidrule(l){5-7}
Method & RMSE & MAE & \Bias & RMSE & MAE & \Bias\\
\midrule
\textbf{CPF} & \textbf{.0374} & \textbf{.0322} & \textbf{.0221} & \textbf{.0347} & \textbf{.0287} & \textbf{.0179}\\
\midrule
GRF-cov & .0492 & .0439 & .0365 & .0405 & .0344 & .0228\\
RSF-T & .0672 & .0564 & .0423 & .0576 & .0462 & .0300\\
RSF-S & .0734 & .0679 & .0675 & .0664 & .0593 & .0588\\
CSF-X & .0799 & .0738 & .0694 & .0777 & .0702 & .0655\\
GRF-X & .0782 & .0728 & .0691 & .0762 & .0693 & .0652\\
\bottomrule
\end{tabular}
\end{table}

\paragraph{Other representation learners.} \cref{tab:reps} gives MAE alongside \cref{tab:main} and \cref{tab:repsSP} the survival probability. Without style PLS ties CPF ($-$\pct{0.2}, $z=-0.5$): when no modality-specific variance dominates, the whitening that separates CCA from PLS changes little.

\begin{table}[h]
\centering\small
\caption{RMST CATE error with proxies from other representation learners in the same NC-CSF estimator, and covariate adjustment for every raw feature (GRF-all), group means. $\Delta$RMSE: change versus CPF (positive = worse); $^\dagger$: $|z|<2$.}
\label{tab:reps}
\setlength{\tabcolsep}{2.5pt}
\IfFileExists{r3/tab_reps_body.tex}{\begin{tabular}{@{}l rrrr rrrr@{}}
\toprule
& \multicolumn{4}{c}{Group A (style)} & \multicolumn{4}{c}{Group B}\\
\cmidrule(lr){2-5}\cmidrule(l){6-9}
Proxies & RMSE & MAE & \Bias & $\Delta$RMSE & RMSE & MAE & \Bias & $\Delta$RMSE\\
\midrule
CPF (ours) & \textbf{.0374} & \textbf{.0321} & \textbf{.0221} & -- & \textbf{.0346} & \textbf{.0286} & \textbf{.0178} & --\\
\midrule
NC-CSF (PCA) & .0452 & .0397 & .0319 & $+20.8$ & .0356 & .0295 & .0191 & $+2.9$\\
PLS & .0438 & .0382 & .0299 & $+17.1$ & .0345 & .0286 & .0177 & $-0.2$\ns\\
Deep CCA & .0430 & .0377 & .0286 & $+14.9$ & .0356 & .0296 & .0189 & $+2.8$\\
Embedding, uncompressed & .0562 & .0507 & .0438 & $+50.1$ & .0377 & .0317 & .0199 & $+9.0$\\
GRF, all features & .0487 & .0436 & .0349 & $+30.1$ & .0424 & .0363 & .0254 & $+22.4$\\
\bottomrule
\end{tabular}
}{\pending{pending}}
\end{table}

\begin{table}[h]
\centering\small
\caption{Survival-probability version of \cref{tab:reps}. CPF and NC-CSF rows come from the run that produced the table and can differ from \cref{tab:summary} in the last digit.}
\label{tab:repsSP}
\setlength{\tabcolsep}{2.5pt}
\IfFileExists{r3/tab_reps_sp_body.tex}{\begin{tabular}{@{}l rrrr rrrr@{}}
\toprule
& \multicolumn{4}{c}{Group A (style)} & \multicolumn{4}{c}{Group B}\\
\cmidrule(lr){2-5}\cmidrule(l){6-9}
Proxies & RMSE & MAE & \Bias & $\Delta$RMSE & RMSE & MAE & \Bias & $\Delta$RMSE\\
\midrule
CPF (ours) & \textbf{.0703} & \textbf{.0566} & \textbf{.0332} & -- & \textbf{.0664} & \textbf{.0520} & \textbf{.0278} & --\\
\midrule
NC-CSF (PCA) & .0803 & .0652 & .0432 & $+14.3$ & .0679 & .0532 & .0287 & $+2.3$\\
PLS & .0779 & .0629 & .0408 & $+10.8$ & .0667 & .0522 & .0274 & $+0.5$\ns\\
Deep CCA & .0776 & .0631 & .0413 & $+10.4$ & .0682 & .0534 & .0296 & $+2.7$\\
Embedding, uncompressed & .0955 & .0799 & .0580 & $+35.9$ & .0721 & .0580 & .0298 & $+8.5$\\
GRF, all features & .0946 & .0811 & .0484 & $+34.6$ & .0859 & .0724 & .0368 & $+29.4$\\
\bottomrule
\end{tabular}
}{\pending{pending}}
\end{table}

\paragraph{Ablation.} In group A (\cref{tab:ablation}), CCA lowered RMSE by \pct{17.3} against PCA on the same dictionary, the raw-input dictionary lowered it by \pct{18.9} (mean bias \pct{31.3}), and CPF as a whole lowered RMSE and mean bias by \pct{38.6} and \pct{55.3} against the common practice, NC-CSF on the PCA of pretrained embeddings. InfoNCE heads instead of the closed form raised RMSE by \pct{20}--\pct{58}; removing the bridges raised it by \pct{17}--\pct{20} for the survival probability and \pct{1}--\pct{7} for RMST; removing $X$-conditioning changed it by less than \pct{1}, as \cref{cor:xcond} predicts when $X$ enters each modality through a single linear index.

\begin{table}[h]
\centering\small
\caption{Ablation ladder with MAE: RMST CATE error (group means); \cref{tab:ablate} is the main-text version. The two `w/o dictionary' rows cover 249 (A) and 427--428 (B) datasets.}
\label{tab:ablation}
\setlength{\tabcolsep}{2.3pt}
\begin{tabular}{@{}l rrr rrr@{}}
\toprule
& \multicolumn{3}{c}{Group A} & \multicolumn{3}{c}{Group B}\\
\cmidrule(lr){2-4}\cmidrule(l){5-7}
Variant & RMSE & MAE & \Bias & RMSE & MAE & \Bias\\
\midrule
\textbf{CPF} & \textbf{.0374} & \textbf{.0322} & .0221 & \textbf{.0347} & \textbf{.0287} & .0179\\
w/o $X$-conditioning & .0376 & .0323 & .0223 & .0349 & .0288 & .0180\\
w/o bridges & .0377 & .0327 & .0205 & .0371 & .0312 & .0174\\
InfoNCE heads & .0483 & .0426 & .0350 & .0547 & .0467 & .0403\\
CCA $\to$ PCA & .0452 & .0398 & .0320 & .0356 & .0295 & .0191\\
w/o dictionary (CPF) & .0473 & .0419 & .0330 & .0402 & .0339 & .0240\\
w/o dictionary (NC-CSF) & .0624 & .0568 & .0508 & .0388 & .0325 & .0227\\
\bottomrule
\end{tabular}
\end{table}

\begin{figure}[h]
\centering
\figorpending{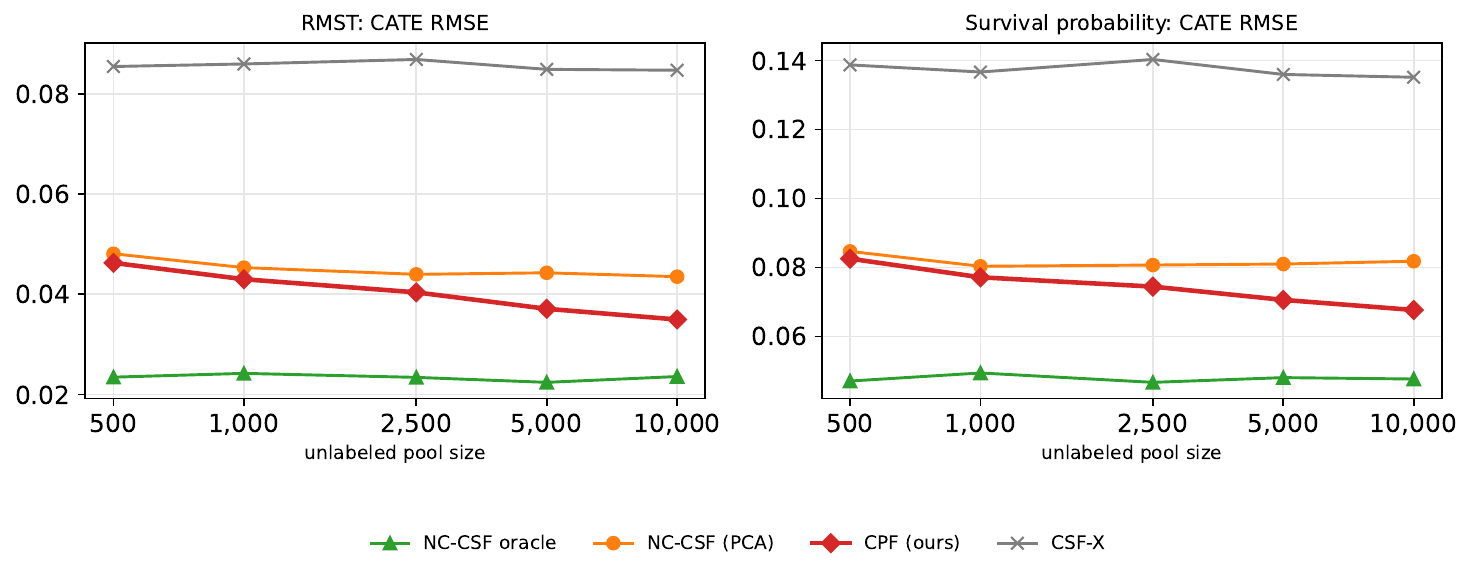}{0.75}
\caption{CATE error versus the unlabeled pool size (S8 design).}
\label{fig:pool}
\end{figure}

\begin{figure}[h]
\centering
\figorpending{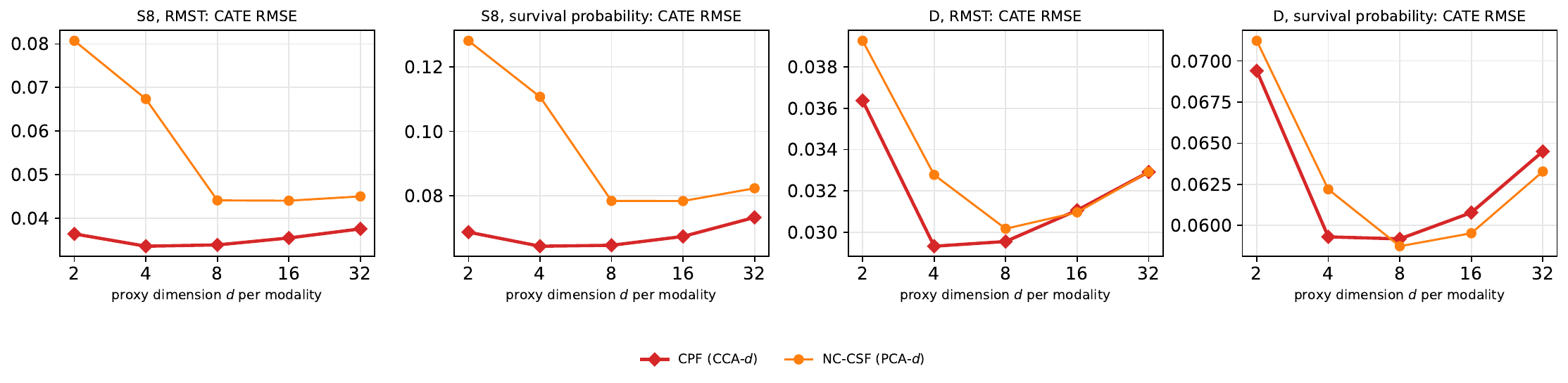}{0.85}
\caption{CATE error versus the proxy dimension $d$ (CPF: CCA-$d$; NC-CSF: PCA-$d$), with style (S8) and without (D).}
\label{fig:dsweep}
\end{figure}

\begin{figure}[h]
\centering
\includegraphics[width=0.8\textwidth]{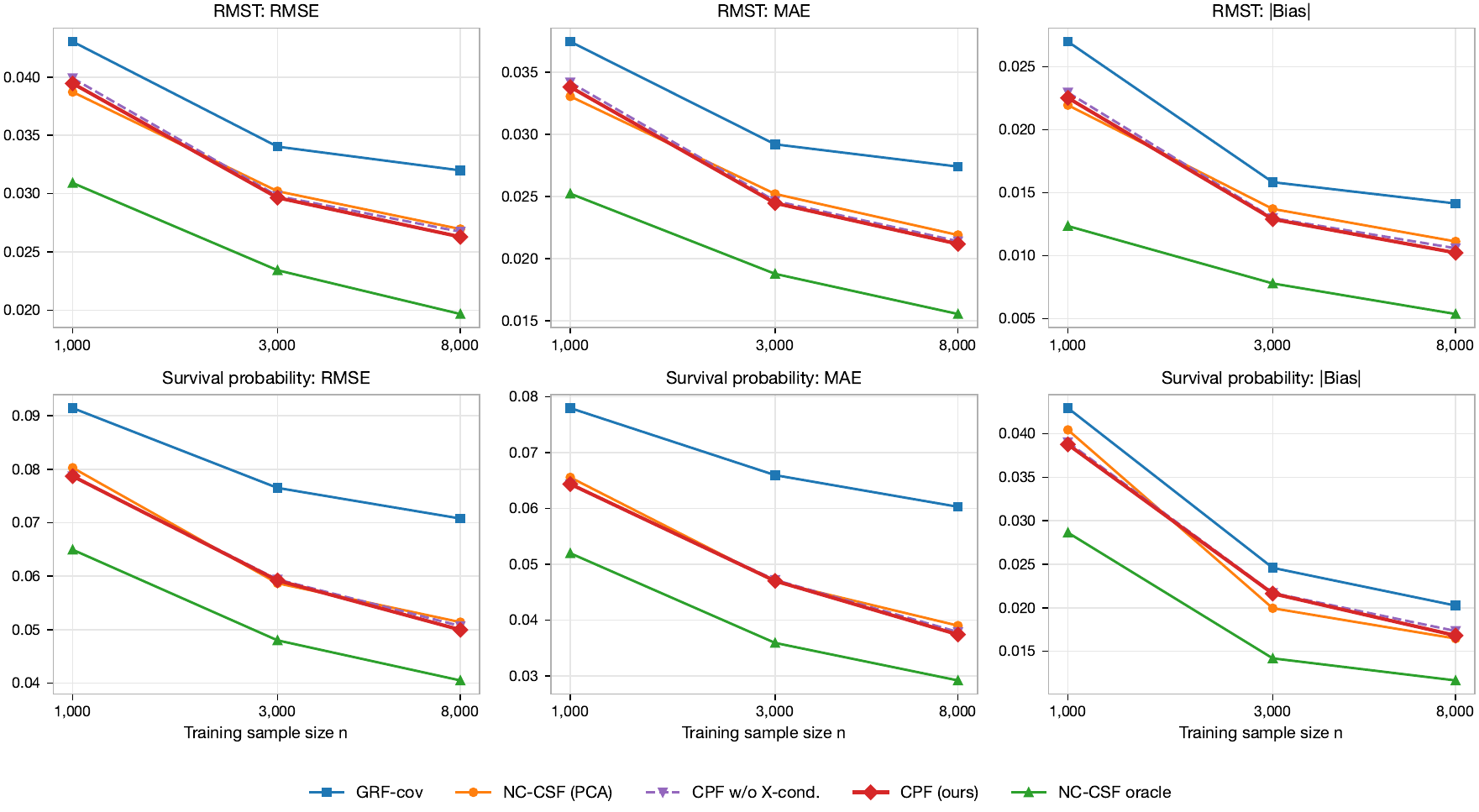}
\caption{CATE error versus training sample size $n$ (base setting D, pool fixed at 10{,}000).}
\label{fig:scale_n}
\end{figure}

\begin{figure}[h]
\centering
\includegraphics[width=0.8\textwidth]{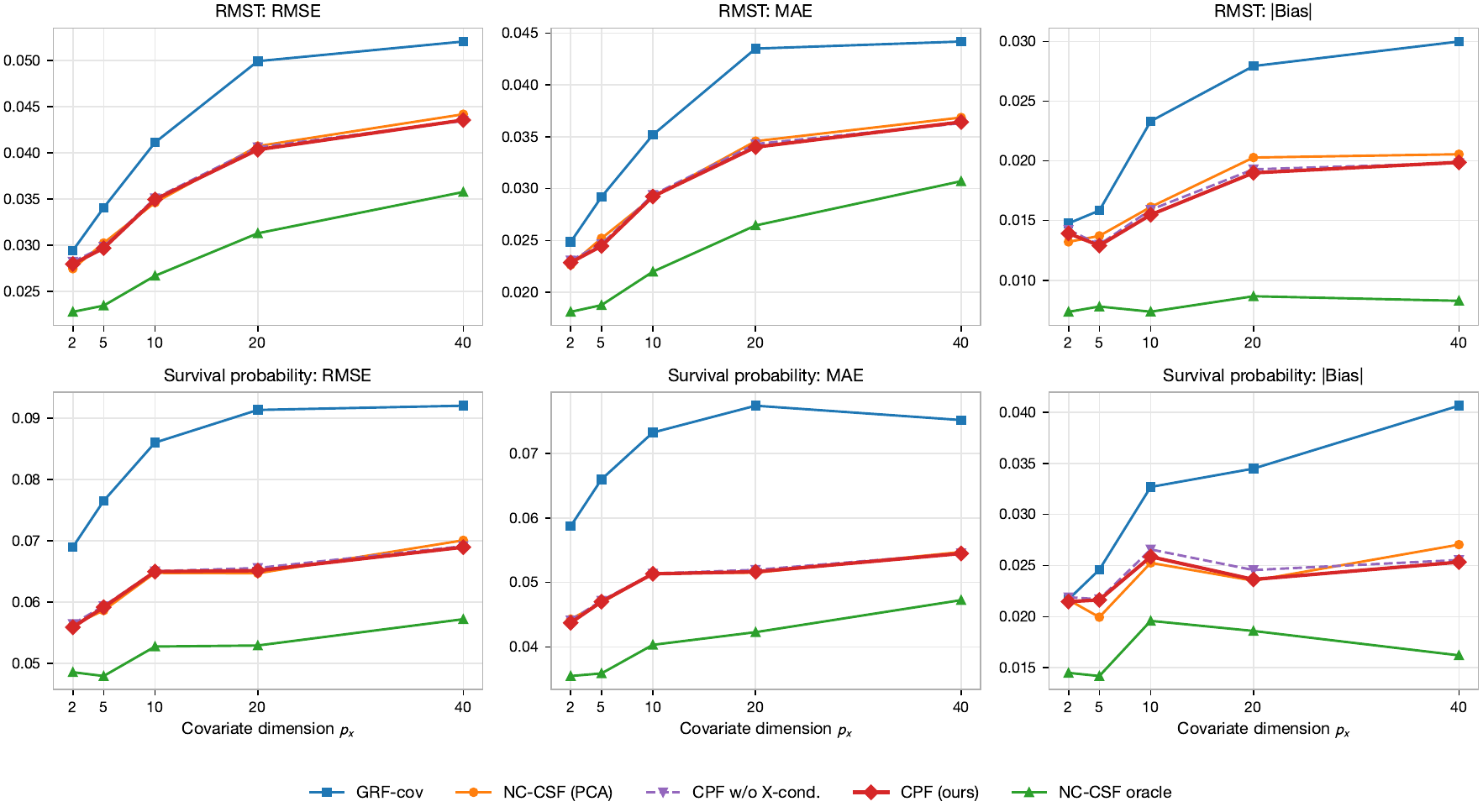}
\caption{CATE error versus covariate dimension $p_x$ (otherwise as in \cref{fig:scale_n}).}
\label{fig:scale_px}
\end{figure}

\paragraph{Style, pool size and proxy dimension.} \cref{tab:phase} gives the numbers behind \cref{fig:phase}. With style (S8) CPF beat NC-CSF at every pool size, and its advantage grew with the pool, from RMSE $-$\pct{3.8} at 500 rows to $-$\pct{19.6} at 10{,}000, consistent with \cref{thm:efficiency}(c) (\cref{tab:pool}, \cref{fig:pool}). Across proxy dimensions (\cref{tab:dsweep}, \cref{fig:dsweep}) CPF's error is lowest at $d=4$--$8$ and rises toward $d=32$; the best PCA ($d=16$, .0440) still had \pct{31} higher error than CPF at $d=4$ (.0336), and CPF beat PCA at every $d$ under style. Around D, CPF matched NC-CSF for training sizes 1{,}000--8{,}000 and covariate dimensions 2--40, and all four significant differences favored CPF (\cref{tab:scale}, Figures~\ref{fig:scale_n}--\ref{fig:scale_px}).

\begin{table}[h]
\centering\small
\caption{Numbers behind \cref{fig:phase} (RMST). $\Delta$: CPF versus NC-CSF ($^\dagger$: $|z|<2$).}
\label{tab:phase}
\IfFileExists{r3/tab_phase_body.tex}{\begin{tabular}{@{}r rrr rr rr@{}}
\toprule
& \multicolumn{3}{c}{RMST CATE RMSE} & \multicolumn{2}{c}{oracle gap closed (\%)} & \multicolumn{2}{c}{linear $R^2$ of $U$}\\
\cmidrule(lr){2-4}\cmidrule(lr){5-6}\cmidrule(l){7-8}
style & CPF & NC-CSF & $\Delta$ (\%) & CPF & NC-CSF & CPF & NC-CSF\\
\midrule
$\times0$ & .0318 & .0317 & $+0.5$\ns & 87 & 87 & 0.85 & 0.85\\
$\times0.25$ & .0296 & .0306 & $-3.3$ & 90 & 89 & 0.85 & 0.84\\
$\times0.5$ & .0300 & .0324 & $-7.5$ & 90 & 86 & 0.85 & 0.82\\
$\times1$ & .0302 & .0343 & $-12.1$ & 89 & 82 & 0.84 & 0.80\\
$\times2$ & .0303 & .0376 & $-19.2$ & 90 & 78 & 0.82 & 0.77\\
$\times4$ & .0333 & .0420 & $-20.7$ & 85 & 71 & 0.79 & 0.71\\
$\times8$ & .0333 & .0442 & $-24.8$ & 84 & 66 & 0.75 & 0.66\\
$\times16$ & .0361 & .0468 & $-22.8$ & 80 & 62 & 0.72 & 0.61\\
\bottomrule
\end{tabular}
}{\pending{pending}}
\end{table}

\begin{table}[h]
\centering\small
\caption{Unlabeled pool size on the S8 design (60 datasets per row). $\Delta$: CPF versus NC-CSF.}
\label{tab:pool}
\IfFileExists{r3/tab_pool_body.tex}{\begin{tabular}{@{}r rrr rrr@{}}
\toprule
& \multicolumn{3}{c}{RMST CATE RMSE} & \multicolumn{3}{c}{Survival probability CATE RMSE}\\
\cmidrule(lr){2-4}\cmidrule(l){5-7}
pool & CPF & NC-CSF & $\Delta$ (\%) & CPF & NC-CSF & $\Delta$ (\%)\\
\midrule
500 & .0463 & .0481 & $-3.8$ & .0825 & .0846 & $-2.5$\\
1,000 & .0430 & .0453 & $-5.1$ & .0771 & .0803 & $-4.0$\\
2,500 & .0404 & .0440 & $-8.2$ & .0744 & .0807 & $-7.7$\\
5,000 & .0371 & .0443 & $-16.2$ & .0706 & .0810 & $-12.8$\\
10,000 & .0350 & .0435 & $-19.6$ & .0677 & .0818 & $-17.3$\\
\bottomrule
\end{tabular}
}{\pending{pending}}
\end{table}

\begin{table}[h]
\centering\small
\caption{Proxy dimension $d$ (RMST CATE RMSE; 60 datasets per cell). NC-CSF uses PCA-$d$, so this also tunes PCA's dimension.}
\label{tab:dsweep}
\IfFileExists{r3/tab_dsweep_body.tex}{\begin{tabular}{@{}r rrr rrr@{}}
\toprule
& \multicolumn{3}{c}{S8 (style $\times8$)} & \multicolumn{3}{c}{D (no style)}\\
\cmidrule(lr){2-4}\cmidrule(l){5-7}
$d$ & CPF & NC-CSF (PCA-$d$) & $\Delta$ (\%) & CPF & NC-CSF (PCA-$d$) & $\Delta$ (\%)\\
\midrule
2 & .0364 & .0807 & $-54.9$ & .0364 & .0393 & $-7.4$\ns\\
4 & .0336 & .0674 & $-50.2$ & .0293 & .0328 & $-10.5$\\
8 & .0339 & .0441 & $-23.2$ & .0296 & .0302 & $-2.0$\ns\\
16 & .0355 & .0440 & $-19.4$ & .0311 & .0310 & $+0.3$\ns\\
32 & .0376 & .0450 & $-16.6$ & .0329 & .0329 & $-0.0$\ns\\
\bottomrule
\end{tabular}
}{\pending{pending}}
\end{table}

\begin{figure}[h]
\centering
\includegraphics[width=0.8\textwidth]{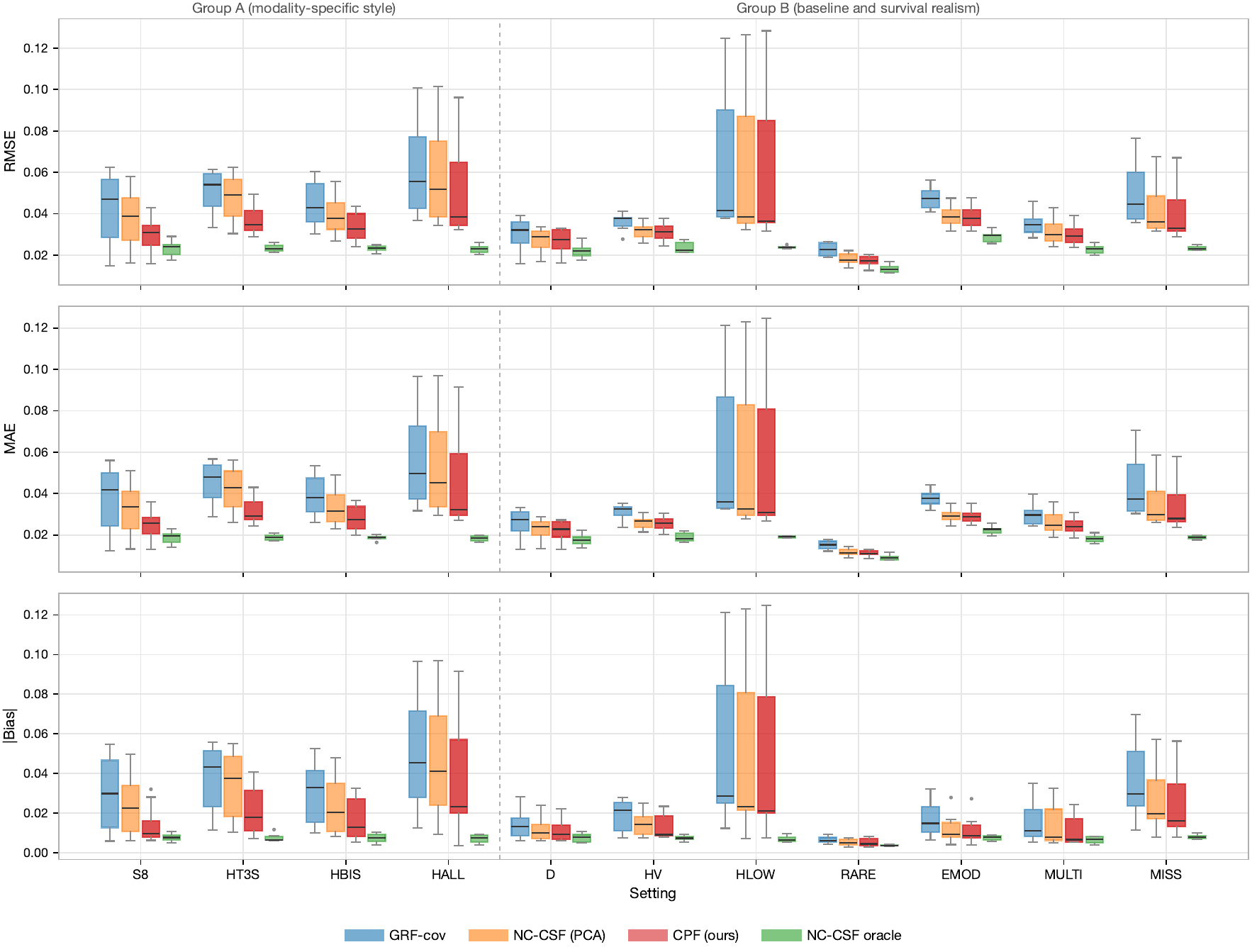}
\caption{Distribution of RMST CATE error by setting (per-case means). Group A lies left of the dashed line.}
\label{fig:box_rmst}
\end{figure}

\begin{figure}[h]
\centering
\includegraphics[width=0.8\textwidth]{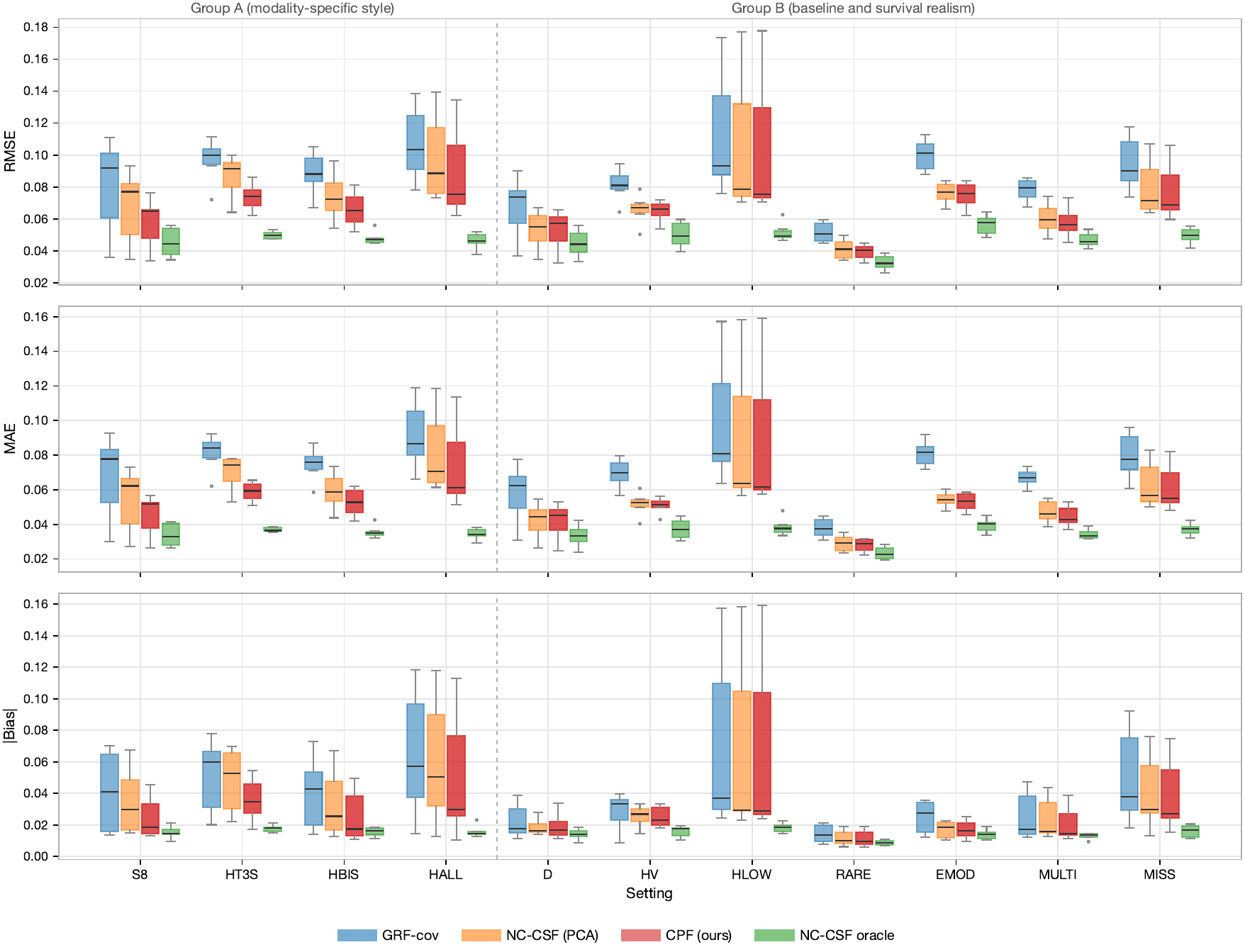}
\caption{Distribution of survival-probability CATE error by setting. Layout as in \cref{fig:box_rmst}.}
\label{fig:box_sp}
\end{figure}

\section{Raw records versus learned proxies: details}\label{app:raw}
The comparison uses the 120 d-sweep datasets (six strong-confounding cases $\times$ \{S8, D\} $\times$ 10 repetitions), whose CPF and PCA arms agree with the d-sweep's $d=8$ results to within .0005 in every group mean. The raw arm hands the estimator the full dictionary that CPF's CCA reads (2{,}103 token and 112 waveform columns on S8), so the two arms receive the same information and differ only in the compression. NC-CSF on the raw dictionary is the engine as released: its treatment bridge is logistic on $[X,Z]$, its outcome bridge a forest on $[X,W]$, and its arm-specific and censoring Cox models see $[X,W,Z]$; nothing is pre-screened, since pre-screening would itself be a compression. NC-CF is run on the raw dictionary in the same way, except that it has no Cox models. \cref{tab:rawfull} reports all four outcomes.

\begin{table}[h]
\centering\small
\caption{Raw records versus learned proxies for all four outcomes: CATE RMSE and absolute mean bias, averaged over the 60 datasets in each setting (NC-CSF for survival, NC-CF otherwise). $\Delta$: change in the RMSE of the raw dictionary relative to CPF (\%), paired over datasets.}
\label{tab:rawfull}
\setlength{\tabcolsep}{4pt}
\begin{tabular}{@{}ll rrrr rrrr r@{}}
\toprule
&& \multicolumn{4}{c}{RMSE} & \multicolumn{4}{c}{\Bias} &\\
\cmidrule(lr){3-6}\cmidrule(lr){7-10}
Outcome (estimator) & Setting & CPF & PCA & Raw & True $U$ & CPF & PCA & Raw & True $U$ & $\Delta$\\
\midrule
RMST (NC-CSF) & S8 & .0341 & .0440 & .0987 & .0248 & .0165 & .0298 & .0698 & .0082 & $+189.7$\\
 & D & .0295 & .0301 & .0911 & .0234 & .0128 & .0137 & .0594 & .0078 & $+208.7$\\
Surv.\ prob.\ (NC-CSF) & S8 & .0648 & .0782 & .2702 & .0485 & .0268 & .0407 & .1868 & .0161 & $+317.1$\\
 & D & .0587 & .0585 & .2579 & .0480 & .0214 & .0199 & .1562 & .0142 & $+339.5$\\
Continuous (NC-CF) & S8 & .1250 & .1846 & .1679 & .0770 & .1046 & .1707 & .1531 & .0334 & $+34.3$\\
 & D & .0922 & .0958 & .1115 & .0816 & .0641 & .0695 & .0865 & .0362 & $+20.9$\\
Binary (NC-CF) & S8 & .0438 & .0498 & .0517 & .0404 & .0198 & .0271 & .0304 & .0152 & $+18.0$\\
 & D & .0415 & .0423 & .0454 & .0418 & .0166 & .0177 & .0227 & .0174 & $+9.3$\\
\bottomrule
\end{tabular}
\end{table}

\textbf{Why the raw records fail.} The raw dictionary was less accurate than CPF's proxies in all eight comparisons ($|z|\ge2.1$), in the style-free setting D as well as in S8, so the loss stems from dimension rather than style. With NC-CSF, most of the loss is error in the average effect: the absolute mean bias was four to five times CPF's for RMST and about seven times for the survival probability, and its sign varied across datasets (positive in 56 of the 120 RMST datasets, against 29 for CPF). This is the signature of unstable nuisance fits rather than of a confounder left unadjusted, and the Cox models are its source. NC-CSF cross-fits its nuisance models over three folds, so each arm-specific Cox model is fitted to about 1{,}000 units with about 650 events, and the censoring model to about 2{,}000 units, while both see about 2{,}200 columns. With more columns than events, the lightly penalized partial likelihood (ridge 0.01, increased only when a fit fails) overfits, and the inverse-censoring-weighted scores of the causal survival forest inherit this instability. NC-CF uses neither Cox models nor censoring weights and lost less: with the raw dictionary it closed \pct{82} (continuous) and \pct{84} (binary) of the gap between $X$-only adjustment and the oracle on S8, compared with \pct{90} and \pct{95} with CPF's proxies; on D the figures were \pct{94} and \pct{95}, compared with \pct{98} and \pct{100}.

\textbf{Other nuisance learners.} A sparser nuisance learner, such as a lasso Cox model, or screening the columns by their association with the outcome, would reduce the dimension within the estimator. It would do so, however, with the treatment and outcome of the analysis sample: a supervised compression that carries none of the guarantees of \cref{sec:theory}. CPF performs the reduction once, on the unlabeled pool, where it preserves the exclusion restrictions (\cref{thm:exclusion}) and keeps completeness under the conditions of \cref{prop:complete}.

\textbf{Computing cost.} Both arms read the same dictionary, so the only Stage-1 cost that CPF adds is its closed-form CCA and label-free selection, incurred once per pool. Under identical computing settings, the median NC-CSF fit took about 20 seconds with CPF's proxies and 5.0 hours (RMST) or 4.7 hours (survival probability) with the raw dictionary, about 870 times as long; individual raw fits took from 3.0 to 29.3 hours. Each raw fit runs about 12 Cox regressions on 2{,}200 columns (three folds and a full-sample fit, each with two arm-specific models and one censoring model), and more whenever a fit fails and is rerun with a larger penalty. NC-CF, which has no Cox models, took a median of 15 seconds with CPF's proxies and 6.6 minutes with the raw dictionary, about 26 times as long.

\section{Beyond survival: continuous and binary outcomes}\label{app:general}
Every main-benchmark dataset of groups A and B except RARE, which changes only the event-time scale, is reused with $X$, $U$, $V$, $A$, both modalities and the Stage-1 fit unchanged; only the outcome is redrawn, from the case's linear predictor $\eta_a(X,U)$, which carries its confounding and outcome surface. The continuous outcome is $Y=-\eta_A(X,U)+\varepsilon$ with $\varepsilon\sim N(0,1)$, so the effect is constant except under EMOD. The binary outcome has $P(Y=1\mid A,X,U)=\operatorname{expit}(b_y-\eta_A(X,U))$, with $b_y$ set so that $P(Y(0)=1)=0.5$; its target is the risk difference, which varies with $X$ and $U$. The estimator is NC-CF, the non-censored NC estimator released with NC-CSF \citep{nccsf}, at the forest sizes of the survival runs. With the CPF proxies it closed \pct{81} (continuous) and \pct{87} (binary) of the gap between $X$-only adjustment and the oracle in group A, against \pct{70} and \pct{78} with PCA proxies, and \pct{82} and \pct{88} in group B, against \pct{81} and \pct{87}. Covariate adjustment for the PCA proxies (a causal forest) kept \pct{44}--\pct{48} more mean bias than NC-CF with the CPF proxies in group A.

\begin{table}[h]
\centering\small
\caption{Beyond survival: CATE error (group means, strong-confounding cases) with a continuous and a binary outcome. Causal forest: EconML's CausalForestDML \citep{econml}. Bottom row: change of the CPF proxies versus the PCA proxies with NC-CF.}
\label{tab:general}
\setlength{\tabcolsep}{2.2pt}
\resizebox{\textwidth}{!}{%
\begin{tabular}{@{}ll rrr rrr rrr rrr@{}}
\toprule
&& \multicolumn{6}{c}{Continuous outcome} & \multicolumn{6}{c}{Binary outcome}\\
\cmidrule(lr){3-8}\cmidrule(l){9-14}
&& \multicolumn{3}{c}{Group A} & \multicolumn{3}{c}{Group B} & \multicolumn{3}{c}{Group A} & \multicolumn{3}{c}{Group B}\\
\cmidrule(lr){3-5}\cmidrule(lr){6-8}\cmidrule(lr){9-11}\cmidrule(l){12-14}
Estimator & Proxies & RMSE & MAE & \Bias & RMSE & MAE & \Bias & RMSE & MAE & \Bias & RMSE & MAE & \Bias\\
\midrule
NC-CF & true $U$ & .0781 & .0633 & .0339 & .0963 & .0760 & .0355 & .0402 & .0325 & .0140 & .0429 & .0347 & .0153\\
 & CPF proxies & \textbf{.1716} & \textbf{.1594} & \textbf{.1536} & \textbf{.1887} & \textbf{.1638} & \textbf{.1361} & \textbf{.0503} & \textbf{.0433} & \textbf{.0282} & \textbf{.0521} & \textbf{.0441} & \textbf{.0260}\\
 & PCA proxies & .2252 & .2154 & .2127 & .1952 & .1698 & .1424 & .0571 & .0500 & .0365 & .0526 & .0447 & .0267\\
 & no proxies ($X$ only) & .5720 & .5607 & .5602 & .6096 & .5844 & .5753 & .1155 & .1072 & .1016 & .1182 & .1087 & .1029\\
\midrule
Causal forest & $X$ + PCA proxies as covariates & .2389 & .2288 & .2267 & .2127 & .1843 & .1568 & .0611 & .0539 & .0408 & .0562 & .0480 & .0298\\
 & $X$ only & .5714 & .5592 & .5585 & .6098 & .5846 & .5754 & .1156 & .1071 & .1014 & .1182 & .1085 & .1026\\
\midrule
\multicolumn{2}{@{}l}{$\Delta$ CPF vs PCA, NC-CF (\%)} & $-23.8$ & $-26.0$ & $-27.8$ & $-3.3$ & $-3.5$ & $-4.4$ & $-11.8$ & $-13.4$ & $-22.6$ & $-1.0$ & $-1.3$ & $-2.8$\\
\bottomrule
\end{tabular}}
\end{table}

\section{Real-data and implementation details}\label{app:real}
\paragraph{MIMIC-IV cohort.} We use MIMIC-IV 3.1 \citep{johnson2024mimiciv31,johnson2023mimiciv} from PhysioNet \citep{goldberger2000physionet}. The cohort is adults discharged alive after a first admission with atrial fibrillation, with no warfarin before the admission and no direct oral anticoagulant during it, excluding hospice discharges and strokes within 7 days. Treatment is warfarin at discharge ($A=1$) against no anticoagulant ($A=0$). The outcomes are death within 360 days and ischemic stroke within 243 days, censored at death. $X$ has 20 covariates, including age, sex, the components of CHA$_2$DS$_2$-VASc \citep{lip2010chads} and HAS-BLED \citep{pisters2010hasbled}, and pre-admission laboratory values. The unlabeled pool holds 6{,}000 patients outside the analysis sample, which has 7{,}336 patients (2{,}756 treated, 1{,}461 deaths, 95 strokes).
\paragraph{MIMIC-IV modalities.} The treatment side is the radiology reports and recent discharge summaries from 30 days before admission to 6 hours after it, with every sentence on anticoagulation or the electrocardiogram removed. Its dictionary is a frozen Bio\_ClinicalBERT embedding \citep{alsentzer2019} plus TF-IDF (2{,}768 columns). The outcome side is the latest 12-lead electrocardiogram before the decision, represented by a pool-fitted autoencoder plus per-lead log spectra (1{,}264 columns, with $B=10$ Hz and $\lambda=1$ chosen without labels). Both modalities precede the first warfarin order, and the discharge summary of the index stay, which lists the discharge medications, is never used.
\paragraph{MIMIC-IV benchmark and hand-picked proxies.} The trial answer is a Cochrane meta-analysis \citep{aguilar2005} of five warfarin trials (2{,}313 participants), with odds ratios of 0.34 (0.23--0.52) for ischemic stroke and 0.69 (0.50--0.94) for death. We treat each odds ratio as a hazard ratio $\theta$ and set $S_1=S_0^{\theta}$ on the cohort's Kaplan--Meier curve \citep{kaplan1958}. This gives trial values of $+1.59$ days (stroke) and $+13.3$ days (death) of RMST, and the confidence limits give the trial ranges. The ranges reflect only the trials' sampling uncertainty, and the trials enrolled outpatients around 1990, unlike this older, hospitalized cohort. Following NC-CSF, the hand-picked proxies are $W$ = pre-admission platelets, albumin, ALT and AST, and $Z$ = English as preferred language, private insurance, weekend admission and calendar year.
\paragraph{NHANES.} We use NHANES 2003--2006 \citep{nhanes} participants aged 40 and over, with mortality linked through 2019 \citep{nchs2019lmf} and a 10-year horizon. The treatment side is a 24-hour dietary recall as a food-group token sequence, and the outcome side is 7-day accelerometry (672 dimensions). $X$ has 19 covariates, and the unlabeled pool has 2{,}805 participants aged 20--39. Supplements have no effect on mortality in RCTs \citep{gaziano2012,sesso2012mvm} but go with healthy lifestyles, and statins reduce mortality \citep{ctt2010} but are prescribed to people at high risk. The CTT all-cause mortality RR of 0.90 (0.87--0.93) per 1 mmol/L LDL reduction gives HR 0.85--0.93 for typical reductions, or $+0.8$ to $+1.8$ months of 10-year RMST in this cohort. The hand-picked proxies are chronic comorbidity ($W$) and health-care use ($Z$).
\paragraph{Style in the MIMIC-IV records.}\label{app:style} To test whether the largest variation in real records is style, we regressed each proxy direction on recording conditions and on patient state, beyond $X$, by 5-fold cross-validated ridge regression on the 7{,}336 analysis rows (\cref{tab:style}). Recording conditions are the composition of the notes (numbers, types and length of reports) and the device and filter settings of the electrocardiogram, counted beyond patient state (laboratory values, and the electrocardiogram's intervals and axes). Documentation composition explains 40\% of the variance of the notes' top eight principal components but only 7\% of CPF's directions, so the notes are in the regime of \cref{prop:pca}. Device settings explain none of the electrocardiogram's directions, and there CPF carries more measured physiology than PCA. These shares are lower bounds on style, because author, template and electrode placement are unobserved, and documentation composition may partly reflect the course of care.

\begin{table}[h]
\centering\small
\caption{What the proxy directions of the MIMIC-IV modalities encode: 5-fold cross-validated $R^2$ on the 7{,}336 analysis rows, as gains beyond $X$, with PCA directions weighted by their variance. Recording conditions are counted beyond $X$ and patient state.}
\label{tab:style}
\begin{tabular}{@{}ll rrr@{}}
\toprule
Modality & Proxies & $X$ alone & Patient state beyond $X$ & Recording conditions beyond $X$ and patient state\\
\midrule
Clinical notes & PCA-8 & .102 & .003 & \textbf{.399}\\
 & CPF & .045 & .000 & .067\\
\midrule
Electrocardiogram & PCA-8 & .025 & .066 & .000\\
 & CPF & .028 & \textbf{.089} & .000\\
\bottomrule
\end{tabular}
\end{table}

\paragraph{Implementation.} CPF uses $d=8$ proxies per modality and chooses the ridge $\lambda\in\{10^{-3},10^{-2},10^{-1},1\}$ and the waveform bandwidth $B$ on an 80/20 split of the pool. NC-CSF runs with its default settings, and intervals use 250 bootstrap replicates. Deep CCA uses two-layer MLPs with 256 units that maximize the total canonical correlation, and PLS takes the top-$d$ singular pairs of the cross-covariance.

\clearpage
\section{Complete proofs}\label{app:proofs}
This appendix gives complete proofs of the theoretical results, with the numbering of the main text. Where the main text states a result informally, the precise statement is given before its proof. Auxiliary results are numbered within this appendix. Proposition~\ref{prop:complete} is proved in Appendix~\ref{app:theory}.

\begin{center}
\small
\begin{tabular}{lll}
\toprule
Result & Content & Proof \\
\midrule
Theorem 1 & exact solution of $X$-conditional contrastive learning & \ref{s:sec:thm1} \\
Theorem 2 & exact inheritance of the exclusion restrictions & \ref{s:sec:thm2} \\
Theorem 3 & information preservation, efficiency, second-order cost & \ref{s:sec:thm3} \\
Theorem 4 & identification without linearity & \ref{s:sec:thm4nl} \\
Proposition 1 & when the learned proxies are complete & \ref{app:theory} \\
Proposition 2 & failure of PCA and of linear autoencoders & \ref{s:sec:prop1} \\
Theorem 5 & dictionary monotonicity & \ref{s:sec:thm4} \\
Proposition 3 & random-phase waveforms & \ref{s:sec:prop2} \\
Theorem 6 & finite pools and label-free selection & \ref{s:sec:thm5} \\
Lemma 1 & completeness without Gaussianity & \ref{s:sec:lem1} \\
Proposition 4 & why bridges are needed & \ref{s:sec:prop3} \\
Corollary 1 & $X$-conditioning & \ref{s:sec:cor1} \\
Appendix~\ref{app:further}, last paragraph & outcome factors present only on the outcome side & \ref{s:sec:asym} \\
\bottomrule
\end{tabular}
\end{center}

\subsection{Conventions, assumptions and imported results}\label{s:sec:setup}

\subsubsection{Notation}
All random vectors have finite second moments. For a random vector $Y$ and a collection $S$ of random variables, $\proj(Y\mid S)$ is the coordinatewise $L^2$ projection of $Y$ onto the closed linear span $\spn(S)$. The constant $1$ is listed explicitly whenever it belongs to $S$. For a random vector $Y$, $\M(Y)=\E[YY^\top]$. If $\E Y=0$ then $\M(Y)=\Var Y$. $\preceq$ is the Loewner order. $\lVert\cdot\rVert$ is the spectral norm, $\lVert\cdot\rVert_F$ the Frobenius norm, $\sigma_j(\cdot)$ the $j$-th largest singular value, and $\kappa(\Sigma)=\lambda_{\max}(\Sigma)/\lambda_{\min}(\Sigma)$.

For a feature vector $D$, $R(D)=D-\proj(D\mid 1,X)$ is its linear residual. Under (A1) below, $R(D)=D-\E[D\mid X]$. We write $R_z=R(D_z)$, $R_w=R(D_w)$, $\Sigma_{zz}=\M(R_z)$, $\Sigma_{ww}=\M(R_w)$ and $C=\E[R_zR_w^\top]$. (The letter $C$ also denotes the censoring time, but only inside (P), (E1) and (E2).) For $\lambda\ge0$, with $\lambda>0$ whenever a covariance is singular,
\[
K_\lambda=(\Sigma_{zz}+\lambda I)^{-1/2}\,C\,(\Sigma_{ww}+\lambda I)^{-1/2}=\Gamma S\,\Omega^\top
\]
is a singular value decomposition, with singular values $\rho_1(\lambda)\ge\rho_2(\lambda)\ge\dots\ge0$ (the \emph{ridge canonical correlations}, $\rho_j=\rho_j(0)$ are the canonical correlations). $\Gamma_d,\Omega_d,S_d$ are the leading $d$ columns and singular values, and
\[
\Theta_z^{(d)}(\lambda)=(\Sigma_{zz}+\lambda I)^{-1/2}\Gamma_d,\qquad \Theta_w^{(d)}(\lambda)=(\Sigma_{ww}+\lambda I)^{-1/2}\Omega_d
\]
are the top-$d$ directions. Hats denote the same quantities computed from pool moments.

For subspaces $\mathcal U,\mathcal V$ of an inner-product space with orthogonal projectors $\Pi_{\mathcal U},\Pi_{\mathcal V}$ and $\dim\mathcal U=\dim\mathcal V$, $\dist(\mathcal U,\mathcal V)=\lVert\Pi_{\mathcal U}-\Pi_{\mathcal V}\rVert$ is the sine of the largest principal angle. On $\R^p$ the default inner product is Euclidean, and $\dist_\Sigma$ uses $\langle a,b\rangle_\Sigma=a^\top\Sigma b$ for $\Sigma\succ0$.

The survival targets of the main text enter only through NC-CSF (Theorem~\ref{s:thm:S0}). The results on efficiency use a scalar outcome $Y$ observed without censoring, for example $\min(T,t_0)$ or an inverse-probability-of-censoring-weighted transform of it.

\subsubsection{Assumptions}
\begin{description}
\item[(A1)] $\E[D_z\mid X]$ and $\E[D_w\mid X]$ are affine in $X$.
\item[(A2)] $\Cov(D_z\mid X)$ and $\Cov(D_w\mid X)$ do not depend on $X$.
\item[(F)] $R_z=\Lambda_z\Lb+\varepsilon_z$ and $R_w=\Lambda_w\Lb+\varepsilon_w$, where $\Lb=L-\E[L\mid X]$ with $\Sigma_L=\Var\Lb\succ0$. $\Lambda_z\in\R^{p_z\times k}$ and $\Lambda_w\in\R^{p_w\times k}$ have rank $k$. $\varepsilon_z$, $\varepsilon_w$ and $(\Lb,X)$ are mutually independent, and $\Psi_z=\Var\varepsilon_z\succ0$, $\Psi_w=\Var\varepsilon_w\succ0$ are otherwise arbitrary. Under (F), $\Sigma_{zz}=\Lambda_z\Sigma_L\Lambda_z^\top+\Psi_z$, $\Sigma_{ww}=\Lambda_w\Sigma_L\Lambda_w^\top+\Psi_w$ and $C=\Lambda_z\Sigma_L\Lambda_w^\top$.
\item[(P)] $M_z\Ind(T(0),T(1),C,M_w)\mid A,L,X$ and $M_w\Ind(A,M_z)\mid L,X$.
\item[(S)] The pool $\mathfrak P$ and the seeds $\varsigma$ used by Stage~1 are independent of the analysis sample, whose units are i.i.d.
\item[(Pos)] $0<P(A=1\mid L,X)<1$ almost surely.
\end{description}
\noindent\textbf{Linear proximal model.} Used for efficiency (Theorem~3(b,c), Theorem~\ref{s:thm:monotone}(c))
\begin{description}
\item[(L1)] (F) holds and $\E[L\mid X]$ is affine in $X$.
\item[(L2)] $Y=c_0+\tau A+\beta^\top\Lb+\gamma^\top X+e$.
\item[(L3)] $e$, $\varepsilon_z$, $\varepsilon_w$ and $(A,X,\Lb)$ are mutually independent. (In particular $Z\Ind Y\mid A,L,X$, $W\Ind(A,Z)\mid L,X$, and $Z$ has no direct effect on $A$.)
\item[(L4)] $\E[(1,X^\top)^\top(1,X^\top)]\succ0$. $s_A=\Var(A-\proj(A\mid1,X))>0$, and $\Var(\Lb-\proj(\Lb\mid1,X,A))\succ0$ (no linear combination of $\Lb$ is an affine function of $(X,A)$).
\end{description}
No Gaussianity is assumed anywhere except where stated.

\begin{sdef}[Completeness]\label{s:def:complete}
A proxy $V$ is \emph{complete for $L$} if, for almost every $(a,x)$ and every $g$ with $\E[|g(L)|\mid X=x]<\infty$, $\E[g(L)\mid V,A=a,X=x]=0$ almost surely implies $g(L)=0$ almost surely given $(A,X)=(a,x)$.
\end{sdef}

\subsubsection{Results taken as given}
\begin{sthm}[NC-CSF identification; \citealp{nccsf}]\label{s:thm:S0}
Let $(Z^\ast,W^\ast)$ satisfy (E1) $Z^\ast\Ind(T(0),T(1),C,W^\ast)\mid A,L,X$ and (E2) $W^\ast\Ind(A,Z^\ast)\mid L,X$, NC-CSF's latent ignorability, positivity and censoring assumptions with latent $L$, completeness for $L$, and existence of the bridge functions. Then NC-CSF identifies $\tau_{\rm RMST}$ and $\tau_{\rm SP}$, and its estimator has the guarantees of \citet{nccsf}.
\end{sthm}
CPF therefore has to deliver proxies that satisfy (E1)--(E2) (Theorem~\ref{s:thm:exclusion}) and are relevant and complete (Theorem~3(a), Lemma~\ref{s:lem:complete}, Corollary~\ref{s:cor:cpfcomplete}).

\begin{slemma}[A shared non-confounder enlarges the latent]\label{s:lem:sharedV}
Suppose $V\Ind(A,T(0),T(1),C)\mid U,X$ and latent ignorability holds with $U$, i.e.\ $(T(0),T(1),C)\Ind A\mid U,X$. Then latent ignorability holds with $L=(U,V)$. If moreover each modality is a measurable function of $(U,V,X)$ and of noise independent of everything else and of the other modality's noise, then (P) holds with $L$.
\end{slemma}
\begin{proof}
Write $\mathbf T=(T(0),T(1),C)$. By the first assumption $P(\mathbf T,A\mid U,V,X)=P(\mathbf T,A\mid U,X)=P(\mathbf T\mid U,X)P(A\mid U,X)$, and again by the first assumption $P(\mathbf T\mid U,X)=P(\mathbf T\mid U,V,X)$ and $P(A\mid U,X)=P(A\mid U,V,X)$. Hence $\mathbf T\Ind A\mid L,X$. Given $(L,X)$ the modalities are functions of independent noises, so they are independent of each other and of $(A,\mathbf T)$, which gives (P).
\end{proof}
A shared non-confounder is therefore harmless for validity, but completeness is required for $L$, of dimension $k=\dim U+\dim V$, so $d\ge k$ proxies are needed.

\subsubsection{Elementary facts}
\begin{slemma}[Matrix facts]\label{s:lem:matrix}
\begin{enumerate}
\item[(i)] (Eckart--Young--Mirsky.) Let $K\in\R^{p\times q}$ have singular values $\rho_1\ge\rho_2\ge\dots$ and SVD $K=\Gamma S\Omega^\top$. Over all $N$ with $\rank N\le d$, $\min\lVert K-N\rVert_F^2=\sum_{j>d}\rho_j^2$, attained at $N=\Gamma_dS_d\Omega_d^\top$ for every SVD of $K$. If $\rho_d>\rho_{d+1}$ the minimizer is unique. Moreover $\min_{\rank N<j}\lVert K-N\rVert=\rho_j$.
\item[(ii)] $\sigma_j(ABC)\le\lVert A\rVert\,\sigma_j(B)\,\lVert C\rVert$ for conformable matrices.
\item[(iii)] If $\mathcal S_1\subseteq\mathcal S_2$ are closed subspaces of mean-zero $L^2$ variables and $Y\in(L^2)^k$, then $\M(\proj(Y\mid\mathcal S_2))=\M(\proj(Y\mid\mathcal S_1))+\M(\proj(Y\mid\mathcal S_2)-\proj(Y\mid\mathcal S_1))\succeq\M(\proj(Y\mid\mathcal S_1))$.
\item[(iv)] If $0\prec H_1\preceq H_2$ then $\alpha^\top H_2^{-1}\alpha\le\alpha^\top H_1^{-1}\alpha$ for every $\alpha$. For any $H_1,H_2\succ0$, $|\alpha^\top H_1^{-1}\alpha-\alpha^\top H_2^{-1}\alpha|\le\lVert\alpha\rVert^2\lVert H_1^{-1}\rVert\lVert H_2^{-1}\rVert\lVert H_1-H_2\rVert$.
\end{enumerate}
\end{slemma}
\begin{proof}
(i) is the theorem of \citet{eckart1936} and \citet{mirsky1960}. Uniqueness under a gap is standard \citep[\S I.4]{stewart1990}. (ii) By (i) there is $N$ of rank $<j$ with $\lVert B-N\rVert=\sigma_j(B)$. Then $ANC$ has rank $<j$, so by (i) $\sigma_j(ABC)\le\lVert ABC-ANC\rVert\le\lVert A\rVert\sigma_j(B)\lVert C\rVert$. (iii) Let $\Delta=\proj(Y\mid\mathcal S_2)-\proj(Y\mid\mathcal S_1)=(Y-\proj(Y\mid\mathcal S_1))-(Y-\proj(Y\mid\mathcal S_2))$. Both brackets are orthogonal to $\mathcal S_1$, so every coordinate of $\Delta$ is orthogonal to every coordinate of $\proj(Y\mid\mathcal S_1)$ and the cross moments vanish. (iv) $H_1\preceq H_2$ gives $H_2^{-1/2}H_1H_2^{-1/2}\preceq I$, hence $H_2^{1/2}H_1^{-1}H_2^{1/2}\succeq I$, i.e.\ $H_1^{-1}\succeq H_2^{-1}$. The bound follows from $H_1^{-1}-H_2^{-1}=H_1^{-1}(H_2-H_1)H_2^{-1}$.
\end{proof}

\begin{slemma}[Distances between subspaces]\label{s:lem:subspace}
Let $\mathcal U,\mathcal V\subseteq\R^p$ have equal dimension.
\begin{enumerate}
\item[(i)] $\dist$ is a metric on subspaces of a given dimension, $\dist(\mathcal U,\mathcal V)=\lVert(I-\Pi_{\mathcal V})\Pi_{\mathcal U}\rVert$, and $\lVert u-\Pi_{\mathcal V}u\rVert\le\dist(\mathcal U,\mathcal V)\lVert u\rVert$ for $u\in\mathcal U$ \citep[\S I.5]{stewart1990}.
\item[(ii)] For $\Sigma\succ0$, $\dist_\Sigma(\mathcal U,\mathcal V)\le\kappa(\Sigma)^{1/2}\dist(\mathcal U,\mathcal V)$.
\item[(iii)] If $B\in\R^{p\times k}$ has rank $k$ and $\lVert B'-B\rVert\le\epsilon<\sigma_{\min}(B)$, then $\dist(\col B',\col B)\le\epsilon/(\sigma_{\min}(B)-\epsilon)$.
\item[(iv)] There are orthonormal bases $U$ of $\mathcal U$ and $V$ of $\mathcal V$ with $\lVert U-V\rVert\le\sqrt2\,\dist(\mathcal U,\mathcal V)$.
\end{enumerate}
\end{slemma}
\begin{proof}
(i) The triangle inequality is that of the operator norm. The rest is cited. (ii) Let $\Pi^\Sigma_{\mathcal V}$ be the $\Sigma$-orthogonal projector. For $u\in\mathcal U$, $\lVert u-\Pi^\Sigma_{\mathcal V}u\rVert_\Sigma\le\lVert u-\Pi_{\mathcal V}u\rVert_\Sigma\le\lambda_{\max}^{1/2}\lVert u-\Pi_{\mathcal V}u\rVert\le\lambda_{\max}^{1/2}\dist(\mathcal U,\mathcal V)\lVert u\rVert\le\kappa^{1/2}\dist(\mathcal U,\mathcal V)\lVert u\rVert_\Sigma$. Take the maximum over $\lVert u\rVert_\Sigma=1$ and apply (i) in the $\Sigma$-geometry. (iii) $\sigma_{\min}(B')\ge\sigma_{\min}(B)-\epsilon>0$, so $\rank B'=k$. For $y=B'c$, $\lVert y-\Pi_{\col B}y\rVert\le\lVert B'c-Bc\rVert\le\epsilon\lVert c\rVert\le\epsilon\lVert y\rVert/\sigma_{\min}(B')$. Apply (i). (iv) Take any orthonormal bases $U_0,V_0$ and the SVD $V_0^\top U_0=W_1(\cos\Theta)W_2^\top$. With $U=U_0W_2$ and $V=V_0W_1$, $(U-V)^\top(U-V)=2(I-\cos\Theta)$, so $\lVert U-V\rVert^2=2(1-\cos\theta_{\max})\le2\sin^2\theta_{\max}$.
\end{proof}

\subsection{Theorem 1. Exact solution of contrastive learning}\label{s:sec:thm1}

\paragraph{Setting.} Linear heads $f=\Tz^\top R_z$ and $g=\Tw^\top R_w$ with $\Tz\in\R^{p_z\times d}$, $\Tw\in\R^{p_w\times d}$, and critic $s=f^\top g$. A positive pair is $(D_z,D_w)$ of one unit. A negative pair is $(D_z,D_w')$, where $D_w'$ is drawn from the law of $D_w$ given the same $X$, independently of $D_z$. Each view of each pair receives its own feature noise, giving $R_z+\xi_z$ and $R_w+\xi_w$ (or $R'_w+\xi'_w$), where the $\xi$'s are mutually independent, independent of everything else, mean zero, with covariance $\lambda I$. The loss is $\mathcal L_\lambda(\Tz,\Tw)=-2\,\E_{\rm pos}[s]+\E_{\rm neg}[s^2]$.

\paragraph{Empirical loss.} Given $m$ pool rows, let $r_{z,i},r_{w,i}$ be the residuals of the least-squares regression of the dictionaries on $[1,X]$ in the pool, and $\hat\Sigma_{zz},\hat\Sigma_{ww},\hat C$ their second moments ($\hat C=m^{-1}\sum_ir_{z,i}r_{w,i}^\top$). The empirical loss $\hat{\mathcal L}_\lambda$ draws positive pairs as $(r_{z,i},r_{w,i})$ with $i$ uniform on the pool and negative pairs as $(r_{z,i},r_{w,j})$ with $i,j$ independent and uniform, each view with independent $N(0,\lambda I)$ noise.

\begin{mt}{Theorem}{1}[Exact solution of contrastive learning]\label{s:thm:contrastive}
\begin{enumerate}
\item[(a)] Under (A1)--(A2), with $\tilde\Theta_z=(\Sigma_{zz}+\lambda I)^{1/2}\Tz$ and $\tilde\Theta_w=(\Sigma_{ww}+\lambda I)^{1/2}\Tw$,
\begin{align*}
\mathcal L_\lambda(\Tz,\Tw)&=-2\tr(\Tz^\top C\Tw)+\tr\big(\Tz^\top(\Sigma_{zz}+\lambda I)\Tz\,\Tw^\top(\Sigma_{ww}+\lambda I)\Tw\big)\\
&=\lVert K_\lambda-\tilde\Theta_z\tilde\Theta_w^\top\rVert_F^2-\lVert K_\lambda\rVert_F^2 .
\end{align*}
\item[(b)] $\min\mathcal L_\lambda=-\sum_{j\le d}\rho_j(\lambda)^2$, attained at $(\Theta_z^{(d)}(\lambda)S_d^{1/2},\Theta_w^{(d)}(\lambda)S_d^{1/2})$. If $\rho_d(\lambda)>\rho_{d+1}(\lambda)$, the set of minimizers is exactly
\[
\big\{\big(\Theta_z^{(d)}(\lambda)S_d^{1/2}M,\ \Theta_w^{(d)}(\lambda)S_d^{1/2}M^{-\top}\big):M\in GL(d)\big\},
\]
so the features of every minimizer are an invertible linear map of the top-$d$ ridge canonical variates. This holds for every $\lambda>0$ even when $\Sigma_{zz}$ or $\Sigma_{ww}$ is singular.
\item[(c)] Without any assumption, (a) and (b) hold for $\hat{\mathcal L}_\lambda$ with $(\hat\Sigma_{zz},\hat\Sigma_{ww},\hat C)$ in place of $(\Sigma_{zz},\Sigma_{ww},C)$. Hence the directions CPF computes, $\hat\Theta_z=\hat\Theta_z^{(d)}(\lambda)$ and $\hat\Theta_w=\hat\Theta_w^{(d)}(\lambda)$, rescaled by $\hat S_d^{1/2}$, are a global minimizer of $\hat{\mathcal L}_\lambda$, and CPF's standardized proxies are an invertible linear map of that minimizer's features, up to an affine function of $X$ (when $\hat\rho_d(\lambda)>0$).
\end{enumerate}
\end{mt}

\begin{proof}
\emph{Positive term.} Write $N=\Tz\Tw^\top$. Since the noises are independent, mean zero and independent of the data, $\E_{\rm pos}[s]=\E[(R_z+\xi_z)^\top N(R_w+\xi_w)]=\E[R_z^\top NR_w]=\tr(N^\top C)=\tr(\Tz^\top C\Tw)$.

\emph{Negative term.} Let $a=R_z+\xi_z$ and $b=R'_w+\xi'_w$, where $R'_w=D'_w-\E[D_w\mid X]$. By (A1) $R_z=D_z-\E[D_z\mid X]$, so $\E[a\mid X]=0$ and $\E[b\mid X]=0$, and by construction $a\Ind b\mid X$. Hence
\[
\E[s^2\mid X]=\E[\tr(N^\top aa^\top Nbb^\top)\mid X]=\tr\big(N^\top\E[aa^\top\mid X]\,N\,\E[bb^\top\mid X]\big).
\]
By (A2), $\E[aa^\top\mid X]=\Cov(D_z\mid X)+\lambda I=\Sigma_{zz}+\lambda I$ and $\E[bb^\top\mid X]=\Sigma_{ww}+\lambda I$, so $\E_{\rm neg}[s^2]=\tr(\Tz^\top(\Sigma_{zz}+\lambda I)\Tz\,\Tw^\top(\Sigma_{ww}+\lambda I)\Tw)$, which gives the first identity.

\emph{Frobenius form.} $\lVert K_\lambda-\tilde\Theta_z\tilde\Theta_w^\top\rVert_F^2=\lVert K_\lambda\rVert_F^2-2\tr(\tilde\Theta_z^\top K_\lambda\tilde\Theta_w)+\tr(\tilde\Theta_z^\top\tilde\Theta_z\tilde\Theta_w^\top\tilde\Theta_w)$. Substituting, $\tilde\Theta_z^\top K_\lambda\tilde\Theta_w=\Tz^\top C\Tw$ and $\tilde\Theta_z^\top\tilde\Theta_z=\Tz^\top(\Sigma_{zz}+\lambda I)\Tz$, $\tilde\Theta_w^\top\tilde\Theta_w=\Tw^\top(\Sigma_{ww}+\lambda I)\Tw$, which proves (a).

\emph{Minimization.} The matrices $(\Sigma+\lambda I)^{1/2}$ are invertible, so as $(\Tz,\Tw)$ ranges over all pairs, $\tilde\Theta_z\tilde\Theta_w^\top$ ranges over all matrices of rank $\le d$. By Lemma~\ref{s:lem:matrix}(i) the minimum is $\sum_{j>d}\rho_j(\lambda)^2-\sum_j\rho_j(\lambda)^2=-\sum_{j\le d}\rho_j(\lambda)^2$, attained at $\tilde\Theta_z\tilde\Theta_w^\top=\Gamma_dS_d\Omega_d^\top$, i.e.\ at the stated pair. If $\rho_d(\lambda)>\rho_{d+1}(\lambda)$, then $\rho_d(\lambda)>0$ and the minimizer is unique, so any minimizer satisfies $\tilde\Theta_z\tilde\Theta_w^\top=\Gamma_dS_d\Omega_d^\top$, a matrix of rank $d$. Then $\col\tilde\Theta_z\supseteq\col\Gamma_d$, and since $\tilde\Theta_z$ has $d$ columns, $\tilde\Theta_z=\Gamma_dS_d^{1/2}M$ for some $M\in GL(d)$. Substituting, $\Gamma_dS_d^{1/2}M\tilde\Theta_w^\top=\Gamma_dS_d\Omega_d^\top$. Multiplying by $\Gamma_d^\top$ on the left gives $\tilde\Theta_w=\Omega_dS_d^{1/2}M^{-\top}$. Conversely every such pair attains the minimum. Undoing the whitening proves (b). Only $\Sigma+\lambda I\succ0$ was used.

(c) For the empirical loss, $\E_{\rm pos}[s]=m^{-1}\sum_ir_{z,i}^\top Nr_{w,i}=\tr(\Tz^\top\hat C\Tw)$. For the negative term, $a=r_{z,i}+\xi_z$ and $b=r_{w,j}+\xi'_w$ are independent because $i$ and $j$ are, with $\E[aa^\top]=m^{-1}\sum_ir_{z,i}r_{z,i}^\top+\lambda I=\hat\Sigma_{zz}+\lambda I$ and likewise for $b$. Hence $\E[s^2]=\tr(N^\top(\hat\Sigma_{zz}+\lambda I)N(\hat\Sigma_{ww}+\lambda I))$. The rest of the proof is algebra and applies verbatim. CPF computes $\hat\Theta_z=(\hat\Sigma_{zz}+\lambda I)^{-1/2}\hat\Gamma_d$ and $\hat\Theta_w=(\hat\Sigma_{ww}+\lambda I)^{-1/2}\hat\Omega_d$ (the implementation adds $\lambda I$ to the pool second moments of the residuals), so $(\hat\Theta_z\hat S_d^{1/2},\hat\Theta_w\hat S_d^{1/2})$ is a minimizer for every SVD, ties included. The proxies are $\mathrm{diag}(\hat\sigma_z)^{-1}\hat\Theta_z^\top(D_z-\bar D_z)$, where $\bar D_z$ and $\hat\sigma_z$ are pool statistics. They differ from the minimizer's features $\hat S_d^{1/2}\hat\Theta_z^\top r_z$ by the invertible map $\mathrm{diag}(\hat\sigma_z)^{-1}\hat S_d^{-1/2}$ and by a term affine in $X$, and likewise on the outcome side.
\end{proof}

\begin{slemma}[Ridge limits and monotonicity]\label{s:lem:ridge}
(i) As $\lambda\to0$, $K_\lambda\to K_0$ when $\Sigma_{zz},\Sigma_{ww}\succ0$, so CPF reduces to $X$-conditional CCA. (ii) As $\lambda\to\infty$, $\lambda K_\lambda\to C$. Hence, if the $d$-th singular value of $C$ exceeds the $(d+1)$-th, $\col\Theta_z^{(d)}(\lambda)$ and $\col\Theta_w^{(d)}(\lambda)$ converge to the top-$d$ singular subspaces of $C$, i.e.\ partial least squares (PLS-SVD) of the conditional cross-covariance. (iii) Every $\rho_j(\lambda)$ is non-increasing in $\lambda\ge0$. The same holds for the sample quantities.
\end{slemma}
\begin{proof}
(i) is continuity of $\Sigma\mapsto(\Sigma+\lambda I)^{-1/2}$ on $\Sigma\succ0$. (ii) $(\Sigma+\lambda I)^{-1/2}=\lambda^{-1/2}(I+\Sigma/\lambda)^{-1/2}$, so $\lambda K_\lambda=(I+\Sigma_{zz}/\lambda)^{-1/2}C(I+\Sigma_{ww}/\lambda)^{-1/2}\to C$. Singular subspaces are continuous under a gap (Wedin's theorem, \citealp{wedin1972}), $\lambda K_\lambda$ and $K_\lambda$ have the same singular vectors, and $\Theta^{(d)}(\lambda)=\lambda^{-1/2}(I+\Sigma/\lambda)^{-1/2}\Gamma_d$ with $(I+\Sigma/\lambda)^{-1/2}\to I$. (iii) For $\lambda'>\lambda$, $K_{\lambda'}=F_z^\top K_\lambda F_w$ with $F_z=(\Sigma_{zz}+\lambda I)^{1/2}(\Sigma_{zz}+\lambda'I)^{-1/2}$ and $F_w$ likewise. These factors are functions of the same symmetric matrix, so they commute, and $F_z^\top F_z=(\Sigma_{zz}+\lambda'I)^{-1}(\Sigma_{zz}+\lambda I)\preceq I$. Thus $\lVert F_z\rVert,\lVert F_w\rVert\le1$, and Lemma~\ref{s:lem:matrix}(ii) gives $\rho_j(\lambda')\le\rho_j(\lambda)$. The argument uses only $\Sigma\succeq0$ and applies to sample moments.
\end{proof}

\subsection{Theorem 2. Exact inheritance of exclusion}\label{s:sec:thm2}

\begin{mt}{Theorem}{2}[Exact inheritance of exclusion]\label{s:thm:exclusion}
Let $\omega=(\mathfrak P,\varsigma)$ be the pool and seeds and let $a_\omega(m_z,x)$ and $b_\omega(m_w,x)$ be maps that are jointly measurable in $(\omega,m,x)$. Assume (P) and (S). Then, conditionally on $\omega$, for every analysis unit the proxies $\Zt=a_\omega(M_z,X)$ and $\Wt=b_\omega(M_w,X)$ satisfy (E1) and (E2). This applies to CPF's maps for every pool size and every outcome of the label-free selection. In contrast, there is a law satisfying (P) under which CCA fitted on the analysis sample itself violates (E2) for every unit.
\end{mt}

\begin{proof}
\emph{Fixed maps.} If $Q_1\Ind Q_2\mid Q_3$, then $h_1(Q_1,Q_3)\Ind h_2(Q_2,Q_3)\mid Q_3$ for measurable $h_1,h_2$ \citep[Lemma~4.2]{dawid1979}. For (E1), take $Q_1=M_z$, $Q_2=(T(0),T(1),C,M_w)$, $Q_3=(A,L,X)$, $h_1(m_z,(\alpha,l,x))=a(m_z,x)$ and $h_2((t_0,t_1,c,m_w),(\alpha,l,x))=(t_0,t_1,c,b(m_w,x))$. (P) gives the premise. For (E2), take $Q_1=M_w$, $Q_2=(A,M_z)$, $Q_3=(L,X)$, $h_1(m_w,(l,x))=b(m_w,x)$ and $h_2((\alpha,m_z),(l,x))=(\alpha,a(m_z,x))$.

\emph{Random maps.} Let $O$ collect the variables of one analysis unit. By (S), $O\Ind\omega$, so for every bounded measurable $\phi$, $\E[\phi(O,\omega)\mid\omega=\omega_0]=\E[\phi(O,\omega_0)]$ for almost every $\omega_0$. Hence, conditionally on $\omega=\omega_0$, the joint law of $(O,a_\omega(M_z,X),b_\omega(M_w,X))$ is that of $(O,a_{\omega_0}(M_z,X),b_{\omega_0}(M_w,X))$ with fixed maps, to which the first part applies.

\emph{CPF's maps.} Every step of Stage~1 is a measurable function of $\omega$ alone. Dropping one-hot columns with fewer than five occurrences or absences in the pool. The spectral features. Standardization with pool means and standard deviations. Residualization on $[1,X]$ with pool regression coefficients. The selection of $(B,\lambda)$, which uses a seeded permutation of the pool rows, held-out correlations computed on the pool, and an $\arg\max$ over a finite grid with a fixed tie-break, and the final ridge CCA refitted on the pool, with the pool means and standard deviations of the variates. Treatment, time and event never enter. The analysis unit enters only through its own $(M_z,M_w)$, so the first two parts apply.

\emph{In-sample fitting fails.} Let $M_z$ and $M_w$ be scalar, independent $N(0,1)$, and independent of $(L,X,A,T(0),T(1),C)$. (P) holds trivially. Fit one-dimensional CCA on an analysis sample of $n\ge2$ units. With sample standard deviations $\hat s_z,\hat s_w$ and sample covariance $\hat c_{zw}$, the variates are $\Zt_i=(M_{z,i}-\bar M_z)/\hat s_z$ and $\Wt_i=\operatorname{sgn}(\hat c_{zw})(M_{w,i}-\bar M_w)/\hat s_w$, the sign making the correlation positive. Then $n^{-1}\sum_j\Zt_j\Wt_j=|\hat r|$, the absolute sample correlation, which is positive almost surely. By exchangeability of the units, $\E[\Zt_i\Wt_i]=\E|\hat r|>0$. Replacing every $M_{z,j}$ by $-M_{z,j}$ preserves the joint law and maps $(\Zt_i,\Wt_i)$ to $(-\Zt_i,-\Wt_i)$, so $\E\Zt_i=\E\Wt_i=0$ (the variates are bounded by $\sqrt n$). If $\Zt_i\Ind\Wt_i$, then $\E[\Zt_i\Wt_i]=0$, a contradiction. Since $(\Zt_i,\Wt_i)$ is independent of $(L_i,X_i,A_i)$, it is also dependent conditionally on $(L_i,X_i)$, so (E2) fails for every unit.
\end{proof}

In a latent-factor example with $n=400$ and $p=60$, the partial correlation of the two proxies given $(L,X)$ is $+0.64$ when CCA is fitted on the analysis sample and $-0.02$ when it is fitted on a disjoint pool.

\subsection{Theorem 3. Information preservation and efficiency}\label{s:sec:thm3}

\subsubsection{Part (a). What the canonical variates contain}
\begin{slemma}[Canonical structure under (F)]\label{s:lem:structure}
Assume (F) and let $\lambda\ge0$.
\begin{enumerate}
\item[(i)] $C=\Lambda_z\Sigma_L\Lambda_w^\top$ has rank $k$, $\col C=\col\Lambda_z$ and $\col C^\top=\col\Lambda_w$. Hence $K_\lambda$ has exactly $k$ non-zero singular values.
\item[(ii)] $\col\Theta_z^{(k)}(\lambda)=\col\big((\Sigma_{zz}+\lambda I)^{-1}\Lambda_z\big)$ and $\col\Theta_w^{(k)}(\lambda)=\col\big((\Sigma_{ww}+\lambda I)^{-1}\Lambda_w\big)$.
\item[(iii)] $\Cov(\Lb,\Theta_z^{(k)}(\lambda)^\top R_z)=\Sigma_L\Lambda_z^\top(\Sigma_{zz}+\lambda I)^{-1}\Lambda_zN_\lambda$ for an invertible $N_\lambda$. It is non-singular for every $\lambda\ge0$, and likewise on the outcome side.
\end{enumerate}
\end{slemma}
\begin{proof}
(i) $\Sigma_L\Lambda_w^\top$ has full row rank $k$, so $\col C=\col\Lambda_z$, and symmetrically for $C^\top$. Multiplication by invertible matrices preserves rank. (ii) The left singular vectors with non-zero singular values form an orthonormal basis of $\col K_\lambda=(\Sigma_{zz}+\lambda I)^{-1/2}\col C=(\Sigma_{zz}+\lambda I)^{-1/2}\col\Lambda_z$, a space of dimension $k$. Applying $(\Sigma_{zz}+\lambda I)^{-1/2}$ maps it onto $\col((\Sigma_{zz}+\lambda I)^{-1}\Lambda_z)$. (iii) By (ii), $\Theta_z^{(k)}(\lambda)=(\Sigma_{zz}+\lambda I)^{-1}\Lambda_zN_\lambda$ with $N_\lambda$ invertible, and $\Cov(\Lb,R_z)=\Sigma_L\Lambda_z^\top$ because $\varepsilon_z\Ind\Lb$. The matrix $\Lambda_z^\top(\Sigma_{zz}+\lambda I)^{-1}\Lambda_z$ is positive definite because $\Lambda_z$ has full column rank.
\end{proof}

\begin{mt}{Theorem}{3(a)}[Information preservation]\label{s:thm:effa}
Under (F), the top-$k$ canonical variates satisfy $\Theta_z^{(k)}(0)^\top R_z=N^\top\hat L_z$ for an invertible $N$, where $\hat L_z=\proj(\Lb\mid R_z)=\Sigma_L\Lambda_z^\top\Sigma_{zz}^{-1}R_z=\E_{\rm lin}[\Lb\mid R_z]$. Consequently, under (A1), $\proj(\Lb\mid 1,X,D_z)\in\spn(1,X,\Zt_{1:k})$ for the CPF proxies $\Zt_{1:k}=\mathrm{diag}(\sigma)^{-1}\Theta_z^{(k)}(0)^\top(D_z-\bar D_z)$. The $k$ proxies retain all linear information about $\Lb$ beyond $X$ that the dictionary carries, whatever $\Psi_z$ is. The same holds on the outcome side.
\end{mt}
\begin{proof}
$\E[\Lb R_z^\top]=\Sigma_L\Lambda_z^\top$, so $\proj(\Lb\mid R_z)=\Sigma_L\Lambda_z^\top\Sigma_{zz}^{-1}R_z=\Upsilon^\top R_z$ with $\Upsilon=\Sigma_{zz}^{-1}\Lambda_z\Sigma_L$. By Lemma~\ref{s:lem:structure}(ii), $\col\Theta_z^{(k)}(0)=\col(\Sigma_{zz}^{-1}\Lambda_z)=\col\Upsilon$, and both matrices have $k$ independent columns, so $\Theta_z^{(k)}(0)=\Upsilon N$ with $N$ invertible. Under (A1), $D_z-\bar D_z=R_z+(\E[D_z\mid X]-\bar D_z)$, where the second term is affine in $X$ given the pool, so $\spn(1,X,\Zt_{1:k})=\spn(1,X,\hat L_z)$. Since $\Lb$ and $R_z$ are uncorrelated with every affine function of $X$, $\spn(1,X,D_z)=\spn(1,X)\oplus\spn(R_z)$ is an orthogonal sum and $\proj(\Lb\mid1,X,D_z)=\proj(\Lb\mid R_z)=\hat L_z$.
\end{proof}

\subsubsection{Part (b). The variance factorizes and CCA minimizes both factors}
\paragraph{Estimator.} For $P_z\in\R^{p_z\times d_z}$ and $P_w\in\R^{p_w\times k}$, let $\hat\tau(P_z,P_w)$ be the coefficient of $A$ in the two-stage least squares (2SLS) regression of $Y$ on $\mathsf D=(1,A,X,P_w^\top D_w)$ with instruments $\mathsf I=(1,A,X,P_z^\top D_z)$, from an i.i.d.\ analysis sample of size $n$. Call $P_w$ \emph{admissible} if $\Lambda_w^\top P_w$ is invertible. Define
\[
\begin{gathered}
\Zt=P_z^\top R_z,\qquad \Zt^\circ=\Zt-\proj(\Zt\mid1,X,A),\qquad H(P_z)=\M\{\proj(\Lb\mid\Zt^\circ)\},\\
\zeta(P_z)=\tfrac1{s_A}+\alpha^\top H(P_z)^{-1}\alpha,\qquad \vartheta_{P_w}=P_w(\Lambda_w^\top P_w)^{-1}\beta,\qquad \sigma_u^2(P_w)=\sigma_e^2+\vartheta_{P_w}^\top\Psi_w\vartheta_{P_w},
\end{gathered}
\]
where $\alpha$ is the coefficient of $A$ in $\proj(\Lb\mid1,X,A)$.

\begin{slemma}[Variance factorization]\label{s:lem:factor}
Assume (L1)--(L4), $P_w$ admissible and $H(P_z)\succ0$. Then $\hat\tau(P_z,P_w)$ is consistent and $\sqrt n(\hat\tau-\tau)\Rightarrow N(0,\mathcal V(P_z,P_w))$ with
\[
\mathcal V(P_z,P_w)=\sigma_u^2(P_w)\,\zeta(P_z).
\]
\end{slemma}
\begin{proof}
\emph{Reductions.} Replacing $P_z$ by a basis of $\col P_z$ does not change $\spn(\mathsf I)$, hence not the 2SLS fit, so assume $P_z$ has full column rank. Under (A1), $P_z^\top D_z=\Zt+(\text{affine in }X)$ and $P_w^\top D_w=\Wt+(\text{affine in }X)$ with $\Wt=P_w^\top R_w$. The instrument span is unchanged, and the regressors change by an invertible linear map that leaves the column of $A$ untouched, so the coefficient of $A$ is unchanged. We may therefore use $\Zt$ and $\Wt$.

\emph{(i) Structural equation.} With $\theta=(\Lambda_w^\top P_w)^{-1}\beta$, $\theta^\top\Wt=\vartheta_{P_w}^\top R_w=\vartheta_{P_w}^\top\Lambda_w\Lb+\vartheta_{P_w}^\top\varepsilon_w=\beta^\top\Lb+\vartheta_{P_w}^\top\varepsilon_w$, since $\Lambda_w^\top\vartheta_{P_w}=\beta$. By (L2), $Y=c_0+\tau A+\theta^\top\Wt+\gamma^\top X+u$ with $u=e-\vartheta_{P_w}^\top\varepsilon_w$.

\emph{(ii) Exogeneity and homoscedasticity.} $\mathsf I$ is a function of $(A,X,\Lb,\varepsilon_z)$ and $u$ is a function of $(e,\varepsilon_w)$, so $u\Ind\mathsf I$ by (L3). Thus $\E[u\mid\mathsf I]=0$ and $\E[u^2\mid\mathsf I]=\sigma_u^2(P_w)$.

\emph{(iii) 2SLS asymptotics.} $\M(\mathsf I)\succ0$. By (L4) $(1,X,A)$ has a non-singular second-moment matrix, and $\Var(\Zt^\circ)\succeq P_z^\top\Psi_zP_z\succ0$. If $\M(\hat{\mathsf D})\succ0$ with $\hat{\mathsf D}=\proj(\mathsf D\mid\mathsf I)$ (the rank condition, verified below), standard 2SLS theory \citep[Thms 5.1--5.2]{wooldridge2010} gives consistency and asymptotic normality with variance $\sigma_u^2\,\M(\hat{\mathsf D})^{-1}$. The central limit theorem applies because $\E[\lVert\mathsf I\rVert^2u^2]=\sigma_u^2\E\lVert\mathsf I\rVert^2<\infty$. By the partitioned-inverse (Frisch--Waugh--Lovell) formula, the $A$-entry of $\M(\hat{\mathsf D})^{-1}$ is $1/\E[\check A^2]$, where $\check A=A-\proj(A\mid1,X,\hat W)$ and $\hat W=\proj(\Wt\mid\mathsf I)$ (note $\proj(A\mid\mathsf I)=A$ and $\proj(X\mid\mathsf I)=X$).

\emph{(iv) The first stage.} $\varepsilon_w$ is independent of $\mathsf I$ and mean zero, so $\hat W=P_w^\top\Lambda_w\,G$ with $G=\proj(\Lb\mid\mathsf I)$. As $P_w^\top\Lambda_w$ is invertible, $\spn(1,X,\hat W)=\spn(1,X,G)$. Decompose $\spn(\mathsf I)=\spn(1,X,A)\oplus\spn(\Zt^\circ)$ (orthogonal). Then $G=\proj(\Lb\mid1,X,A)+h$ with $h=\proj(\Lb\mid\Zt^\circ)$, and $\proj(\Lb\mid1,X,A)=(\text{affine in }X)+\alpha A$.

\emph{(v) The residual variance.} Let $a=A-\proj(A\mid1,X)$, so $\E a^2=s_A$, and $v=\alpha a+h$. Since $h\perp\spn(1,X,A)$, $\spn(1,X,G)=\spn(1,X)\oplus\spn(v)$ and $\check A=a-\proj(a\mid v)$. Now $\M(v)=s_A\alpha\alpha^\top+H$ and $\E[av]=s_A\alpha$, where $H=H(P_z)$. By the Sherman--Morrison formula, with $q=\alpha^\top H^{-1}\alpha$,
\[
\E\check A^2=s_A-s_A^2\alpha^\top(H+s_A\alpha\alpha^\top)^{-1}\alpha=s_A-s_A^2\frac{q}{1+s_Aq}=\frac{s_A}{1+s_Aq}>0 .
\]
The Gram matrix of $(a,v)$ has Schur complement $H\succ0$ with respect to $a$, so $(1,X,A,\hat W)$ are linearly independent and the rank condition holds. Hence $\mathcal V=\sigma_u^2(1+s_Aq)/s_A=\sigma_u^2(P_w)\zeta(P_z)$.
\end{proof}

The outcome-side compression enters only through the error variance $\sigma_u^2(P_w)$, and the treatment-side compression only through the explained latent variance $H(P_z)$.

\begin{mt}{Theorem}{3(b)}[Efficiency]\label{s:thm:effb}
Assume (L1)--(L4).
\begin{enumerate}
\item[(i)] \emph{Treatment side.} For every $P_z$, $H(P_z)\preceq H(I_{p_z})$, so $\zeta(P_z)\ge\zeta(I_{p_z})$. Equality $H(P_z)=H(I_{p_z})$ holds whenever $\col P_z\supseteq\col(\Sigma_{zz}^{-1}\Lambda_z)=\col\Theta_z^{(k)}(0)$. Moreover $H(I_{p_z})\succ0$.
\item[(ii)] \emph{Outcome side.} For every admissible $P_w$, $\sigma_u^2(P_w)\ge\sigma_e^2+\beta^\top(\Lambda_w^\top\Psi_w^{-1}\Lambda_w)^{-1}\beta$. Equality holds for every $\beta$ if and only if $\col P_w=\col(\Sigma_{ww}^{-1}\Lambda_w)=\col\Theta_w^{(k)}(0)$.
\item[(iii)] \emph{Jointly.} For every $d\ge k$, CPF's population compression $(\Theta_z^{(d)}(0),\Theta_w^{(k)}(0))$ minimizes $\mathcal V$ over all $P_z$ and all admissible $P_w$, simultaneously for every outcome coefficient $\beta$, with minimum
\[
\mathcal V^\ast=\big(\sigma_e^2+\beta^\top(\Lambda_w^\top\Psi_w^{-1}\Lambda_w)^{-1}\beta\big)\big(\tfrac1{s_A}+\alpha^\top H(I_{p_z})^{-1}\alpha\big).
\]
Since the minimizer does not depend on $(\tau,\beta)$, the best label-free compression is the best compression that uses labels.
\end{enumerate}
\end{mt}
\begin{proof}
(i) \emph{Inequality.} $\spn(\Zt^\circ)\subseteq\spn(R_z^\circ)$, where $R_z^\circ=R_z-\proj(R_z\mid1,X,A)$. Lemma~\ref{s:lem:matrix}(iii) gives $H(P_z)\preceq H(I)$, and Lemma~\ref{s:lem:matrix}(iv) gives $\zeta(P_z)\ge\zeta(I)$ when $H(P_z)\succ0$ (and $\zeta(P_z)=\infty$ by convention otherwise).

\emph{Equality.} Let $\Theta_\ast=\Sigma_{zz}^{-1}\Lambda_z$, $S_z=\Theta_\ast^\top R_z$, and let $\mathcal N\in\R^{p_z\times(p_z-k)}$ be a basis of $\{x:\Lambda_z^\top x=0\}$. The matrix $[\Theta_\ast,\mathcal N]$ is invertible. If $\Theta_\ast c\in\col\mathcal N$ then $0=\Lambda_z^\top\Theta_\ast c=\Lambda_z^\top\Sigma_{zz}^{-1}\Lambda_zc$, so $c=0$. Hence $\spn(R_z)=\spn(S_z)+\spn(\mathcal N^\top R_z)$, with $\mathcal N^\top R_z=\mathcal N^\top\varepsilon_z$. By (L3), $\mathcal N^\top\varepsilon_z$ is independent of $(A,X,\Lb)$ and has mean zero, so it equals its own residual on $(1,X,A)$, and $\spn(R_z^\circ)=\spn(S_z^\circ)+\spn(\mathcal N^\top\varepsilon_z)$. Moreover $\mathcal N^\top\varepsilon_z\perp S_z^\circ$. $\E[\mathcal N^\top\varepsilon_zS_z^\top]=\mathcal N^\top\Psi_z\Theta_\ast=\mathcal N^\top\Sigma_{zz}\Theta_\ast=\mathcal N^\top\Lambda_z=0$ (using $\mathcal N^\top\Lambda_z=0$ twice), and $\mathcal N^\top\varepsilon_z$ is orthogonal to the function $\proj(S_z\mid1,X,A)$ of $(X,A)$. Also $\mathcal N^\top\varepsilon_z\perp\Lb$. Therefore $\Lb-\proj(\Lb\mid S_z^\circ)$ is orthogonal to $S_z^\circ$ and to $\mathcal N^\top\varepsilon_z$, hence to $\spn(R_z^\circ)$, so $\proj(\Lb\mid R_z^\circ)=\proj(\Lb\mid S_z^\circ)$. If $\col P_z\supseteq\col\Theta_\ast$, then $\spn(S_z^\circ)\subseteq\spn(\Zt^\circ)\subseteq\spn(R_z^\circ)$, and projecting onto the intermediate space gives the same result, so $H(P_z)=H(I)$.

\emph{Positivity.} $R_z^\circ=\Lambda_z\Lb^\circ+\varepsilon_z$ with $\Lb^\circ=\Lb-\proj(\Lb\mid1,X,A)$, so $\Cov(\Lb,R_z^\circ)=\Var(\Lb^\circ)\Lambda_z^\top$ and $H(I)=\Var(\Lb^\circ)\Lambda_z^\top\Var(R_z^\circ)^{-1}\Lambda_z\Var(\Lb^\circ)\succ0$ by (L4).

(ii) Minimize $f(\vartheta)=\vartheta^\top\Psi_w\vartheta$ subject to $\Lambda_w^\top\vartheta=\beta$. The problem is strictly convex. Its Lagrange conditions give the unique minimizer $\vartheta^\ast=\Psi_w^{-1}\Lambda_w(\Lambda_w^\top\Psi_w^{-1}\Lambda_w)^{-1}\beta$ with value $\beta^\top(\Lambda_w^\top\Psi_w^{-1}\Lambda_w)^{-1}\beta$. For admissible $P_w$, $\vartheta_{P_w}$ is the unique feasible point in $\col P_w$, so equality holds iff $\vartheta^\ast\in\col P_w$. As $\beta$ ranges over $\R^k$, $\vartheta^\ast$ ranges over the $k$-dimensional space $\col(\Psi_w^{-1}\Lambda_w)$, and $\col P_w$ has dimension $k$. So equality for every $\beta$ holds iff $\col P_w=\col(\Psi_w^{-1}\Lambda_w)$. By the Woodbury identity, with $\mathcal M=\Lambda_w^\top\Psi_w^{-1}\Lambda_w$,
\[
\Sigma_{ww}^{-1}\Lambda_w=\Psi_w^{-1}\Lambda_w-\Psi_w^{-1}\Lambda_w(\Sigma_L^{-1}+\mathcal M)^{-1}\mathcal M=\Psi_w^{-1}\Lambda_w(\Sigma_L^{-1}+\mathcal M)^{-1}\Sigma_L^{-1},
\]
so $\col(\Sigma_{ww}^{-1}\Lambda_w)=\col(\Psi_w^{-1}\Lambda_w)$, which equals $\col\Theta_w^{(k)}(0)$ by Lemma~\ref{s:lem:structure}(ii). This choice is admissible since $\Lambda_w^\top\Sigma_{ww}^{-1}\Lambda_w\succ0$.

(iii) By Lemma~\ref{s:lem:factor}, $\mathcal V$ is the product of a factor that depends only on $P_w$ and a factor that depends only on $P_z$, both positive. By (i), $\col\Theta_z^{(d)}(0)\supseteq\col\Theta_z^{(k)}(0)$ minimizes the second factor. By (ii), $\Theta_w^{(k)}(0)$ minimizes the first for every $\beta$.
\end{proof}

\begin{scor}[Consequences]\label{s:cor:effcons}
(i) If $d_z<k$, then $\rank H(P_z)\le d_z<k$, so $H(P_z)$ is singular and the 2SLS rank condition fails, so $d\ge k$ is necessary. (ii) In the regime of Proposition~\ref{s:prop:pca}, PCA proxies have $H(P_{\rm PCA})=0$, so $\mathcal V(\text{PCA})=\infty$ while $\mathcal V^\ast$ does not depend on the magnitude of style. PCA can be arbitrarily inefficient. (iii) With $d>k$ outcome-side directions, the extra $d-k$ directions are pure noise ($\Theta^\top\Lambda_w=0$ for the null canonical directions), so the linear bridge is not unique. A regularized bridge gives them zero weight asymptotically. On the treatment side extra directions are harmless by Theorem~3(b)(i).
\end{scor}
\begin{proof}
(i) $\proj(\Lb\mid\Zt^\circ)$ lies in the span of $d_z$ variables. (ii) is Proposition~\ref{s:prop:pca}(ii) with (i) of Theorem~3(b). (iii) A null outcome-side direction is $\theta=\Sigma_{ww}^{-1/2}\omega$ with $K_0\omega=0$, i.e.\ $C\theta=\Lambda_z\Sigma_L\Lambda_w^\top\theta=0$. Since $\Lambda_z\Sigma_L$ has full column rank, $\Lambda_w^\top\theta=0$ and $\theta^\top R_w=\theta^\top\varepsilon_w$.
\end{proof}

\subsubsection{Part (c). The cost of the ridge and of Stage-1 estimation is second order}

\begin{mt}{Theorem}{3(c)}[Second-order cost]\label{s:thm:effc}
Assume (L1)--(L4) and fix $\bar\lambda<\infty$.
\begin{enumerate}
\item[(i)] \emph{Population.} Let $P_z$ and $P_w$ have $k$ columns, with $\dist_{\Sigma^\circ}(\col P_z,\col\Theta_z^{(k)}(0))\le\theta$ and $\dist(\col P_w,\col\Theta_w^{(k)}(0))\le\theta$, where $\Sigma^\circ=\M(R_z^\circ)$. Then $0\le\mathcal V(P_z,P_w)-\mathcal V^\ast\le c\,\theta^2$ for all $\theta\le\theta_0$, with constants $c,\theta_0>0$ depending only on the model.
\item[(ii)] \emph{Ridge.} $0\le\mathcal V(\Theta_z^{(k)}(\lambda),\Theta_w^{(k)}(\lambda))-\mathcal V^\ast\le c\lambda^2$ for $\lambda\in[0,\bar\lambda]$. If $\Psi_z$ and $\Psi_w$ are multiples of the identity, the ridge costs nothing in population.
\item[(iii)] \emph{Stage-1 estimation.} Let the pool consist of $m$ i.i.d.\ rows with finite fourth moments, fixed dimensions, and be independent of the analysis sample. For the sample directions $\hat P_z(\lambda)=\hat\Theta_z^{(d)}(\lambda)$, $d\ge k$, and $\hat P_w(\lambda)=\hat\Theta_w^{(k)}(\lambda)$,
\[
0\le\mathcal V(\hat P_z(\lambda),\hat P_w(\lambda))-\mathcal V^\ast\le\hat c_m\,(\lambda+m^{-1/2})^2\quad\text{for all }\lambda\in[0,\bar\lambda],
\]
where $\hat c_m=O_p(1)$ does not depend on $\lambda$ and $\mathcal V$ is evaluated conditionally on the pool. In particular the bound holds at the label-free choice $\hat\lambda$.
\end{enumerate}
\end{mt}
\begin{proof}
\emph{Outcome-side factor.} For feasible $\vartheta$ ($\Lambda_w^\top\vartheta=\beta$), $f(\vartheta)-f(\vartheta^\ast)=(\vartheta-\vartheta^\ast)^\top\Psi_w(\vartheta-\vartheta^\ast)$, because the cross term $2(\vartheta-\vartheta^\ast)^\top\Psi_w\vartheta^\ast=2(\vartheta-\vartheta^\ast)^\top\Lambda_w\mathcal M^{-1}\beta$ vanishes. The map $P_w\mapsto\vartheta_{P_w}$ depends only on $\col P_w$ and is smooth on the set where $\Lambda_w^\top P_w$ is invertible. At an orthonormal basis $P_w^\ast$ of $\col\Theta_w^{(k)}(0)$ it is therefore Lipschitz on a neighbourhood. By Lemma~\ref{s:lem:subspace}(iv) there are orthonormal bases with $\lVert P_w-P_w^\ast\rVert\le\sqrt2\theta$, so $\lVert\vartheta_{P_w}-\vartheta^\ast\rVert\le c\theta$ and $\sigma_u^2(P_w)-\sigma_u^{2\ast}\le c\lVert\Psi_w\rVert\theta^2$ for $\theta\le\theta_0$.

\emph{Treatment-side factor.} Let $\ell=\proj(\Lb\mid R_z^\circ)$, which lies in $(\mathcal S^\ast)^k$ for $\mathcal S^\ast=\spn(\Theta_z^{(k)}(0)^\top R_z^\circ)$ by the proof of Theorem~3(b)(i), and let $\mathcal S=\spn(P_z^\top R_z^\circ)$. The $L^2$ inner product of $a^\top R_z^\circ$ and $b^\top R_z^\circ$ is $a^\top\Sigma^\circ b$, so $\dist(\mathcal S,\mathcal S^\ast)=\dist_{\Sigma^\circ}(\col P_z,\col\Theta_z^{(k)}(0))\le\theta$. Since $\Lb-\ell\perp\spn(R_z^\circ)\supseteq\mathcal S$, $\proj(\Lb\mid\mathcal S)=\proj(\ell\mid\mathcal S)$, and by Lemma~\ref{s:lem:matrix}(iii)
\[
H(I)-H(P_z)=\M\big(\ell-\proj(\ell\mid\mathcal S)\big),\qquad \lVert H(I)-H(P_z)\rVert\le\tr\M\big(\ell-\proj(\ell\mid\mathcal S)\big)\le\theta^2\tr H(I),
\]
using Lemma~\ref{s:lem:subspace}(i) coordinatewise. For $\theta^2\tr H(I)\le\lambda_{\min}(H(I))/2$, $\lVert H(P_z)^{-1}\rVert\le2\lVert H(I)^{-1}\rVert$, and Lemma~\ref{s:lem:matrix}(iv) gives $\zeta(P_z)-\zeta(I)\le2\lVert\alpha\rVert^2\lVert H(I)^{-1}\rVert^2\tr H(I)\,\theta^2$.

(i) $\mathcal V-\mathcal V^\ast=(\sigma_u^2(P_w)-\sigma_u^{2\ast})\zeta(P_z)+\sigma_u^{2\ast}(\zeta(P_z)-\zeta(I))$, and both terms are $O(\theta^2)$ with $\zeta(P_z)\le2\zeta(I)$ for small $\theta$. Non-negativity is Theorem~3(b).

(ii) $\lVert(\Sigma+\lambda I)^{-1}\Lambda-\Sigma^{-1}\Lambda\rVert=\lambda\lVert(\Sigma+\lambda I)^{-1}\Sigma^{-1}\Lambda\rVert\le\lambda\lVert\Sigma^{-1}\rVert^2\lVert\Lambda\rVert$, so by Lemma~\ref{s:lem:structure}(ii) and Lemma~\ref{s:lem:subspace}(iii),(ii) both distances are $O(\lambda)$, and (i) gives the bound for $\lambda\le\lambda_0$. For $\lambda\in[\lambda_0,\bar\lambda]$, the ridge directions are always admissible with $H\succ0$ (Lemma~\ref{s:lem:structure}(iii)), so $\mathcal V$ is continuous and bounded there and the bound holds with a larger $c$. If $\Psi=\sigma^2I$, then $\Sigma+\lambda I=\Lambda\Sigma_L\Lambda^\top+(\sigma^2+\lambda)I$ and the Woodbury identity gives $(\Sigma+\lambda I)^{-1}\Lambda=\Lambda\times(\text{invertible})$, so $\col((\Sigma+\lambda I)^{-1}\Lambda)=\col\Lambda=\col(\Sigma^{-1}\Lambda)$ for all $\lambda$.

(iii) \emph{Directions.} With finite fourth moments and fixed dimensions, the pool second moments of the residuals on $[1,X]$ satisfy $\lVert\hat\Sigma-\Sigma\rVert=O_p(m^{-1/2})$ for each block. On $\{\Sigma\succeq c_\Sigma I\}$, $\Sigma\mapsto(\Sigma+\lambda I)^{-1/2}$ is Lipschitz uniformly in $\lambda\ge0$, so $\sup_{\lambda\le\bar\lambda}\lVert\hat K_\lambda-K_\lambda\rVert=O_p(m^{-1/2})$. By Lemma~\ref{s:lem:structure}(i), $\rho_{k+1}(\lambda)=0$, and $\lambda\mapsto\rho_k(\lambda)$ is continuous and positive on $[0,\bar\lambda]$, hence bounded below by some $\rho_\ast>0$. Wedin's theorem \citep{wedin1972} with Weyl's inequality bounds the distance between the top-$k$ singular subspaces by $\lVert\hat K_\lambda-K_\lambda\rVert/(\rho_\ast-\lVert\hat K_\lambda-K_\lambda\rVert)$. With aligned orthonormal bases (Lemma~\ref{s:lem:subspace}(iv)), $\lVert\hat\Theta^{(k)}(\lambda)-\Theta^{(k)}(\lambda)O\rVert\le\lVert(\hat\Sigma+\lambda I)^{-1/2}-(\Sigma+\lambda I)^{-1/2}\rVert+\lVert(\Sigma+\lambda I)^{-1/2}\rVert\sqrt2\,\dist(\hat\Gamma_k,\Gamma_k)=O_p(m^{-1/2})$ uniformly in $\lambda$, and $\sigma_{\min}(\Theta^{(k)}(\lambda))\ge\lambda_{\max}(\Sigma+\bar\lambda I)^{-1/2}>0$, so Lemma~\ref{s:lem:subspace}(iii),(ii) give $\dist(\col\hat\Theta^{(k)}(\lambda),\col\Theta^{(k)}(\lambda))=O_p(m^{-1/2})$ uniformly in $\lambda$, in both metrics.

\emph{Variance.} By the triangle inequality for $\dist$, the sample top-$k$ directions are within $O(\lambda)+O_p(m^{-1/2})$ of the CCA directions. Conditionally on the pool, the maps are fixed and independent of the analysis sample, so Lemma~\ref{s:lem:factor} applies with $(\hat P_z,\hat P_w)$. Admissibility and $H\succ0$ hold with probability tending to one. Part (i) bounds $\mathcal V(\hat\Theta_z^{(k)},\hat P_w)-\mathcal V^\ast$, and adding the remaining $d-k$ treatment-side directions can only lower $\mathcal V$ by Theorem~3(b)(i) and Lemma~\ref{s:lem:matrix}(iii),(iv). On the event where the distances exceed $\theta_0$, whose probability tends to zero, the bound is absorbed into $\hat c_m$.
\end{proof}

\subsection{Theorem 4. Identification without linearity}\label{s:sec:thm4nl}

Theorems 1 and 3 use (A1), (A2) and the linear factor model (F). This section drops all three and keeps the proximal condition (P), positivity and, where stated, completeness of the raw modalities. Every statement is conditional on $X=x$ and holds for almost every $x$. To lighten notation we suppress $x$, so $P$ and $\E$ denote the conditional law and expectation given $X=x$. The variables $L$, $M_z$ and $M_w$ take values in standard Borel spaces.

\paragraph{Operators.} Let $H_z=L^2(P_{M_z})$, $H_w=L^2(P_{M_w})$ and $H_L=L^2(P_L)$, and let a superscript $\circ$ denote the subspace of mean-zero functions. Define
\[
\mathcal S_z\phi=\E[\phi(L)\mid M_z],\qquad \mathcal S_w\phi=\E[\phi(L)\mid M_w],\qquad \mathcal Tg=\E[g(M_w)\mid M_z],
\]
mapping $H_L\to H_z$, $H_L\to H_w$ and $H_w\to H_z$. Their adjoints are $\mathcal S_z^\ast f=\E[f(M_z)\mid L]$, $\mathcal S_w^\ast g=\E[g(M_w)\mid L]$ and $\mathcal T^\ast f=\E[f(M_z)\mid M_w]$, because, for instance, $\E[\E[\phi(L)\mid M_z]f(M_z)]=\E[\phi(L)f(M_z)]=\E[\phi(L)\E[f(M_z)\mid L]]$. All are contractions (Jensen), map each constant to itself and map mean-zero functions to mean-zero functions. A superscript $\circ$ on an operator denotes its restriction to mean-zero functions.

\paragraph{Nonlinear canonical analysis.} For $f\in H_z^\circ$ and $g\in H_w^\circ$ of unit variance, $\Corr(f(M_z),g(M_w))=\E[f(M_z)\E[g(M_w)\mid M_z]]=\langle f,\mathcal T^\circ g\rangle$, so $\lVert\mathcal T^\circ\rVert$ is the maximal correlation of \citet{renyi1959}. When $\mathcal T^\circ$ is compact it has a singular value decomposition
\[
\mathcal T^\circ=\sum_{j\le r}\sigma_j\,f_j\otimes g_j,\qquad (f\otimes g)h=\langle g,h\rangle f,\qquad \sigma_1\ge\sigma_2\ge\dots>0,\quad r\le\infty,
\]
with orthonormal $\{f_j\}\subset H_z^\circ$ and $\{g_j\}\subset H_w^\circ$, and then $\Corr(f_j(M_z),g_k(M_w))=\sigma_j\delta_{jk}$. The $f_j,g_j$ are the \emph{nonlinear canonical functions} and the $\sigma_j$ the \emph{nonlinear canonical correlations} \citep{lancaster1958,buja1990}. Here they are $X$-conditional, since $x$ is fixed. $\mathcal T^\circ$ is Hilbert--Schmidt (hence compact) whenever the joint law of $(M_z,M_w)$ has a density with respect to $P_{M_z}\otimes P_{M_w}$ that is square-integrable under the product law, and it has finite rank whenever $L$ takes finitely many values (by part (a) below, $\mathcal T$ factors through $H_L$).

\paragraph{Completeness conditions on the raw modalities.} (Cz) $M_z$ is $L^2$-complete for $L$. If $\phi\in H_L$ and $\E[\phi(L)\mid M_z]=0$ a.s., then $\phi(L)=0$ a.s.. Equivalently, $\mathcal S_z$ is injective. (Cw) is the same for $M_w$. When $L$ takes $K$ values $l_1,\dots,l_K$ with probabilities $p_k>0$, $\mathcal S_z\phi=\sum_k\phi(l_k)P(L=l_k\mid M_z)$ and $P(L=l_k\mid M_z)=p_k\,dP(M_z\in\cdot\mid l_k)/dP_{M_z}$, so (Cz) holds iff the $K$ conditional laws $P(M_z\in\cdot\mid L=l_k)$ are linearly independent, which is the rank condition for categorical proxies of \citet{miao2018}.

\begin{mt}{Theorem}{4}[Identification without linearity]\label{s:thm:nonlinear}
Assume (P) and fix $x$.
\begin{enumerate}
\item[(a)] \emph{Factorization.} $\mathcal T=\mathcal S_z\mathcal S_w^\ast$. Hence, if $\mathcal T^\circ$ is compact, every canonical pair consists of posterior means of the confounder
\[
f_j(M_z)=\E[\phi_j(L)\mid M_z],\qquad g_j(M_w)=\E[\psi_j(L)\mid M_w],
\]
where $\phi_j=\E[g_j(M_w)\mid L]/\sigma_j$ and $\psi_j=\E[f_j(M_z)\mid L]/\sigma_j$ lie in $H_L^\circ$.
\item[(b)] \emph{Style invariance.} Suppose $M_z=h(Y,S)$ for an injective Borel map $h$, with $S\Ind(Y,M_w)$. Then the nonlinear canonical correlations of $(M_z,M_w)$ are those of $(Y,M_w)$, the $g_j$ are the same, and $f_j(h(y,s))=f_j^Y(y)$ for every $s$, where $f_j^Y$ are the canonical functions of $(Y,M_w)$. Neither the law of $S$ nor $h$ affects the canonical analysis. The same holds on the outcome side.
\item[(c)] \emph{Contrastive learning with nonlinear heads.} Let $\mathcal T^\circ$ be Hilbert--Schmidt. For heads $u=(u_1,\dots,u_d)\in(H_z^\circ)^d$ and $v\in(H_w^\circ)^d$ and critic $s=u(M_z)^\top v(M_w)$, the $X$-conditional spectral contrastive loss of Section~\ref{s:sec:thm1}, with general heads and $\lambda=0$, equals
\[
\mathcal L(u,v)=-2\,\E_{\rm pos}[s]+\E_{\rm neg}[s^2]=\Big\lVert\mathcal T^\circ-\sum_{j\le d}u_j\otimes v_j\Big\rVert_{\rm HS}^2-\lVert\mathcal T^\circ\rVert_{\rm HS}^2 .
\]
Its minimum is $-\sum_{j\le d}\sigma_j^2$ (with $\sigma_j=0$ for $j>r$), attained at $u_j=\sigma_j^{1/2}f_j$, $v_j=\sigma_j^{1/2}g_j$. If $\sigma_d>\sigma_{d+1}$, the minimizers are exactly $u=M^\top(\sigma_j^{1/2}f_j)_{j\le d}$, $v=M^{-1}(\sigma_j^{1/2}g_j)_{j\le d}$ with $M\in GL(d)$, so the learned features are an invertible linear map of the top-$d$ canonical functions.
\item[(d)] \emph{Dictionary limit.} Let $\mathcal D^z_1\subseteq\mathcal D^z_2\subseteq\dots\subseteq H_z^\circ$ and $\mathcal D^w_1\subseteq\mathcal D^w_2\subseteq\dots\subseteq H_w^\circ$ be finite-dimensional subspaces whose unions are dense (the conditionally centred spans of growing dictionaries), and let $\rho_j(K)$ be the linear canonical correlations between the $K$-th dictionaries. If $\mathcal T^\circ$ is compact, then $\rho_j(K)$ is non-decreasing in $K$ and converges to $\sigma_j$ for every $j$, and if $\sigma_d>\sigma_{d+1}$ the top-$d$ linear canonical subspaces converge, in the gap metric, to $\spn\{f_1,\dots,f_d\}$ and $\spn\{g_1,\dots,g_d\}$. If (A1)--(A2) hold for every dictionary, CPF's population CCA ($\lambda=0$) computes these linear canonical correlations and directions at every $x$.
\item[(e)] \emph{Sufficiency.} If $\mathcal T^\circ$ is compact and (Cz) holds, then $\Wt_r=(g_j(M_w))_{j\le r}$ is sufficient for $L$ in $M_w$. $P(L\in\cdot\mid M_w)=P(L\in\cdot\mid\Wt_r)$ a.s. Symmetrically, $\Zt_r=(f_j(M_z))_{j\le r}$ is sufficient for $L$ in $M_z$ under (Cw).
\item[(f)] \emph{Completeness.} Assume (Pos), that $\mathcal T^\circ$ is compact, (Cz), and that $M_w$ is complete for $L$ (Definition~\ref{s:def:complete}). Then every proxy of the form $\Wt=(\Wt_r,v(M_w))$, with $v$ measurable, is complete for $L$. Symmetrically, $\Zt=(\Zt_r,u(M_z))$ is complete for $L$ under (Cw), completeness of $M_z$ for $L$, and $M_z\Ind A\mid L,X$.
\item[(g)] \emph{Finitely many states.} If $L$ takes $K$ values with positive probabilities and (Cz) and (Cw) hold, then $r=K-1$, and $(g_1,\dots,g_{K-1})(M_w)$ is an invertible affine function of the posterior probabilities $(P(L=l_k\mid M_w))_{k<K}$. Symmetrically for $(f_1,\dots,f_{K-1})(M_z)$. Hence $d\ge K-1$ proxies per modality suffice for (e) and (f).
\end{enumerate}
\end{mt}

\begin{proof}
(a) For $g\in H_w$, the tower property and $M_w\Ind M_z\mid L$, which (P) implies, give $\E[g(M_w)\mid M_z]=\E[\E[g(M_w)\mid M_z,L]\mid M_z]=\E[\E[g(M_w)\mid L]\mid M_z]=\mathcal S_z\mathcal S_w^\ast g$. The same computation on mean-zero functions gives $\mathcal T^\circ=\mathcal S_z^\circ\mathcal S_w^{\circ\ast}$ and $\mathcal T^{\circ\ast}=\mathcal S_w^\circ\mathcal S_z^{\circ\ast}$. From the SVD, $\mathcal T^\circ g_j=\sigma_jf_j$, so $f_j=\mathcal S_z(\mathcal S_w^\ast g_j)/\sigma_j=\E[\phi_j(L)\mid M_z]$. Similarly $\mathcal T^{\circ\ast}f_j=\sigma_jg_j$ gives $g_j=\E[\psi_j(L)\mid M_w]$. Since $g_j$ has mean zero, so has $\phi_j$, and likewise $\psi_j$.

(b) Because $h$ is an injective Borel map between standard Borel spaces, its image is Borel and its inverse is Borel \citep[Thm~15.1]{kechris1995}, so $\sigma(M_z)=\sigma(Y,S)$ and we may take $M_z=(Y,S)$. For $g\in H_w$ and Borel sets $B,C$, independence of $S$ and $(Y,M_w)$ gives
\begin{align*}
\E[g(M_w)1\{Y\in B\}1\{S\in C\}]&=\E[g(M_w)1\{Y\in B\}]\,P(S\in C)\\
&=\E\big[\E[g(M_w)\mid Y]1\{Y\in B\}\big]\,P(S\in C)\\
&=\E\big[\E[g(M_w)\mid Y]1\{Y\in B\}1\{S\in C\}\big],
\end{align*}
and a $\pi$--$\lambda$ argument extends this to all of $\sigma(Y,S)$, so $\E[g(M_w)\mid Y,S]=\E[g(M_w)\mid Y]$. Hence $\mathcal T^\circ=J\mathcal T_Y^\circ$, where $\mathcal T_Y^\circ g=\E[g(M_w)\mid Y]$ and $J:L^2(P_Y)\to L^2(P_{Y,S})$, $Ju(y,s)=u(y)$, is an isometry. Then $\mathcal T^{\circ\ast}\mathcal T^\circ=\mathcal T_Y^{\circ\ast}J^\ast J\mathcal T_Y^\circ=\mathcal T_Y^{\circ\ast}\mathcal T_Y^\circ$, so the singular values and the right singular functions coincide, and $f_j=\mathcal T^\circ g_j/\sigma_j=J(\mathcal T^\circ_Yg_j)/\sigma_j=Jf_j^Y$. The outcome side is symmetric.

(c) Write $M_w'$ for a draw from $P_{M_w}$ independent of $M_z$, which is the $X$-conditional negative at the same $x$. Then $\E_{\rm pos}[s]=\sum_j\E[u_j(M_z)v_j(M_w)]=\sum_j\langle u_j,\mathcal T^\circ v_j\rangle$ and $\E_{\rm neg}[s^2]=\sum_{i,j}\E[u_iu_j(M_z)]\E[v_iv_j(M_w')]=\sum_{i,j}\langle u_i,u_j\rangle\langle v_i,v_j\rangle$. Since $\langle\mathcal T^\circ,u\otimes v\rangle_{\rm HS}=\langle u,\mathcal T^\circ v\rangle$ and $\langle u_i\otimes v_i,u_j\otimes v_j\rangle_{\rm HS}=\langle u_i,u_j\rangle\langle v_i,v_j\rangle$, expanding the Hilbert--Schmidt norm gives the identity. As $(u,v)$ ranges over all heads, $\sum_ju_j\otimes v_j$ ranges over all operators of rank at most $d$. By Schmidt's approximation theorem, the Hilbert--Schmidt version of Lemma~\ref{s:lem:matrix}(i) \citep{stewart1993}, the best rank-$d$ approximation has error $\sum_{j>d}\sigma_j^2$ and is the truncated SVD, uniquely when $\sigma_d>\sigma_{d+1}$. This gives the minimum. For the minimizer set, write $\sum_ju_j\otimes v_j=UV^\ast$ with $U:\R^d\to H_z^\circ$, $Ue_j=u_j$, and $V$ likewise, and $F_d$, $G_d$ for the maps $e_j\mapsto f_j$, $e_j\mapsto g_j$. Then $UV^\ast=F_dS_dG_d^\ast$, an operator of rank $d$ (since $\sigma_d>0$). Hence $\operatorname{range}U\supseteq\spn\{f_1,\dots,f_d\}$, so $U=F_dS_d^{1/2}M$ with $M$ invertible, and applying $F_d^\ast$ gives $V=G_dS_d^{1/2}M^{-\top}$, exactly as in the proof of Theorem~\ref{s:thm:contrastive}.

(d) For $g\in\mathcal D^w_K\subseteq H_w^\circ$, the $L^2$ projection of $g(M_w)$ onto $H_z^\circ$ is $\E[g(M_w)\mid M_z]=\mathcal T^\circ g$, so the projection onto $\mathcal D^z_K\subseteq H_z^\circ$ is $\Pi^z_K\mathcal T^\circ g$. Therefore the linear canonical correlations of the $K$-th dictionaries, defined as in Section~\ref{s:sec:thm4} as the singular values of the projection between the two spans, are the singular values of $A_K=\Pi^z_K\mathcal T^\circ\Pi^w_K$. Monotonicity in $K$ is Theorem~\ref{s:thm:monotone}(a). Because the subspaces are nested with dense unions, $\Pi^z_K\to I$ and $\Pi^w_K\to I$ strongly. Thus
\[
\lVert A_K-\mathcal T^\circ\rVert\le\lVert(\Pi^z_K-I)\mathcal T^\circ\rVert+\lVert\mathcal T^\circ(\Pi^w_K-I)\rVert=\lVert(\Pi^z_K-I)\mathcal T^\circ\rVert+\lVert(\Pi^w_K-I)\mathcal T^{\circ\ast}\rVert\to0,
\]
because a uniformly bounded, strongly convergent sequence of operators converges uniformly on the precompact images of the unit ball under the compact operators $\mathcal T^\circ$ and $\mathcal T^{\circ\ast}$. Weyl's inequality $|\sigma_j(A_K)-\sigma_j(\mathcal T^\circ)|\le\lVert A_K-\mathcal T^\circ\rVert$, from the min--max characterization, gives $\rho_j(K)\to\sigma_j$. For the subspaces, apply the $\sin\Theta$ theorem of \citet{davis1970} to the compact self-adjoint operators $A_KA_K^\ast$ and $\mathcal T^\circ\mathcal T^{\circ\ast}$, whose top-$d$ eigenspaces are the left canonical subspaces. Their eigengap is $\sigma_d^2-\sigma_{d+1}^2>0$, and $\lVert A_KA_K^\ast-\mathcal T^\circ\mathcal T^{\circ\ast}\rVert\le2\lVert A_K-\mathcal T^\circ\rVert\to0$. The right side is the same with $A_K^\ast A_K$. Under (A1)--(A2) for a dictionary, the second moments of its residuals on $(1,X)$ equal its conditional covariances at every $x$, which do not depend on $x$. The linear CCA computed from the residuals is therefore the conditional linear CCA at every $x$.

(e) By (a), $\mathcal T^{\circ\ast}=\mathcal S_w^\circ\mathcal S_z^{\circ\ast}$. (Cz) makes $\mathcal S_z^\circ$ injective, so $\operatorname{range}\mathcal S_z^{\circ\ast}$ is dense in $H_L^\circ$ (its closure is $(\ker\mathcal S_z^\circ)^\perp$). As $\mathcal S_w^\circ$ is bounded,
\[
\mathcal S_w^\circ(H_L^\circ)\subseteq\overline{\mathcal S_w^\circ(\operatorname{range}\mathcal S_z^{\circ\ast})}=\overline{\operatorname{range}\mathcal T^{\circ\ast}}=\overline{\spn}\{g_j:j\le r\}.
\]
So for every $\phi\in H_L$, $\E[\phi(L)\mid M_w]=\E\phi+\mathcal S_w^\circ(\phi-\E\phi)$ is an $L^2$ limit of linear combinations of the $g_j(M_w)$. Along a subsequence the limit is almost sure, so $\E[\phi(L)\mid M_w]$ has a $\sigma(\Wt_r)$-measurable version. Taking $\phi=1_B$ for $B$ in a countable $\pi$-system generating the Borel sets of the $L$-space, $P(L\in B\mid M_w)$ is $\sigma(\Wt_r)$-measurable, hence equals $P(L\in B\mid\Wt_r)$ a.s., because $\sigma(\Wt_r)\subseteq\sigma(M_w)$. Two probability kernels that agree on a generating $\pi$-system agree, so $P(L\in\cdot\mid M_w)=P(L\in\cdot\mid\Wt_r)$ a.s.

(f) Fix $a$ and let $g$ satisfy $\E|g(L)|<\infty$ and $\E[g(L)\mid\Wt,A=a]=0$ a.s. Write $\pi_a(l)=P(A=a\mid L=l)$, which is positive by (Pos). By (P), $M_w\Ind A\mid L$, so Bayes' rule gives
\[
\E[g(L)\mid M_w,A=a]=\frac{\int g\,\pi_a\,dP(L\in\cdot\mid M_w)}{\int\pi_a\,dP(L\in\cdot\mid M_w)}=:\Phi(\Wt_r),
\]
where the second equality and the $\sigma(\Wt_r)$-measurability follow from (e). As $\sigma(\Wt_r)\subseteq\sigma(\Wt)\subseteq\sigma(M_w)$, the tower property gives $\E[g(L)\mid\Wt,A=a]=\E[\E[g(L)\mid M_w,A=a]\mid\Wt,A=a]=\Phi(\Wt_r)$. Hence $\E[g(L)\mid M_w,A=a]=0$ a.s., and completeness of $M_w$ gives $g(L)=0$ a.s. given $A=a$. The treatment side is identical once $M_z\Ind A\mid L,X$ is assumed, which gives the same Bayes form.

(g) $H_L$ has dimension $K$ and $H_L^\circ$ dimension $K-1$. (Cw) makes $\mathcal S_w^\circ$ injective, so $\mathcal S_w^{\circ\ast}$ maps onto $H_L^\circ$. (Cz) makes $\mathcal S_z^\circ$ injective, so $\operatorname{rank}\mathcal T^\circ=\dim\mathcal S_z^\circ(H_L^\circ)=K-1$. Also $\spn\{g_1,\dots,g_{K-1}\}=\operatorname{range}\mathcal T^{\circ\ast}=\mathcal S_w^\circ(\operatorname{range}\mathcal S_z^{\circ\ast})=\mathcal S_w^\circ(H_L^\circ)$, because $\mathcal S_z^{\circ\ast}$ maps onto $H_L^\circ$ by (Cz). This space is spanned by $\mathcal S_w^\circ(1\{L=l_k\}-p_k)=P(L=l_k\mid M_w)-p_k$, $k=1,\dots,K$. These $K$ functions sum to zero and span a space of dimension $K-1$, so the first $K-1$ of them form a basis, and $(g_1,\dots,g_{K-1})(M_w)$ is an invertible linear function of $(P(L=l_k\mid M_w)-p_k)_{k<K}$. The treatment side is symmetric.
\end{proof}

\subsection{Proposition 2. Failure of generic compression}\label{s:sec:prop1}

\begin{mt}{Proposition}{2}[Failure of generic compression]\label{s:prop:pca}
Assume (F) for $D_z$ with $\Psi_z=\sigma^2I+\sigma_s^2\Pi_{\mathcal S}$, where $\sigma^2>0$, $\sigma_s\ge0$, and the style subspace $\mathcal S\perp\col\Lambda_z$ has dimension $m_s\ge d$.
\begin{enumerate}
\item[(i)] The canonical correlations $\rho_j$ and the canonical directions on both sides do not depend on $\sigma_s$.
\item[(ii)] If $\sigma_s^2>\lambda_{\max}(\Lambda_z\Sigma_L\Lambda_z^\top)$, every top-$d$ principal subspace of $\Sigma_{zz}$ lies in $\mathcal S$, and the PCA-$d$ proxies $\Zt_{\rm PCA}=V_d^\top(D_z-\bar D_z)$ satisfy $\Cov(\Lb,\Zt_{\rm PCA}\mid X)=0$. In fact $\Zt_{\rm PCA}-\E[\Zt_{\rm PCA}\mid X]=V_d^\top\varepsilon_z$ is independent of $(\Lb,X)$. If in addition $\varepsilon_z\Ind A\mid L,X$ (automatic on the outcome side by (P)), $\Zt_{\rm PCA}$ is not complete for $L$ unless $L$ is almost surely a function of $(A,X)$.
\item[(iii)] Let a linear autoencoder with $h\le m_s$ code units minimize $\E\lVert R_z-\mathcal B\mathcal AR_z\rVert^2$ over $\mathcal A\in\R^{h\times p_z}$ and $\mathcal B\in\R^{p_z\times h}$. Its code is $\mathcal AR_z$. Under the condition of (ii), every global minimizer has $\spn(\mathcal AR_z)=\spn(V^\top R_z)$ for an $h$-dimensional $\col V\subseteq\mathcal S$, so its code, and every linear function of it, is uncorrelated with $\Lb$. (Every other critical point is a saddle, \citealp{baldi1989}.)
\end{enumerate}
\end{mt}

\begin{proof}
(i) For $x\in\mathcal S$, $\Lambda_z^\top x=0$, so $\Sigma_{zz}x=(\sigma^2+\sigma_s^2)x$. $\mathcal S$ is an eigenspace of $\Sigma_{zz}$ and $\mathcal S^\perp$ is invariant, with $\Sigma_{zz}|_{\mathcal S^\perp}=(\Lambda_z\Sigma_L\Lambda_z^\top+\sigma^2I)|_{\mathcal S^\perp}$, free of $\sigma_s$. Since $\col C=\col\Lambda_z\subseteq\mathcal S^\perp$, both $\Sigma_{zz}^{-1}C$ and $C^\top\Sigma_{zz}^{-1}C$ are free of $\sigma_s$. Hence $K_0^\top K_0=\Sigma_{ww}^{-1/2}C^\top\Sigma_{zz}^{-1}C\Sigma_{ww}^{-1/2}$, its eigenvalues $\rho_j^2$ and its eigenvectors $\Omega$ are free of $\sigma_s$, and so are $\Theta_w=\Sigma_{ww}^{-1/2}\Omega$ and $\Theta_z=\Sigma_{zz}^{-1}C\Theta_w\,\mathrm{diag}(\rho)^{-1}$ for the non-null pairs.

(ii) The eigenvalues of $\Sigma_{zz}$ on $\mathcal S^\perp$ are at most $\sigma^2+\lambda_{\max}(\Lambda_z\Sigma_L\Lambda_z^\top)<\sigma^2+\sigma_s^2$, so the top-$m_s$ eigenspace is exactly $\mathcal S$, and since $d\le m_s$ every top-$d$ eigenspace lies in it. For $v\in\mathcal S$, $v^\top R_z=v^\top\varepsilon_z$, which is independent of $(\Lb,X)$ by (F). Under (A1), $v^\top(D_z-\bar D_z)=v^\top\varepsilon_z+(\text{affine in }X)$. For completeness, $\varepsilon_z\Ind\Lb\mid X$ (from (F)) and $\varepsilon_z\Ind A\mid L,X$ give $\varepsilon_z\Ind(L,A)\mid X$, so $\E[g(L)\mid\Zt_{\rm PCA},A=a,X=x]=\E[g(L)\mid A=a,X=x]$. Take $g_{a,x}(l)=c^\top(l-\E[L\mid A=a,X=x])$ with $c$ such that $\Var(c^\top L\mid A=a,X=x)>0$, which exists unless $L$ is a.s.\ a function of $(A,X)$. Then $\E[g(L)\mid\Zt_{\rm PCA},A,X]=0$ although $g(L)\ne0$.

(iii) The reconstruction error is $\tr((I-\mathcal B\mathcal A)\Sigma_{zz}(I-\mathcal B\mathcal A)^\top)=\lVert\Sigma_{zz}^{1/2}-\mathcal B\mathcal A\Sigma_{zz}^{1/2}\rVert_F^2$, and $\mathcal B\mathcal A\Sigma_{zz}^{1/2}$ ranges over all matrices of rank $\le h$. By Lemma~\ref{s:lem:matrix}(i), the minimizers are the truncated eigendecompositions $\Pi_{\mathcal U}\Sigma_{zz}^{1/2}$, where $\mathcal U$ is spanned by $h$ orthonormal eigenvectors with the $h$ largest eigenvalues. Here $\mathcal U\subseteq\mathcal S$ by (ii). Since $\Sigma_{zz}^{1/2}$ is invertible, $\mathcal B\mathcal A=\Pi_{\mathcal U}$, so the row space of $\mathcal A$, which has $h$ rows, equals $\mathcal U$. The code therefore spans $\{u^\top R_z:u\in\mathcal U\}$, which is uncorrelated with $\Lb$ by (ii).
\end{proof}

\subsection{Theorem 5. Dictionary monotonicity}\label{s:sec:thm4}

\paragraph{Definitions.} Let $\mathcal H_z(D)$ be the linear span of the coordinates of $R(D_z)$ in $L^2$, and $\mathcal H_w(D)$ likewise. The canonical correlations $\rho_j(D)$ are the singular values of $\Gamma_D=\Pi_{\mathcal H_z(D)}|_{\mathcal H_w(D)}:\mathcal H_w(D)\to\mathcal H_z(D)$. In the orthonormal bases $\Sigma_{zz}^{-1/2}R_z$ and $\Sigma_{ww}^{-1/2}R_w$, $\Gamma_D$ has matrix $K_0$, so this agrees with the definition above when the covariances are invertible, and it also covers singular covariances (the one-hot columns of one position sum to one). $D$ is a \emph{compression} of $D'$ if $D_z=T_zD'_z$ and $D_w=T_wD'_w$ for fixed matrices. A column subset is the case $T$ = a row selection of the identity. The linear confounding information is $\mathcal I_z(D)=\M(\proj(\Lb\mid1,X,D_z))$.

\begin{mt}{Theorem}{5}[Dictionary monotonicity]\label{s:thm:monotone}
Let $D$ be a compression of $D'$.
\begin{enumerate}
\item[(a)] $\rho_j(D)\le\rho_j(D')$ for every $j$. If $T_zT_z^\top=I$ and $T_wT_w^\top=I$ (in particular for column subsets), also $\rho_j^\lambda(D)\le\rho_j^\lambda(D')$ for every $\lambda>0$.
\item[(b)] The optimal contrastive loss $-\sum_{j\le d}\rho_j(\lambda)^2$ is no larger on $D'$ (for every $\lambda\ge0$ under the condition of (a)), and $\mathcal I_z(D)\preceq\mathcal I_z(D')$. If (F) holds for $D'$, the $k$ CCA proxies of $D'$ attain $\mathcal I_z(D')$.
\item[(c)] Assume (L1)--(L4) for $D'$ with loadings $\Lambda'$ and noise covariances $\Psi'$, and let $T_z,T_w$ have full row rank (as for column subsets). Then (L1)--(L4) hold for $D$ with loadings $T\Lambda'$ and noise covariances $T\Psi'T^\top$, except possibly the rank condition on the loadings. If $T_z\Lambda'_z$ or $T_w\Lambda'_w$ has rank $<k$, no compression of $D$ identifies $\tau$ (Corollary~\ref{s:cor:effcons}(i)) and we set $\mathcal V^\ast(D)=+\infty$. Then $\mathcal V^\ast(D')\le\mathcal V^\ast(D)$, and CPF on $D'$ attains $\mathcal V^\ast(D')$.
\item[(d)] PCA has no such property. There are $D\subset D'$ (one added column) for which the relevance of PCA-1 drops from $R^2=0.5$ to $0$ while the canonical correlation stays at $0.5$.
\end{enumerate}
\end{mt}

\begin{proof}
(a), $\lambda=0$. Residualization is linear, so $R(D_z)=T_zR(D'_z)$ and $\mathcal H_z(D)\subseteq\mathcal H_z(D')$, $\mathcal H_w(D)\subseteq\mathcal H_w(D')$. For $g\in\mathcal H_w(D)$, $\Pi_{\mathcal H_z(D)}g=\Pi_{\mathcal H_z(D)}\Pi_{\mathcal H_z(D')}g$, so $\Gamma_D=\Pi_{\mathcal H_z(D)}\,\Gamma_{D'}\,\iota$ with $\iota$ the inclusion $\mathcal H_w(D)\hookrightarrow\mathcal H_w(D')$. Both outer factors have norm $\le1$, and Lemma~\ref{s:lem:matrix}(ii) (in orthonormal bases) gives the claim.

(a), $\lambda>0$. Let $F_z=(\Sigma'_{zz}+\lambda I)^{1/2}T_z^\top(T_z\Sigma'_{zz}T_z^\top+\lambda I)^{-1/2}$ and $F_w$ likewise. Since $C(D)=T_zC'T_w^\top$ and $\Sigma_{zz}(D)=T_z\Sigma'_{zz}T_z^\top$, substitution gives $K_\lambda(D)=F_z^\top K_\lambda(D')F_w$. With $T_zT_z^\top=I$, $F_z^\top F_z=(T_z\Sigma'_{zz}T_z^\top+\lambda I)^{-1/2}(T_z\Sigma'_{zz}T_z^\top+\lambda T_zT_z^\top)(T_z\Sigma'_{zz}T_z^\top+\lambda I)^{-1/2}=I$, so $F_z$ and $F_w$ are isometries, and Lemma~\ref{s:lem:matrix}(ii) applies.

(b) The first claim follows from (a) and Theorem~\ref{s:thm:contrastive}(b). Since $\spn(1,X,D_z)\subseteq\spn(1,X,D'_z)$, Lemma~\ref{s:lem:matrix}(iii) gives $\mathcal I_z(D)\preceq\mathcal I_z(D')$. The last claim is Theorem~3(a).

(c) By Theorem~3(b), $\mathcal V^\ast(D)=\sigma_u^{2\ast}(D)\,\zeta^\ast(D)$ with $\zeta^\ast(D)=1/s_A+\alpha^\top H(I_D)^{-1}\alpha$ and $\sigma_u^{2\ast}(D)=\sigma_e^2+\min\{\vartheta^\top T_w\Psi'_wT_w^\top\vartheta:(T_w\Lambda'_w)^\top\vartheta=\beta\}$. \emph{Treatment side.} $H(I_D)=\M(\proj(\Lb\mid R^\circ(D_z)))\preceq H(I_{D'})$ because $\spn(R^\circ(D_z))\subseteq\spn(R^\circ(D'_z))$ (Lemma~\ref{s:lem:matrix}(iii)), and $\zeta$ is Loewner-decreasing (Lemma~\ref{s:lem:matrix}(iv)). \emph{Outcome side.} Every $\vartheta$ feasible for $D$ maps to $\vartheta'=T_w^\top\vartheta$, which is feasible for $D'$ ($\Lambda_w'^\top T_w^\top\vartheta=\beta$) with the same value $\vartheta'^\top\Psi'_w\vartheta'$. Hence the minimum over $D'$ is no larger. Both factors are positive, so $\mathcal V^\ast(D')\le\mathcal V^\ast(D)$. CPF's population directions attain $\mathcal V^\ast(D')$ by Theorem~3(b)(iii).

(d) Let $L,n,n_w$ be independent with unit variance and $s$ independent of them with variance $100$, no $X$, $D_z=(L+n)$, $D'_z=(L+n,s)$ and $D_w=(L+n_w)$. (F) holds for $D'$ with $\Lambda'_z=(1,0)^\top$ and $\Psi'_z=\mathrm{diag}(1,100)$. PCA-1 of $D_z$ is $L+n$, with $R^2(L)=1/2$. The covariance of $D'_z$ is $\mathrm{diag}(2,100)$, so PCA-1 of $D'_z$ is $s$, with $R^2(L)=0$. The canonical correlation is $1/2$ for both, with $K_0(D')=\mathrm{diag}(2,100)^{-1/2}(1,0)^\top\cdot2^{-1/2}=(1/2,0)^\top$.
\end{proof}

\subsection{Proposition 3. Random-phase waveforms}\label{s:sec:prop2}

\begin{mt}{Proposition}{3}[Random-phase waveforms]\label{s:prop:phase}
Let $x_t=\mu\sin(2\pi ft/N+\varphi)+s_t+\xi_t$ for $t=0,\dots,N-1$, where the amplitude $\mu=\mu(L,X)$ and frequency $f=f(L,X)$ (cycles per window) and the phase-locked part $s_t=s_t(L,X)$ are functions of $(L,X)$, $\varphi\sim{\rm Unif}[0,2\pi)$ is independent of $(L,X,\xi,M_z)$, and $\E[\xi\mid L,X,M_z]=0$.
\begin{enumerate}
\item[(a)] $\E[x_t\mid L,X,M_z]=s_t$ for every $t$. Hence, for every fixed $c\in\R^N$ and all measurable $g,h$, $\Cov(c^\top x,g(L,X))=\Cov(c^\top s,g(L,X))$ and $\Cov(c^\top x,h(M_z)\mid X)=\Cov(c^\top s,\E[h(M_z)\mid L,X]\mid X)$. In particular, if $s\equiv0$, every linear function of the raw samples is uncorrelated with every function of $(L,X)$ and, given $X$, with every function of $M_z$. PCA, CCA, linear heads and linear bridges on the raw samples see none of the information in $\mu$ and $f$.
\item[(b)] Let $D_N(\nu)=\sum_{t<N}e^{2\pi i\nu t/N}$, so $|D_N(\nu)|=|\sin\pi\nu/\sin(\pi\nu/N)|$, and let $F_k=\big|\sum_{t<N}(x_t-\bar x)e^{-2\pi i(k/q)t/N}\big|$ be the magnitude spectrum with zero-padding factor $q$ ($q=4$ in CPF). For the oscillatory component ($s\equiv0$, $\xi\equiv0$) and every $\varphi$,
\[
\Big|F_k-\tfrac{|\mu|}2\big|D_N(f-k/q)\big|\Big|\le\tfrac{|\mu|}2\big|D_N(f+k/q)\big|+\tfrac{|\mu|}N\big|D_N(f)\big|\,\big|D_N(k/q)\big| .
\]
The main term depends on $(\mu,f)$, i.e.\ on $(L,X)$, only. The right-hand side is the negative-frequency leakage plus the leakage of the mean-centring. The centring term vanishes when $f$ is an integer with $0<f<N$, and the negative-frequency term when, in addition, $f+k/q$ is an integer not divisible by $N$. In general, $F_k$ of $x$ differs from $F_k$ of the oscillatory component by at most the magnitude of the centred DFT of $s+\xi$ at $k/q$.
\end{enumerate}
\end{mt}

\begin{proof}
(a) Conditionally on $(L,X,\xi,M_z)$, $\E[\sin(\theta+\varphi)]=\int_0^{2\pi}\sin(\theta+\varphi)\,d\varphi/2\pi=0$ for every fixed $\theta$, so the oscillatory term has conditional mean zero given $(L,X,M_z)$. With $\E[\xi\mid L,X,M_z]=0$ this gives $\E[x_t\mid L,X,M_z]=s_t$. Then $\Cov(c^\top x,g)=\E[\E[c^\top x\mid L,X]g]-\E[c^\top x]\E g=\Cov(c^\top s,g)$, and $\E[c^\top x\,h\mid X]=\E[\E[c^\top x\mid L,X,M_z]h\mid X]=\E[c^\top s\,h\mid X]=\E[c^\top s\,\E[h\mid L,X]\mid X]$, since $s$ is a function of $(L,X)$. The same holds for the means.

(b) With $\nu=k/q$ and $\sin\theta=(e^{i\theta}-e^{-i\theta})/2i$,
\[
\sum_{t<N}\sin(2\pi ft/N+\varphi)e^{-2\pi i\nu t/N}=\frac1{2i}\Big(e^{i\varphi}D_N(f-\nu)-e^{-i\varphi}\,\overline{D_N(f+\nu)}\Big).
\]
The mean is $\bar x=\frac\mu N\operatorname{Im}(e^{i\varphi}D_N(f))$, so $|\bar x|\le\frac{|\mu|}N|D_N(f)|$, and centring subtracts $\bar x\,\overline{D_N(\nu)}$. Write the centred DFT as $u+v$ with $u=\frac{\mu}{2i}e^{i\varphi}D_N(f-\nu)$, $|u|=\frac{|\mu|}2|D_N(f-\nu)|$. Then $||u+v|-|u||\le|v|\le\frac{|\mu|}2|D_N(f+\nu)|+|\bar x||D_N(\nu)|$. The formula for $|D_N|$ is the geometric sum $(e^{2\pi i\nu}-1)/(e^{2\pi i\nu/N}-1)$. The last statement is the triangle inequality, since the centred DFT is linear in $x$.
\end{proof}

\subsection{Theorem 6. Finite pools and label-free selection}\label{s:sec:thm5}

\paragraph{Setting.} The pool has $m$ rows. $r_z$ (an $m\times p_z$ matrix) and $r_w$ are the residuals of least squares on $[1,X]$, and $m'=m-\rank[1,X]$. Sample canonical correlations are the singular values of $\Pi_{\col r_z}|_{\col r_w}$ in $\R^m$, and $\hat\rho_j(\lambda)=\sigma_j(\hat K_\lambda)$ with $\hat K_\lambda=(S_{zz}+\lambda I)^{-1/2}S_{zw}(S_{ww}+\lambda I)^{-1/2}$, $S_{zz}=r_z^\top r_z/m$, $S_{zw}=r_z^\top r_w/m$, $S_{ww}=r_w^\top r_w/m$. For selection, the candidates are $c=(B,\lambda)\in\mathcal C$. A seeded permutation, independent of the data, splits the pool into a training part $\mathcal T$ and a validation part of $m_{\rm va}$ rows. For each $c$, directions $(a_{cj},b_{cj})_{j\le d}$ are fitted on $\mathcal T$, $f_{cj}=a_{cj}^\top(D_z-\hat m_z)$ and $g_{cj}=b_{cj}^\top(D_w-\hat m_w)$ with training means $\hat m$, $\hat r_{cj}$ is their Pearson correlation on the validation rows, $\hat J(c)=\sum_j\hat r_{cj}$, and $\hat c\in\arg\max\hat J$. Conditionally on $\mathcal T$, $r_{cj}=\Corr(f_{cj},g_{cj})$ under the law of a pool row, $J(c)=\sum_jr_{cj}$, and $\eta=\max_{c,j}|\hat r_{cj}-r_{cj}|$.

\begin{mt}{Theorem}{6}[Finite pool and label-free selection]\label{s:thm:selection}
\begin{enumerate}
\item[(a)] If $\rank r_z+\rank r_w>m'$, then at least $\rank r_z+\rank r_w-m'$ unregularized sample canonical correlations equal $1$, for every data set and hence under every law, independence included.
\item[(b)] Every $\hat\rho_j(\lambda)$ is non-increasing in $\lambda$, and
\[
\hat\rho_1(\lambda)\le\frac{\lVert S_{zw}\rVert}{\sqrt{(\lambda_{\min}(S_{zz})+\lambda)(\lambda_{\min}(S_{ww})+\lambda)}}\le\frac{\lVert S_{zw}\rVert}{\lambda}.
\]
\item[(c)] $\hat c$ and the final maps are measurable functions of the pool's $(M_z,M_w,X)$ and of the seeds. Treatment, time and event never enter. Hence Theorem~\ref{s:thm:exclusion} holds whatever the selection chose.
\item[(d)] Conditionally on $\mathcal T$, deterministically, $J(\hat c)\ge\max_cJ(c)-2d\eta$. If, conditionally on $\mathcal T$, each standardized $f_{cj}$ and $g_{cj}$ is sub-Gaussian with $\psi_2$-norm at most $\kappa$, then with probability at least $1-\delta$,
\[
\eta\le c_0\kappa^2\Big(\sqrt{\ell/m_{\rm va}}+\ell/m_{\rm va}\Big),\qquad\ell=\log(10|\mathcal C|d/\delta),
\]
for an absolute constant $c_0$.
\end{enumerate}
\end{mt}

\begin{proof}
(a) $\col r_z$ and $\col r_w$ lie in the $m'$-dimensional orthogonal complement of $\col[1,X]$. By Grassmann's formula, $\mathcal J=\col r_z\cap\col r_w$ has dimension $j_0\ge\rank r_z+\rank r_w-m'$. For $y\in\mathcal J$, $\Pi_{\col r_z}y=y$, so the operator $\Pi_{\col r_z}|_{\col r_w}$ is an isometry on $\mathcal J$, and by the min--max characterization its $j_0$-th singular value is at least $\min_{y\in\mathcal J,\lVert y\rVert=1}\lVert\Pi_{\col r_z}y\rVert=1$. All singular values of a product of projections are at most $1$. A unit vector $y=r_za=r_wb\in\mathcal J$ is orthogonal to the constant, so the variates $r_za$ and $r_wb$ have sample correlation exactly $1$.

(b) Monotonicity is Lemma~\ref{s:lem:ridge}(iii) for sample moments. For the bound, $\hat\rho_1(\lambda)=\lVert\hat K_\lambda\rVert\le\lVert(S_{zz}+\lambda I)^{-1/2}\rVert\lVert S_{zw}\rVert\lVert(S_{ww}+\lambda I)^{-1/2}\rVert$, and $\lVert(S+\lambda I)^{-1/2}\rVert=(\lambda_{\min}(S)+\lambda)^{-1/2}\le\lambda^{-1/2}$.

(c) is the third part of the proof of Theorem~\ref{s:thm:exclusion}. The $\arg\max$ over a finite set with a fixed tie-break is measurable.

(d) \emph{Deterministic part.} $|\hat J(c)-J(c)|\le d\eta$ for every $c$. With $c^\ast\in\arg\max J$, $J(\hat c)\ge\hat J(\hat c)-d\eta\ge\hat J(c^\ast)-d\eta\ge J(c^\ast)-2d\eta$.

\emph{Perturbation of one correlation.} Pearson correlation is invariant to affine maps with positive slope, so let $u,v$ be the standardized versions of $f_{cj},g_{cj}$ (population mean $0$, variance $1$, correlation $r$, given $\mathcal T$). On the validation rows let $S_{uu}=\overline{u^2}-\bar u^2$, $S_{vv}$ likewise and $S_{uv}=\overline{uv}-\bar u\bar v$, and suppose $|S_{uu}-1|,|S_{vv}-1|,|S_{uv}-r|\le\epsilon\le\frac12$. Then
\[
|\hat r-r|\le\frac{|S_{uv}-r|}{\sqrt{S_{uu}S_{vv}}}+|r|\Big|\frac1{\sqrt{S_{uu}S_{vv}}}-1\Big|\le\frac{\epsilon}{1-\epsilon}+\frac{\epsilon}{1-\epsilon}\le4\epsilon .
\]
\emph{Concentration.} Conditionally on $\mathcal T$, the validation rows are i.i.d.\ from the pool-row law, because the split is independent of the data. For each $(c,j)$ consider the five centred averages $\bar u$, $\bar v$, $\overline{u^2}-1$, $\overline{v^2}-1$, $\overline{uv}-r$. Products of sub-Gaussians are sub-exponential with $\lVert uv\rVert_{\psi_1}\le\lVert u\rVert_{\psi_2}\lVert v\rVert_{\psi_2}\le\kappa^2$, and centring costs an absolute constant \citep[Lemma~2.7.7, Exercise~2.7.10]{vershynin2018}. Bernstein's inequality \citep[Thm~2.8.1]{vershynin2018} bounds the last three averages by $c_1\kappa^2(\sqrt{L/m_{\rm va}}+L/m_{\rm va})$ with probability $\ge1-2e^{-L}$ each, and Hoeffding's inequality \citep[Thm~2.6.2]{vershynin2018} bounds $|\bar u|,|\bar v|$ by $c_1\kappa\sqrt{L/m_{\rm va}}$. A unit-variance variable has $\kappa\ge c_\ast>0$, so $\kappa\le\kappa^2/c_\ast$. A union bound over the $5|\mathcal C|d$ averages with $L=\ell=\log(10|\mathcal C|d/\delta)$ shows that, with probability $\ge1-\delta$, all of them are at most $b=c_2\kappa^2(\sqrt{\ell/m_{\rm va}}+\ell/m_{\rm va})$. On this event, if $b\le\frac14$, then $\epsilon\le b+b^2\le2b\le\frac12$ and $\eta\le4\epsilon\le8b$. If $b>\frac14$, then $\eta\le2<8b$ trivially. Take $c_0=8c_2$.
\end{proof}

\subsection{Lemma 1. Completeness without Gaussianity}\label{s:sec:lem1}

\begin{mt}{Lemma}{1}[Completeness without Gaussianity]\label{s:lem:complete}
Suppose $\Wt=M\Lb+c(X)+\xi$ with $M\in\R^{d\times k}$ of rank $k$ and $\xi\Ind(L,A)\mid X$. (Pos) holds, and for a left inverse $G$ of $M$ ($GM=I_k$) the characteristic function of $G\xi$ given $X=x$ is non-zero on a dense subset of $\R^k$ for almost every $x$. Then $\Wt$ is complete for $L$ (Definition~\ref{s:def:complete}), whatever the distributions of $L$ and $\xi$. The same holds for $\Zt=M\Lb+c(X)+\xi$ if $\xi\Ind(L,A)\mid X$, which requires $M_z\Ind A\mid L,X$.
\end{mt}
\begin{proof}
Fix $x$ and $a$, and let $g$ satisfy $\E[|g(L)|\mid X=x]<\infty$ and $\E[g(L)\mid\Wt,A=a,X=x]=0$. Since $\sigma(G\Wt)\subseteq\sigma(\Wt)$, the tower property gives, for every bounded measurable $\phi$, $\E[g(L)1\{A=a\}\phi(G\Wt)\mid X=x]=0$. Now $G\Wt=\Lb+Gc(x)+\eta$ with $\eta=G\xi\Ind(L,A)\mid X$. Taking $\phi(y)=e^{it^\top y}$,
\[
0=e^{it^\top Gc(x)}\,\varphi_\eta(t)\,\E\big[g(L)1\{A=a\}e^{it^\top\Lb}\mid X=x\big]=e^{it^\top(Gc(x)-\E[L\mid x])}\,\varphi_\eta(t)\,\hat\mu_g(t),
\]
where $\mu_g(dl)=g(l)\pi_a(l,x)P_{L\mid X=x}(dl)$ is a finite signed measure ($\pi_a(l,x)=P(A=a\mid L=l,X=x)\le1$) and $\hat\mu_g(t)=\int e^{it^\top l}\mu_g(dl)$. Hence $\hat\mu_g=0$ on the dense set where $\varphi_\eta\ne0$, and since $\hat\mu_g$ is continuous, $\hat\mu_g\equiv0$. Write $\mu_g=\mu_g^+-\mu_g^-$ (Jordan decomposition). Then $\mu_g^+$ and $\mu_g^-$ are finite positive measures with equal total mass ($\hat\mu_g(0)=0$) and equal Fourier transforms, so they are equal by the uniqueness theorem for characteristic functions, and $\mu_g=0$. Thus $g\,\pi_a=0$ $P_{L\mid X=x}$-a.e., and since $\pi_a>0$ by (Pos), $g(L)=0$ a.s.\ given $X=x$, hence given $(A,X)=(a,x)$ (positivity makes $P_{L\mid a,x}\ll P_{L\mid x}$). For $\Zt$ the argument is identical.
\end{proof}

\begin{scor}[CPF's proxies are complete]\label{s:cor:cpfcomplete}
Assume (F) for the dictionaries, (P) and (Pos), and let $\Wt$ be CPF's outcome-side proxy with $d\ge k$ directions containing the top-$k$ ridge directions, for any $\lambda\ge0$. Then $\Wt=M\Lb+c(X)+\xi$ with $M=\mathrm{diag}(\sigma)^{-1}\Theta_w^\top\Lambda_w$ of rank $k$ and $\xi=\mathrm{diag}(\sigma)^{-1}\Theta_w^\top\varepsilon_w\Ind(L,A)\mid X$. Hence $\Wt$ is complete for $L$ whenever the characteristic function of $G\xi$ does not vanish on a dense set, e.g.\ if $\varepsilon_w$ is Gaussian, or has independent coordinates with non-vanishing characteristic functions (Gaussian, Laplace). The same holds for $\Zt$ if $M_z\Ind A\mid L,X$.
\end{scor}
\begin{proof}
Under (A1), $\Wt=\mathrm{diag}(\sigma)^{-1}\Theta_w^\top(R_w+\text{affine}(X))$. The block of $\Theta_w^\top\Lambda_w$ formed by the top-$k$ directions is $N_\lambda^\top\Lambda_w^\top(\Sigma_{ww}+\lambda I)^{-1}\Lambda_w$, which is invertible (Lemma~\ref{s:lem:structure}), so $M$ has rank $k$. By (F), $\varepsilon_w\Ind\Lb\mid X$. Given $(L,X)$, $\varepsilon_w$ is a function of $M_w$, so (P) gives $\varepsilon_w\Ind A\mid L,X$. Together, $\varepsilon_w\Ind(L,A)\mid X$. If $\varepsilon_w$ has independent coordinates, $\varphi_{G\xi}(t)=\prod_j\varphi_{\varepsilon_{w,j}}((\Theta_w\mathrm{diag}(\sigma)^{-1}G^\top t)_j)$, which is non-zero everywhere when every factor is.
\end{proof}

\begin{sex}[The condition is not vacuous]\label{s:ex:fejer}
Let $h(l)=\frac1{2\pi}\big(\frac{\sin(l/2)}{l/2}\big)^2$, the Fej\'er density, whose characteristic function is $(1-|t|)_+$. Let $L\sim h$, $\xi\sim h$ independent, $W=L+\xi$, and $A$ independent of $(L,\xi)$. For $g(l)=\cos3l$, $\E[g(L)\mid W=w]=((gh)\ast h)(w)/(h\ast h)(w)$. The Fourier transform of $gh$ is $\frac12(\hat h(t-3)+\hat h(t+3))$, supported in $[2,4]\cup[-4,-2]$, while $\hat h$ is supported in $[-1,1]$, so the Fourier transform of $(gh)\ast h$, their product, is zero. As $(gh)\ast h$ is continuous and integrable, it vanishes identically. The denominator is positive everywhere. Hence $\E[\cos3L\mid W,A]=0$ although $\cos3L\ne0$. $W$ is not complete. With Laplace noise, whose characteristic function never vanishes, Lemma~\ref{s:lem:complete} applies and the same conditional expectation is not identically zero.
\end{sex}

\subsection{Proposition 4. Why bridges are needed}\label{s:sec:prop3}

\paragraph{Setting.} Scalar latent $L$, $A=\pi L+\kappa_A^\top X+\nu$, $\Zt=aL+\delta_z$, $\Wt=bL+\delta_w$ with $a,b\ne0$, and $Y=\tau A+\beta L+\gamma^\top X+e$, where $\nu,\delta_z,\delta_w,e$ are mutually independent, independent of $(L,X)$, mean zero, with finite variances $\sigma_\nu^2,\sigma_z^2,\sigma_w^2,\sigma_e^2$. This model satisfies (P) and (L3). Gaussianity is not needed.

\begin{mt}{Proposition}{4}[Why bridges are needed]\label{s:prop:bridge}
(a) OLS of $Y$ on $(1,A,X,\Zt,\Wt)$ converges to $\tau+\beta\Cov(\check A,\check L)/\Var\check A$, where checks denote residuals on $(1,X,\Zt,\Wt)$. Here $\Cov(\check A,\check L)=\pi\Var\check L$, so the bias is non-zero whenever $\beta\ne0$, $\pi\ne0$ and $\Var\check L>0$ (the proxies do not determine $L$ linearly). (b) 2SLS of $Y$ on $(1,A,X,\Wt)$ with instruments $(1,A,X,\Zt)$ is consistent for $\tau$ whatever $\sigma_z^2$ and $\sigma_w^2$ are, provided $ab\Var(L^\circ)\ne0$ with $L^\circ=L-\proj(L\mid1,A,X)$.
\end{mt}
\begin{proof}
(a) The OLS limit is the population projection coefficient. By the Frisch--Waugh--Lovell theorem it equals $\Cov(\check A,Y)/\Var\check A$ with $\check A=A-\proj(A\mid1,X,\Zt,\Wt)$. Now $\Cov(\check A,Y)=\tau\Var\check A+\beta\Cov(\check A,L)+\gamma^\top\Cov(\check A,X)+\Cov(\check A,e)$. The third term is zero by construction and the fourth because $e$ is independent of $(A,X,\Zt,\Wt)$. Since $\check A\perp\spn(1,X,\Zt,\Wt)$, $\Cov(\check A,L)=\Cov(\check A,\check L)$. Finally $\nu$ is mean zero and independent of $(L,X,\delta_z,\delta_w)$, so $\proj(\nu\mid1,X,\Zt,\Wt)=0$ and $\check A=\pi\check L+\nu$, giving $\Cov(\check A,\check L)=\pi\Var\check L$ and $\Var\check A=\pi^2\Var\check L+\sigma_\nu^2$.

(b) Substituting $L=(\Wt-\delta_w)/b$ gives $Y=\tau A+(\beta/b)\Wt+\gamma^\top X+u$ with $u=e-(\beta/b)\delta_w$. The instruments $(1,A,X,\Zt)$ are functions of $(L,X,\nu,\delta_z)$, and $u$ is a function of $(e,\delta_w)$, which is independent of them, so $\E[u\,(1,A,X,\Zt)]=0$. The model is just identified, and the rank condition reduces, after partialling out $(1,A,X)$, to $\Cov(\Zt^\circ,\Wt^\circ)\ne0$. Since $\delta_z$ and $\delta_w$ are independent of $(A,X,L)$ and of each other, $\Cov(\Zt^\circ,\Wt^\circ)=ab\Var(L^\circ)$. Consistency follows \citep[Thm~5.1]{wooldridge2010}.
\end{proof}

The bias in (a) is $\beta\pi\Var\check L/(\pi^2\Var\check L+\sigma_\nu^2)$.

\subsection{Corollary 1. $X$-conditioning}\label{s:sec:cor1}

\paragraph{Setting.} (F) and (A1) hold, $\E[D_z\mid X]=\mu_z+G_zh_z(X)$ and $\E[D_w\mid X]=\mu_w+G_wh_w(X)$ with affine indices $h_z\in\R^{r_z}$, $h_w\in\R^{r_w}$, $G_z,G_w$ of full column rank and $\Sigma_{h_z}=\Var h_z\succ0$. Plain (unconditional) CCA uses the marginal covariance $\bar\Sigma_{zz}=\Sigma_{zz}+G_z\Sigma_{h_z}G_z^\top$ and cross-covariance $\bar C=C+G_z\Cov(h_z,h_w)G_w^\top$ (note $\Cov(R_z,X)=0$). The $X$-part of $D_z$ is \emph{visible} in $D_w$ if $\col\bar C=\col[\Lambda_z,G_z]$. The bound is $\mathcal I_z=\M(\proj(\Lb\mid1,X,D_z))$.

\begin{mt}{Corollary}{1}[$X$-conditioning]\label{s:cor:xcond}
(a) $X$-conditional CCA with $d\ge k$ attains the bound, $\proj(\Lb\mid1,X,D_z)\in\spn(1,X,\Zt)$, whatever $G_z,G_w$ are. (b) If the $X$-part is visible and $d\ge k+r_z$, plain CCA attains the bound as well. (c) Otherwise plain CCA can fall strictly below the bound. Examples~\ref{s:ex:dsmall} ($d<k+r_z$) and~\ref{s:ex:invisible} (invisible $X$-part, where the non-null plain canonical variates fall below the bound for every $d$).
\end{mt}
\begin{proof}
(a) is Theorem~3(a). (b) Plain CCA's non-null directions span $\bar\Sigma_{zz}^{-1}\col\bar C$ (Lemma~\ref{s:lem:structure}(ii) applied to the marginal pair). There are $\rank\bar C\le k+r_z\le d$ of them, so all are kept. By the Woodbury identity, with $\mathcal Q=\Sigma_{h_z}^{-1}+G_z^\top\Sigma_{zz}^{-1}G_z$,
\[
\bar\Sigma_{zz}^{-1}[\Lambda_z,G_z]=\Sigma_{zz}^{-1}[\Lambda_z,G_z]\begin{pmatrix}I&0\\-\mathcal Q^{-1}G_z^\top\Sigma_{zz}^{-1}\Lambda_z&\mathcal Q^{-1}\Sigma_{h_z}^{-1}\end{pmatrix},
\]
and the right factor is invertible, so $\bar\Sigma_{zz}^{-1}\col\bar C=\col(\Sigma_{zz}^{-1}[\Lambda_z,G_z])\supseteq\col(\Sigma_{zz}^{-1}\Lambda_z)$. For $\theta$ in this space, $\theta^\top(D_z-\E D_z)=\theta^\top R_z+\theta^\top G_z(h_z(X)-\E h_z)$, so $\theta^\top R_z\in\spn(1,X,\Zt_{\rm plain})$. With $\theta$ ranging over $\col(\Sigma_{zz}^{-1}\Lambda_z)$ this gives $\hat L_z\in\spn(1,X,\Zt_{\rm plain})$, and Theorem~3(a) concludes.
\end{proof}

\begin{sex}[$d<k+r_z$]\label{s:ex:dsmall}
Let $X,\Lb,e_1,\dots,e_4$ be independent $N(0,1)$, $D_z=(\Lb+e_1,\,cX+e_2)$, $D_w=(\Lb+e_3,\,cX+e_4)$ with $c^2>1$, and $k=d=1$, $r_z=1$. The bound is $R^2(\Lb\mid X,D_z)=1/2$, attained by $X$-conditional CCA (its direction is the first coordinate, with $\rho=1/2$). Plain CCA has $\bar C=\mathrm{diag}(1,c^2)$, $\bar\Sigma_{zz}=\bar\Sigma_{ww}=\mathrm{diag}(2,1+c^2)$ and $\bar K=\mathrm{diag}(1/2,c^2/(1+c^2))$. Since $c^2/(1+c^2)>1/2$, its single direction is $cX+e_2$, and $R^2(\Lb\mid X,cX+e_2)=0$.
\end{sex}

\begin{sex}[Invisible $X$-part]\label{s:ex:invisible}
Let $X,\Lb,e_1,e_2,e_3$ be independent with unit variance, $D_z=(\Lb+e_1+X,\,e_2+X)$ and $D_w=\Lb+e_3$, so $\Lambda_z=(1,0)^\top$, $\Psi_z=I$, $G_z=(1,1)^\top$ and $\Cov(h_z,h_w)=0$. The bound is $R^2(\Lb\mid X,D_z)=1/2$. Here $\bar C=\Lambda_z\Sigma_L\Lambda_w^\top$ has rank $1$, so plain CCA has a single non-null direction $\theta\propto\bar\Sigma_{zz}^{-1}\Lambda_z\propto(2,-1)^\top$, whose variate is $2(\Lb+e_1)-e_2+X$. Since $X$ is independent of the rest, $R^2(\Lb\mid X,\theta^\top D_z)=R^2(\Lb\mid2(\Lb+e_1)-e_2)=4/9<1/2$. The remaining directions have zero canonical correlation and are not determined by the criterion. Only $d=p_z$, i.e.\ no compression, recovers the bound.
\end{sex}

\subsection{Outcome factors present only on the outcome side}\label{s:sec:asym}

\paragraph{Setting.} The linear proximal model (L1)--(L4), except that $R_w=\Lambda_w\Lb+\Lambda_qQ+\varepsilon_w$ and $Y=c_0+\tau A+\beta^\top\Lb+\eta^\top Q+\gamma^\top X+e$, where $Q\Ind(A,X,\Lb,\varepsilon_z,\varepsilon_w,e)$, $\Var Q=\Sigma_Q$, and $(Q,\varepsilon_w,e)$ is jointly Gaussian. Write $N_w=\Lambda_qQ+\varepsilon_w$ (the private part of the outcome side), $\Psi_{\rm eff}=\Var N_w=\Lambda_q\Sigma_Q\Lambda_q^\top+\Psi_w$, $c_Q=\Cov(N_w,Y)=\Lambda_q\Sigma_Q\eta$, and let $\Pi$ be a basis of $\{x:\Lambda_w^\top x=0\}$. The coordinates of $\Pi^\top R_w=\Pi^\top N_w$ are the private outcome-side directions, uncorrelated with $D_z$ given $X$ and so identifiable without labels. Let $\vartheta_{\rm CCA}=\Psi_{\rm eff}^{-1}\Lambda_w(\Lambda_w^\top\Psi_{\rm eff}^{-1}\Lambda_w)^{-1}\beta$. The class $\mathcal E$ consists of the 2SLS estimators of Section~\ref{s:sec:thm3} with an admissible $P_w$ and, in addition, exogenous private controls $w_S=S^\top R_w$, $\col S\subseteq\col\Pi$, added to both the regressors and the instruments.

\begin{sprop}[Valid, not always efficient, and no label-free rule fixes it]\label{s:prop:asym}
\begin{enumerate}
\item[(a)] (P) holds with latent $L$, so Theorem~\ref{s:thm:exclusion}, Lemma~\ref{s:lem:complete} and the consistency of Lemma~\ref{s:lem:factor} apply to CPF's proxies.
\item[(b)] The law of $(M_z,M_w,X,A)$ does not involve $\eta$. Under Gaussianity it is the law of the model without $Q$ and with outcome-side noise covariance $\Psi_{\rm eff}$. Hence CPF's outcome-side direction is that of Theorem~3(b) for $\Psi_{\rm eff}$, and its feasible weight is $\vartheta_{\rm CCA}$.
\item[(c)] For a member of $\mathcal E$ with weight $\vartheta=\vartheta_{P_w}$ and controls $S$, the asymptotic variance is $\big(\sigma^2_u(\vartheta)-G(\vartheta,S)\big)\zeta(P_z)$, with $u_\vartheta=e+\eta^\top Q-\vartheta^\top N_w$, $\sigma_u^2(\vartheta)=\Var u_\vartheta$ and $G(\vartheta,S)=\Var\proj(u_\vartheta\mid S^\top N_w)$. The class optimum is attained by $\vartheta_{\rm CCA}$ with all private controls ($S=\Pi$), and CPF without controls loses exactly $G_{\max}=c_Q^\top\Pi(\Pi^\top\Psi_{\rm eff}\Pi)^{-1}\Pi^\top c_Q$, which is zero iff $c_Q\in\col\Lambda_w$.
\item[(d)] Let a label-free rule choose, from the law of $(M_z,M_w,X,A)$, an admissible $\col P_w$ and controls $S$ with $\col S\subseteq\col\Pi$ and $\rank S<p_w-k$. Then there are two laws with the same law of $(M_z,M_w,X,A)$ and the same $\beta$ under which the rule is inefficient within $\mathcal E$ in at least one.
\end{enumerate}
\end{sprop}
\begin{proof}
(a) $Q$ and $\varepsilon_w$ are independent of $(A,\varepsilon_z,L,X)$, so $M_w\Ind(A,M_z)\mid L,X$, and $(Q,\varepsilon_w,e)$ is independent of $(\varepsilon_z,A,L,X)$, so $M_z\Ind(Y,M_w)\mid A,L,X$. For any feasible $\vartheta$, $u_\vartheta$ is a function of $(e,Q,\varepsilon_w)$ and hence independent of the instruments, which is all Lemma~\ref{s:lem:factor}(i)--(ii) used.

(b) $\eta$ enters only the law of $Y$, and $N_w\sim N(0,\Psi_{\rm eff})$ independently of the rest.

(c) The controls $w_S=S^\top N_w$ are independent of $(A,X,\Lb,\varepsilon_z)$. Let $u'=u_\vartheta-\proj(u_\vartheta\mid w_S)$. By joint Gaussianity $u'\Ind w_S$, and $(u',w_S)$ is independent of $(A,X,\Lb,\varepsilon_z)$, so $u'$ is independent of all instruments and the structural equation holds with error $u'$ and coefficient on $w_S$ absorbed. The first stage is $\proj(\Wt\mid\mathsf I,w_S)=P_w^\top\Lambda_w\proj(\Lb\mid\mathsf I)+\proj(P_w^\top N_w\mid w_S)$, and $w_S\perp(A,X,G)$, so $\check A$ is unchanged and Lemma~\ref{s:lem:factor}(iii)--(v) give $\Var(u')\zeta(P_z)$ with $\Var u'=\sigma^2_u(\vartheta)-G(\vartheta,S)$. For any feasible $\vartheta$, $\vartheta-\vartheta_{\rm CCA}\in\col\Pi$, so $(\vartheta-\vartheta_{\rm CCA})^\top N_w\in\spn(\Pi^\top N_w)$. Hence the residual of $u_\vartheta$ on $\Pi^\top N_w$ does not depend on $\vartheta$, and by Lemma~\ref{s:lem:matrix}(iii) its variance is at most that of the residual of any $u_\vartheta$ on any $S^\top N_w$ with $\col S\subseteq\col\Pi$. The factor $\zeta(P_z)$ is common to the class and minimized as in Theorem~3(b)(i). At $\vartheta_{\rm CCA}$, $\Cov(\Pi^\top N_w,u_{\vartheta_{\rm CCA}})=\Pi^\top(c_Q-\Psi_{\rm eff}\vartheta_{\rm CCA})=\Pi^\top c_Q$, because $\Psi_{\rm eff}\vartheta_{\rm CCA}\in\col\Lambda_w$, so the loss is $G_{\max}$. Since $\Pi^\top\Psi_{\rm eff}\Pi\succ0$, $G_{\max}=0$ iff $\Pi^\top c_Q=0$ iff $c_Q\in\col\Lambda_w$.

(d) Write $u_{\rm CCA}=\varrho+\proj(u_{\rm CCA}\mid\Pi^\top N_w)$ with $\varrho\perp\spn(\Pi^\top N_w)$. By (c), $\Var\varrho$ is the class optimum. For the rule, $u_\vartheta=u_{\rm CCA}-\delta^\top N_w$ with $\delta=\vartheta-\vartheta_{\rm CCA}\in\col\Pi$, so its residual on $S^\top N_w$ is $\varrho+(\omega-\proj(\omega\mid S^\top N_w))$ with $\omega=\proj(u_{\rm CCA}\mid\Pi^\top N_w)-\delta^\top N_w\in\spn(\Pi^\top N_w)$, and the rule is efficient iff $\omega\in\spn(S^\top N_w)$. Since $\Psi_{\rm eff}\succ0$, $x\mapsto x^\top N_w$ is injective, so this is a condition on vectors.

\emph{Law 0} ($\eta=0$). The projection vanishes, and the rule is efficient iff $\delta\in\col S$.

\emph{Law 1.} Since $\dim\col\Pi=p_w-k>\rank S$, there is $v\in\col\Pi$, $v\ne0$, that is $\Psi_{\rm eff}$-orthogonal to $\col S$. Starting from law 0 with outcome-side noise $N(0,\Psi_{\rm eff})$, set $\dim Q=1$, $\Sigma_Q=1$, $\Lambda_q=\delta_0\Psi_{\rm eff}v$, $\eta=\epsilon/\delta_0$ with $\epsilon\ne0$, and $\Psi_w=\Psi_{\rm eff}-\delta_0^2\Psi_{\rm eff}vv^\top\Psi_{\rm eff}$, which is positive definite once $\delta_0^2v^\top\Psi_{\rm eff}v<1$ (Cauchy--Schwarz in the $\Psi_{\rm eff}$ inner product). Then $N_w\sim N(0,\Psi_{\rm eff})$, so the law of $(M_z,M_w,X,A)$, $\vartheta_{\rm CCA}$ and the rule's choice are unchanged, while $c_Q=\epsilon\Psi_{\rm eff}v$. Writing $v=\Pi b$, $\proj(u_{\rm CCA}\mid\Pi^\top N_w)=N_w^\top\Pi(\Pi^\top\Psi_{\rm eff}\Pi)^{-1}\Pi^\top\Psi_{\rm eff}\Pi b\,\epsilon=\epsilon v^\top N_w$, and the rule is efficient iff $\epsilon v-\delta\in\col S$.

Efficiency in both laws would give $\delta\in\col S$ and $\epsilon v\in\col S$, hence $v\in\col S$. But then $v^\top\Psi_{\rm eff}v=0$ by $\Psi_{\rm eff}$-orthogonality, so $v=0$, a contradiction.
\end{proof}

\end{document}